\documentclass{article}

\usepackage{arxiv}

\usepackage[utf8]{inputenc} % allow utf-8 input
\usepackage[T1]{fontenc}    % use 8-bit T1 fonts
\usepackage{hyperref}       % hyperlinks
\usepackage{url}            % simple URL typesetting
\usepackage{booktabs}       % professional-quality tables
\usepackage{amsfonts}       % blackboard math symbols
\usepackage{nicefrac}       % compact symbols for 1/2, etc.
\usepackage{microtype}      % microtypography
\usepackage{graphicx}
\usepackage{doi}
\usepackage{subcaption}
\usepackage{amssymb, amsmath, amsthm}
\usepackage{algorithm}
\usepackage{algpseudocode}
\usepackage{appendix}
\usepackage{enumitem}

\title{t-SNE Is Not Optimized to Reveal Clusters in Data}

\author{ Zhirong Yang$^{1,2}$\thanks{\texttt{zhirong.yang@ntnu.no}}, Yuwei Chen$^{3}$, and Jukka Corander$^{4,5,6}$\\
$^1$Norwegian University of Science and Technology\\
$^2$Aalto University\\
$^3$Finnish Geospatial Research Institute\\
$^4$University of Oslo\\
$^5$University of Helsinki\\
$^6$Wellcome Sanger Institute}

\renewcommand{\shorttitle}{t-SNE May Not Show Data Clusters}

\hypersetup{
	pdftitle={t-SNE May Not Show Data Clusters},
	pdfsubject={},
	pdfauthor={Zhirong Yang, Yuwei Chen, Denis Sedov, Samuel Kaski, and Jukka Corander},
	pdfkeywords={Clustering, Stochastic Neighbor Embedding, Nonlinear Dimensionality Reduction, Visualization},
}

\begin{document}
	\maketitle

\newcommand{\matA}{\mathbf{A}}
\newcommand{\matB}{\mathbf{B}}
\newcommand{\matC}{\mathbf{C}}
\newcommand{\matD}{\mathbf{D}}
\newcommand{\matE}{\mathbf{E}}
\newcommand{\matF}{\mathbf{F}}
\newcommand{\matG}{\mathbf{G}}
\newcommand{\matH}{\mathbf{H}}
\newcommand{\matI}{\mathbf{I}}
\newcommand{\matK}{\mathbf{K}}
\newcommand{\matL}{\mathbf{L}}
\newcommand{\matM}{\mathbf{M}}
\newcommand{\matN}{\mathbf{N}}
\newcommand{\matO}{\mathbf{O}}
\newcommand{\matP}{\mathbf{P}}
\newcommand{\matQ}{\mathbf{Q}}
\newcommand{\matR}{\mathbf{R}}
\newcommand{\matS}{\mathbf{S}}
\newcommand{\matT}{\mathbf{T}}
\newcommand{\matU}{\mathbf{U}}
\newcommand{\matV}{\mathbf{V}}
\newcommand{\matW}{\mathbf{W}}
\newcommand{\matX}{\mathbf{X}}
\newcommand{\matY}{\mathbf{Y}}
\newcommand{\matZ}{\mathbf{Z}}
\newcommand{\matg}{\mathbf{g}}

\newcommand{\calA}{\mathcal{A}}
\newcommand{\calB}{\mathcal{B}}
\newcommand{\calC}{\mathcal{C}}
\newcommand{\calD}{\mathcal{D}}
\newcommand{\calE}{\mathcal{E}}
\newcommand{\calF}{\mathcal{F}}
\newcommand{\calG}{\mathcal{G}}
\newcommand{\calH}{\mathcal{H}}
\newcommand{\calI}{\mathcal{I}}
\newcommand{\calJ}{\mathcal{J}}
\newcommand{\calK}{\mathcal{K}}
\newcommand{\calL}{\mathcal{L}}
\newcommand{\calM}{\mathcal{M}}
\newcommand{\calN}{\mathcal{N}}
\newcommand{\calO}{\mathcal{O}}
\newcommand{\calP}{\mathcal{P}}
\newcommand{\calQ}{\mathcal{Q}}
\newcommand{\calR}{\mathcal{R}}
\newcommand{\calS}{\mathcal{S}}
\newcommand{\calT}{\mathcal{T}}
\newcommand{\calU}{\mathcal{U}}
\newcommand{\calV}{\mathcal{V}}
\newcommand{\calW}{\mathcal{W}}
\newcommand{\calX}{\mathcal{X}}
\newcommand{\calY}{\mathcal{Y}}
\newcommand{\calZ}{\mathcal{Z}}

\newcommand{\bbA}{\mathbb{A}}
\newcommand{\bbB}{\mathbb{B}}
\newcommand{\bbR}{\mathbb{R}}
\newcommand{\bbZ}{\mathbb{Z}}
\newcommand{\bbE}{\mathbb{E}}
\newcommand{\bbH}{\mathbb{H}}

\newcommand{\veca}{\mathbf{a}}
\newcommand{\vecb}{\mathbf{b}}
\newcommand{\vecc}{\mathbf{c}}
\newcommand{\vecd}{\mathbf{d}}
\newcommand{\vece}{\mathbf{e}}
\newcommand{\vecf}{\mathbf{f}}
\newcommand{\vecg}{\mathbf{g}}
\newcommand{\vech}{\mathbf{h}}
\newcommand{\veci}{\mathbf{i}}
\newcommand{\vecj}{\mathbf{j}}
\newcommand{\veck}{\mathbf{k}}
\newcommand{\vecl}{\mathbf{l}}
\newcommand{\vecm}{\mathbf{m}}
\newcommand{\vecn}{\mathbf{n}}
\newcommand{\veco}{\mathbf{o}}
\newcommand{\vecp}{\mathbf{p}}
\newcommand{\vecq}{\mathbf{q}}
\newcommand{\vecr}{\mathbf{r}}
\newcommand{\vecs}{\mathbf{s}}
\newcommand{\vect}{\mathbf{t}}
\newcommand{\vecu}{\mathbf{u}}
\newcommand{\vecv}{\mathbf{v}}
\newcommand{\vecw}{\mathbf{w}}
\newcommand{\vecx}{\mathbf{x}}
\newcommand{\vecy}{\mathbf{y}}
\newcommand{\vecz}{\mathbf{z}}

\newcommand{\vecalpha}{\boldsymbol{\alpha}}
\newcommand{\vecbeta}{\boldsymbol{\beta}}
\newcommand{\veceta}{\boldsymbol{\eta}}
\newcommand{\vectheta}{\boldsymbol{\theta}}
\newcommand{\vecphi}{\boldsymbol{\phi}}
\newcommand{\vecpsi}{\boldsymbol{\psi}}
\newcommand{\vecrho}{\boldsymbol{\rho}}
\newcommand{\vectau}{\boldsymbol{\tau}}
\newcommand{\vecmu}{\boldsymbol{\mu}}
\newcommand{\veceps}{\boldsymbol{\epsilon}}
\newcommand{\vecxi}{\boldsymbol{\xi}}
\newcommand{\vecPhi}{\boldsymbol{\Phi}}
\newcommand{\vecDelta}{\boldsymbol{\Delta}}

\newcommand{\matDelta}{\boldsymbol{\Delta}}
\newcommand{\matEta}{\boldsymbol{\eta}}
\newcommand{\matOmega}{\boldsymbol{\Omega}}
\newcommand{\matPhi}{\boldsymbol{\Phi}}
\newcommand{\matPsi}{\boldsymbol{\Psi}}
\newcommand{\matTheta}{\boldsymbol{\Theta}}
\newcommand{\matLambda}{\boldsymbol{\Lambda}}
\newcommand{\matSigma}{\boldsymbol{\Sigma}}
\newcommand{\matzero}{\mathbf{0}}
\newcommand{\IndexSetI}{\mathcal{I}}
\newcommand{\grad}{\mathcal{\nabla}}

\newcommand{\vecone}{\mathbf{1}}
\newcommand{\veczero}{\mathbf{0}}

\def\maximize{\mathop{{\mathgroup\symoperators maximize}}}
\def\Maximize{\mathop{{\mathgroup\symoperators Maximize}}}
\def\minimize{\mathop{{\mathgroup\symoperators minimize}}}

\def\approach{\mathop{{\mathgroup\symoperators \longrightarrow}}}
\def\defineoperator{\mathop{{\mathgroup\symoperators =}}}
\newcommand{\define}{\defineoperator^{\text{def}}}

\newcommand{\Tr}{\text{Tr}}
\newcommand{\trace}{\text{trace}}
\newcommand{\diag}{\text{diag}}
\newcommand{\gradWJ}{\nabla_{\scriptscriptstyle{\matW}}\calJ}
\newcommand{\const}{\text{constant}}
\newcommand{\fracpartial}[2]{\frac{\partial #1}{\partial  #2}}

\newcommand{\defeq}{\stackrel{\text{def}}{=}}

\begin{abstract}
Cluster visualization is an essential task for nonlinear dimensionality reduction as a data analysis tool. It is often believed that Student t-Distributed Stochastic Neighbor Embedding (t-SNE) can show clusters for well clusterable data, with a smaller Kullback-Leibler divergence corresponding to a better quality. There was even theoretical proof for the guarantee of this property. However, we point out that this is not necessarily the case---t-SNE may leave clustering patterns hidden despite strong signals present in the data. Extensive empirical evidence is provided to support our claim. First, several real-world counter-examples are presented, where t-SNE fails even if the input neighborhoods are well clusterable. Tuning hyperparameters in t-SNE or using better optimization algorithms does not help solve this issue because a better t-SNE learning objective can correspond to a worse cluster embedding. Second, we check the assumptions in the clustering guarantee of t-SNE and find they are often violated for real-world data sets.
	%Finally, why t-SNE fails and a potential solution to overcome this are also discussed.
\end{abstract}

% keywords can be removed

\section{Introduction}
\label{sec:intro}
The rapid growth in the amount of data processed by analysts demands more efficient information digestion and communication methods. Data visualization by dimensionality reduction facilitates a viewer to digest information in massive data sets quickly. Therefore, it is increasingly applied as a critical component in scientific research, digital libraries, data mining, financial data analysis, market studies, manufacturing production control, drug discovery, etc.

Stochastic Neighbor Embedding (SNE) \cite{hinton2003sne} is a widely used nonlinear dimensionality reduction (NLDR) method, which approximately preserves the pairwise probabilities of being neighbors (neighboring probabilities for short) in the input space. In particular, the Student t-Distributed Stochastic Neighbor Embedding (t-SNE) \cite{vandermaaten2008tsne} has become one of the most popular nonlinear dimensionality reduction methods for data visualization. The t-SNE method employs a heavy-tailed distribution for the neighboring probabilities in the embedding and minimizes their Kullback-Leibler divergence against the precomputed input probabilities.

Discovery of large-scale patterns such as clusters is an essential task of NLDR. It is often believed that t-SNE can show clusters for well clusterable data, with a smaller Kullback-Leibler divergence corresponding to a better quality. Earlier work has indeed derived theoretical proof for the guarantee of this property \cite{tsneprove,tsneanalysis}. 

However, recently we found many counter-examples where t-SNE may not correctly visualize the clusters even if the input neighborhoods are well clusterable. We have tried to use different hyperparameters in t-SNE or different optimization algorithms, which nonetheless does not help solve this issue because a better t-SNE learning objective can correspond to a worse cluster embedding.

We find that the theoretical clustering guarantees of t-SNE in \cite{tsneprove,tsneanalysis} are problematic as well. Their proof requires a bunch of assumptions on the data distribution. However, these assumptions are often violated when we check them for various real-world data sets, which means the assumptions are oversimplified, and the so-called proven guarantee is often unrealistic.

The remainder of the paper is organized as follows. We first briefly review t-SNE in Section \ref{sec:tsne} and define our research question in \ref{sec:cluvis}. Next, in Section \ref{sec:counterexamples}, we demonstrate several counter-examples where t-SNE fails to show the data clusters. We have tried various ways to tune t-SNE and using other optimization algorithms for t-SNE, where the results are presented in Section \ref{sec:tuningtsne} and \ref{sec:tsneopt}, respectively. We then report the checking of the assumptions in the existing theoretical guarantee in Section \ref{sec:assumptionchecking}. Finally we give conclusion and discussion in Section \ref{sec:conclusion}.

\section{Student t-Distributed Stochastic Neighbor Embedding}
\label{sec:tsne}
Student t-Distributed Stochastic Neighbor Embedding (t-SNE) is a Nonlinear Dimensionality Reduction (NLDR) method. Given a set of multivariate data points $\{x_1,x_2,\dots,x_N\}$, where $x_i\in\bbR^D$, their pairwise similarities are encoded in a nonnegative square matrix $P$. t-SNE finds a mapping $x_i\mapsto y_i\in\bbR^d$ for $i=1,\dots,N$ ($d=2$ or $d=3$ for visualization) such that the similarities in the mapped space, encoded by $Q_{ij}$, approximate those in $P$.

The t-SNE method comprises two major steps. First, it converts vectorial data $\{x_1,x_2,\dots,x_N\}$ to similarity matrix $P$. 
Second, t-SNE finds $y_i$'s given $P$ by minimizing the Kullback-Leibler (KL) divergence between $P$ and $Q$ over $Y=\{y_i\}_{i=1}^N$:
\begin{align}
\minimize_{Y}~D_\text{KL}(P||Q)=\sum_{ij}P_{ij}\ln\frac{P_{ij}}{Q_{ij}},
\end{align}
where $Q_{ij}=q_{ij}/\sum_{ab}q_{ab}$ and $q_{ij}=(1+\|y_i-y_j\|^2)^{-1}$. Here the matrixwise summation is over off-diagonal elements, i.e.~$\sum_{ij}A_{ij}\defeq\sum_{ij:i\neq j}A_{ij}$.

To our knowledge, the first step is a largely unsolved problem, and there is no universally best solution. In practice, users often use Entropic Affinity (EA) \cite{hinton2003sne} or $k$-Nearest Neighbor ($k$-NN) and tune the perplexity (or $k$) parameter. In this study, we focus on the second step with a precomputed matrix $P$.

The original t-SNE optimization algorithm employs gradient descent with momentum:
\begin{align}
y_i(t+1) = y_i(t) - \eta \fracpartial{D_\text{KL}(P||Q)}{y_i} + \zeta\cdot[y_i(t)-y_i(t-1)],
\end{align}
where $\eta>0$ is the step size and $\zeta>0$ is the momentum extent. The coordinates $y_i$'s are initialized with small numbers, e.g.~sampling from $\calN(0, 10^{-4})$. During the first 250 iterations, the algorithm replaces $P$ with $\beta P$, where $\beta=4$ in \cite{vandermaaten2008tsne} or $\beta=12$ in \cite{vandermaaten2014bhtsne}. This is called ``early-exaggeration'' which encourages the mapped data points to form tighter clumps, with more empty space in the visualization to facilitates subsequent optimizing iterations.

\section{Cluster Visualization}
\label{sec:cluvis}
A clustering divides the data objects into a number of groups, such that objects in the same group (called a cluster) are more similar to each other than to those in other groups (clusters) \cite{tandatamining}. The pairwise similarities can be encoded in a similarity matrix.

A similarity matrix is (well) clusterable if there is a clustering such that high similarities appear (much) more probably within clusters than between clusters. To verify whether a similarity matrix is (well) clusterable or not, we can sort its rows and columns according to the cluster labels. Because permutation does not change the clusterability, we should observe a (clear) diagonal blockwise pattern for a (well) clusterable similarity matrix, where high similarities are (much) denser within the cluster blocks, and otherwise no such pattern. Based on this definition, a data set is well clusterable if its similarity matrix is well-clusterable.

A good cluster visualization is a display where the user can easily see the groups of data points. In scatter plots, there should be clear space separating the groups such that points in the same group are closer to each other than to those in other groups. The clustering shown in the visualization should well correspond to the diagonal blockwise pattern in the similarity matrix sorted as described above. Note that cluster visualization is an unsupervised task where class labels are generally not available. Even if there exist class labels, they may not be aligned with the intrinsic data clusters.

Showing clusters in a clusterable data set is an essential application of t-SNE. In this work, we consider the following research question: \emph{given a well-clusterable similarity matrix $P$, can t-SNE correctly visualize the clusters by minimizing $D_\text{KL}(P||Q)$?}
Because of the apparent success, it is often believed that the answer is true. Some researchers even claimed that t-SNE can provably find the clusters (e.g., \cite{tsneprove,tsneanalysis}).

\section{Counter-Examples of t-SNE}
\label{sec:counterexamples}
However, we discovered that despite apparent successes, t-SNE can also be prone to produce false-negative results. That is, t-SNE results show no clear clusterings for $P$ matrices which are actually well clusterable. In the following, we present five of such counter-example data sets.
\begin{itemize}
	\item \texttt{SHUTTLE}: the Statlog (Shuttle) Data Set in the UCI repository\footnote{available at \url{https://archive.ics.uci.edu/ml/datasets/Statlog+(Shuttle)}}. There are 58000 samples of 9 dimensions in three large and four small classes.
	
	\item \texttt{IJCNN}: the IJCNN 2001 neural network competition data\footnote{available at \url{https://www.csie.ntu.edu.tw/~cjlin/libsvmtools/datasets/binary.html}}. There are 126,701 samples of 22 dimensions and from ten engines (classes).
	
	\item \texttt{TOMORADAR}: The data was collected via a helicopter-borne microwave profiling radar \cite{chen2017uav} termed FGI-Tomoradar to investigate the vertical topography structure of forests.  After preprocessing, the data set contains 120,024 samples of 8192 dimensions from three classes.

	\item \texttt{FLOW-CYTOMETRY}: the single-cell biology data set collected from Flow Repository\footnote{available at \url{https://flowrepository.org/id/FR-FCM-ZZ36}}. After preprocessing, the data set contains 1,000,000 samples of 17 dimensions.
	
	\item \texttt{HIGGS}: the HIGGS Data Set in the UCI repository\footnote{available at \url{https://archive.ics.uci.edu/ml/datasets/HIGGS}}. The data was produced using Monte Carlo simulations of the particles in a physics experiment. There are 11,000,000 data points of 28 dimensions. Previously the data were used for classification between the bosons and the background particles, whereas there is little research on unsupervised learning on the data. Here we compared visualizations to discover the particle clusters.
\end{itemize}

These data sets are well-clusterable because we can get their well-clusterable similarity matrices by using EA or $k$-NN. We have constructed the $P$ matrix by using EA with perplexity 30 for \texttt{SHUTTLE} and \texttt{IJCNN}. We have used symmetrized $k$-NN graph adjacency matrix as $P$ for \texttt{TOMORADAR}, \texttt{FLOW-CYTOMETRY} and \texttt{HIGGS} with $k=50$, $k=15$ and $k=5$, respectively. In this way, the constructed similarity matrices are well clusterable because they comprise diagonal blockwise pattern, as shown in Figure \ref{fig:blockvis}.
%In Section \ref{} we will discuss other $P$ choices.

We have used the official implementation of t-SNE\footnote{available at \url{https://github.com/lvdmaaten/bhtsne}} in \cite{vandermaaten2014bhtsne}, where the maximum iterations in t-SNE were set to 10000 (ten times as the default) to get closer to convergence. The resulting t-SNE visualizations are shown in Figure \ref{fig:tsnevis}.

We can see that t-SNE cannot find clustering patterns in all of the five data sets. For \texttt{SHUTTLE}, \texttt{IJCNN}, \texttt{TOMORADAR}, the t-SNE layouts overall look like a single diamond with too many small groups, where no major clusters can be identified. For \texttt{FLOW-CYTOMETRY} the t-SNE visualization is nearly a single ball, while for \texttt{HIGGS} it is just a hairball.

\section{Tuning t-SNE Does Not Help in Cluster Visualization}
\label{sec:tuningtsne}
We have seen that t-SNE in its default setting cannot show the clusters in the tested data sets. Next, we study some extended t-SNE versions to see whether they can give the correct cluster visualization. Here we used only the three smallest data sets \texttt{SHUTTLE}, \texttt{IJCNN}, and \texttt{TOMORADAR} due to computing load. 

There are several t-SNE variants with different hyperparameters. Similar to the original t-SNE, they comprise two steps: 1) constructing a similarity matrix $P$ and 2) minimizing $D_\text{KL}(P||Q(Y))$ over $Y$, where $Q_{ij}=q_{ij}/\sum_{ab}q_{ab}$ with
\begin{align}
q_{ij}=\left(1+\frac{\|y_i-y_j\|^2}{\nu}\right)^{-\frac{1+\nu}{2}}.
\end{align}
By default, the t-SNE software uses $\nu=1$ and Entropic Affinity \cite{hinton2003sne} with perplexity equal to 30 to construct $P$.

We first used various perplexity ranging from 10 to 150 in constructing $P$; the results for the data sets \texttt{SHUTTLE}, \texttt{IJCNN}, and \texttt{TOMORADAR} are given in Figures \ref{fig:perpsshuttle} to \ref{fig:perpstomoradar}. We can see that t-SNE still cannot find the major clusters in the wide range of perplexity.

Next we used various $\nu$ ranging from 1 to 15; the results are given in Figures \ref{fig:degreesshuttle} to \ref{fig:degreestomoradar}. No major clustering pattern can be seen in these visualizations, which indicates that tuning perplexity does not help either.

% =============================================================================

\section{Other t-SNE Optimization Algorithms}
\label{sec:tsneopt}
It remains questionable whether the failure reason comes from the original t-SNE optimization algorithm. This section presents the results using two other optimization algorithms that are superior in finding better t-SNE objectives for some data sets.

The first alternative optimization algorithm uses Majorization-Minimization (MM) \cite{yang2015mm}. It has two advantages: 1) because partial Hessian information is used, the algorithm can avoid many poor local optima; 2) it guarantees that the objective monotonically decreases after each iteration and thus converges to a stationary point.

Another algorithm we considered is opt-SNE \cite{optsne}. The algorithm automates the selection of three critical parameters for the t-SNE run: 1) initial learning rate, 2) the number of iterations spent in early exaggeration, and 3) the number of total iterations.

The resulting visualizations are shown in Figures \ref{fig:opt_shuttle} to \ref{fig:opt_tomoradar}. We include the visualization with the original algorithm again to ease comparison. The t-SNE objectives are shown in each visualization.

We can see that the two alternative optimization algorithms often find lower (better) t-SNE objectives (except opt-SNE for \texttt{SHUTTLE}). Especially the MM algorithm achieves the lowest for all three data sets. However, the better t-SNE objectives do not correspond to good cluster visualizations compared to the known diagonal blockwise pattern in the similarity matrices. There are still too many small pieces without major clusters identified.

By contrast, we also include visualizations by using another method called Stochastic Cluster Embedding (SCE) \cite{sce}. Although the method employs another learning objective, we can still compute the t-SNE objective by using its resulting coordinates. The results are shown in the (d) panel in Figures \ref{fig:opt_shuttle} to \ref{fig:opt_tomoradar}.

From the figures, we can see that SCE gives much better cluster visualizations than all t-SNE results in showing the clusters clearly. It is more important to notice that the SCE results correspond to higher t-SNE objectives for all three data sets. The effort of searching a lower t-SNE objective does not bring a better cluster visualization, which indicates that the problem is not in the optimization but in the t-SNE objective itself.

% =============================================================================

\section{On Theoretical Guarantees that t-SNE Can Show Data Clusters}
\label{sec:assumptionchecking}
In the above, we have seen counter-examples where t-SNE fails to show the clusters even if the data set is well clusterable and we have tried hard to tune it. The phenomenon contradicts the apparent theoretical findings that t-SNE is guaranteed to show data clusters \cite{tsneprove,tsneanalysis}. In this section, we review their theoretical claims and check their assumptions on the tested data sets\footnote{The code for the assumption checking is available at \url{https://ntnu.box.com/s/1yyf5eisncgfudij2ggv42o6v5nngec1}}.  

\subsection[Guarantee and violations of assumptions by Linderman and Steinerberger]{Guarantee and violations of assumptions in \cite{tsneprove}}
Their main results require the following assumptions:
\begin{itemize}
	\item there exists a clustering such that for all $x_i$ and $x_j$ which belong to the same cluster
	\begin{align}
	\label{eq:linderman1}
	P_{ij}\geq \frac{1}{10N\cdot \Omega(i)},
	\end{align}
	where $\Omega(i)$ is the size of cluster which $x_i$ and $x_j$ belong to;
	\item $\beta$ and $\eta$ are suitably chosen;
	\item $y_i$'s are initialized with small numbers.
\end{itemize}
If the above assumptions hold, Linderman and Steinerberger proved that the diameter of an embedded cluster decays exponentially until it is smaller than a constant determined by $\eta$, $\beta$, and $P$ \cite{tsneprove}, and the t-SNE algorithm is guaranteed to show data clusters.

However, the above condition is generally not true because there can be various cluster distributions. Two points $x_i$ and $x_j$ within a cluster can be distant, and their $P_{ij}$ can be very small or even close to zero.

To illustrate this, let us first look at the \texttt{2-MOON} and \texttt{2-ROLL} examples in Figure \ref{fig:synthetic}, where each data set clearly has two clusters. Such two synthetic examples are widely used in cluster analysis and manifold research. We followed the same method in t-SNE to construct the $P$ matrices by using Entropic Affinity (EA), which employs Gaussian kernels with the variances set according to a specific perplexity \cite{hinton2003sne,vandermaaten2008tsne}. We used the default perplexity=30 in the t-SNE software. We can see that $P_{ij}$ is large only for neighboring data points. When the pair of data points are more distant along the curved clusters, their $P_{ij}$ values decay quickly. In total, there are $92.29\%$ and $94.71\%$ pairs violating the assumption in Eq.~\ref{eq:linderman1}, respectively, for the two data sets.

Besides synthetic data sets, we have also verified the assumption on several real-world data sets that are well clusterable (see e.g.~\cite{vandermaaten2008tsne,yang2016jmlr}).
First we try the typical t-SNE benchmark data sets \texttt{COIL20}\footnote{available at \url{https://www.cs.columbia.edu/CAVE/software/softlib/coil-20.php}} and \texttt{DIGITS}\footnote{available at \url{https://archive.ics.uci.edu/ml/datasets/optical+recognition+of+handwritten+digits}}.
We checked the $P$ matrices calculated by the original t-SNE algorithm. We find that violations occur for $23.12\%$ and $47.97\%$ of the within-cluster data point pairs, respectively, for the two data sets.

Surprisingly, we find that even the ``real-life data'' \texttt{MNIST10K} in their work (see \cite{tsneprove} Section 2.2) fails their assumption as well. Finally, we check the $P$ matrices of the tested data sets in our paper. None of them fulfills the assumption.

In t-SNE, we replace the default perplexity with other values ranging from 5 to 500 in calculating $P$, but violations of the assumption still occur. The violation percentages are reported in Table \ref{tab:lindermancheck}.
Note that perplexity equals the average number of neighbors in the Gaussian neighborhoods. Even larger perplexity values obfuscate the cluster boundary, and only a single big clump remains.

In summary, the ``clustered assumption'' is violated for all the above data sets, both synthetic and real-world, with all tested perplexities. More details about the empirical checking can be found in our Matlab scripts.

\subsection[Guarantee and violations of assumptions by Arora et al.]{Guarantee and violations of assumptions in \cite{tsneanalysis}}
Let $\calC_1, \calC_2,\dots,\calC_k$ be the individual clusters such that for each $l\in[k]$, $|\calC_k|\geq 0.1(N/k)$. The data $X=\{x_i\}_{i=1}^N$ is said to be $\gamma$-spherical and $\gamma$-well-separated if for some $b_1, b_2, \dots, b_k>0$:
\begin{itemize}
    \item 1) $\gamma$-spherical:
    \begin{itemize}
        \item 1a) for any $l\in[k]$ and $i,j\in\calC_l$ ($i\neq j)$, we have $\|x_i-x_j\|^2\geq\frac{b_l}{1+\gamma}$, and 
        \item 1b) for any $i\in\calC_l$ we have 
    $\left|\left\{j\in\calC_l\setminus \{i\}: \|x_i-x_j\|^2\leq b_l\right\}\right|\geq 0.51|\calC_l|$.
    \end{itemize}
    \item 2) $\gamma$-well-separated: for any $l,l'\in [k]$ ($l\neq l')$, $i\in\calC_l$ and $j\in\calC_{l'}$ we have $\|x_i-x_j\|^2\geq(1+\gamma\log N)\max\{b_l, b_{l'}\}$.
\end{itemize}
Their guarantee says if $X$ is $\gamma$-spherical and $\gamma$-well-seperated clusterable data, with $\calC_1, \calC_2, \dots, \calC_k$ defining the individual clusters, t-SNE with early exaggeration on input $X$ outputs a good cluster visualization of X with high probability.

These conditions require that the clusters are roughly spherical and all between-cluster distances should be sufficiently large compared to the sphere sizes. However, these assumptions are oversimplified and usually do not hold when the clusters are distributed in curved manifolds. To see this, look at the two synthetic examples in Figure \ref{fig:synthetic}, where neither $\gamma$-Spherical nor $\gamma$-Well-separated condition is met. Below we will show that these conditions do not hold for real-world data sets either.

For a given data set, we have tried various $\gamma$-values in a wide range: $\gamma=10^h$, where $h\in[-10,10]$. For each $\gamma$, we first identified the smallest $b_l$'s by using Condition 1b), which can be done by using the following bisecting method
\begin{enumerate}
	\item $b_l^\text{min}\leftarrow0$, 
	$b_l^\text{max}\leftarrow\max_{i\in\calC_l,j\in\calC_l}\|x_i-x_j\|^2$
	\item Repeat until $b_l^\text{max}-b_l^\text{min}<\epsilon$
	\begin{enumerate}
		\item $b_l^\text{mid}\leftarrow(b_l^\text{min}+b_l^\text{max})/2$
		\item If Condition 1b) holds with $b_l=b_l^\text{mid}$,\\
		then $b_l^\text{max}\leftarrow b_l^\text{mid}$.\\
		Otherwise $b_l^\text{min}\leftarrow b_l^\text{mid}$
	\end{enumerate}
	\item return $b_l\leftarrow b_l^\text{mid}$
\end{enumerate}

Next we used these $b_l$'s and $\gamma$ to check the conditions 1a) and 2) by inspecting each $(i,j)$ pair. The checks are sufficient because even smaller $b_l$'s will fail 1b). Similarly, if these $b_l$'s already cause violation in 2), then even larger $b_l$'s will produce more failures. Therefore, if we find violations of 1a) or 2), then there will be no suitable $b_l$'s that fulfill both $\gamma$-Spherical and $\gamma$-Well-separated conditions, and the claims in \cite{tsneanalysis} will not hold.

We have checked the conditions for the two synthetic datasets in Figure \ref{fig:synthetic}, two typical t-SNE benchmark data sets \texttt{COIL20}, and \texttt{DIGITS}, and the three experimented data sets \texttt{SHUTTLE}, \texttt{IJCNN}, \texttt{TOMORADAR}. All these data sets are known to be well clusterable (see Section \ref{sec:counterexamples} or the cluster analysis literature e.g.~\cite{yang2016jmlr}).

However, we observe violations for all the checked data sets. The violated percentages of the $(i,j)$ pairs are reported in Table \ref{tab:aroracheck}. We can see that Condition 1a for $\gamma$-Spherical is fulfilled only when $\gamma$ is large. Even worse, Condition 2 for $\gamma$-Well-separated is always violated for all data sets and all $\gamma$ values.  

In summary, none of the tested datasets fulfills both the $\gamma$-Spherical and $\gamma$-Well-separated conditions. More details about the empirical checking can be found in our Matlab scripts.

%Moreover, these papers are only preprints and not peer-reviewed. It is unknown whether their proofs are rigorously correct.

\section{Conclusion}
\label{sec:conclusion}
We have presented counter-examples from various domains for t-SNE cluster visualization, where t-SNE fails to identify the correct clustering although the data sets are well clusterable. Tuning perplexity or degree in t-SNE cannot solve the problem. Using other optimization algorithms does not help either because a lower t-SNE loss does not correspond to a good cluster visualization. We have also found that the conditions of two existing theoretical guarantees about t-SNE clustering are often violated for both synthetic and various real-world data sets.

The above evidence shows that t-SNE may not find data clusters even for well clusterable data sets, making this popular visualization method questionable in exploratory data analysis. Better NLDR methods are needed to overcome the problem. Some recent work, for example, \cite{umap,sce}, could give better cluster visualizations. Furthermore, it is necessary for theoretical study on NLDR for clustering with more realistic conditions.

\clearpage

\newcommand{\spyfigwidth}{5.2cm}
\begin{figure}[t]
	\begin{center}
		\subcaptionbox{SHUTTLE}
		{\includegraphics[width=\spyfigwidth,height=\spyfigwidth]{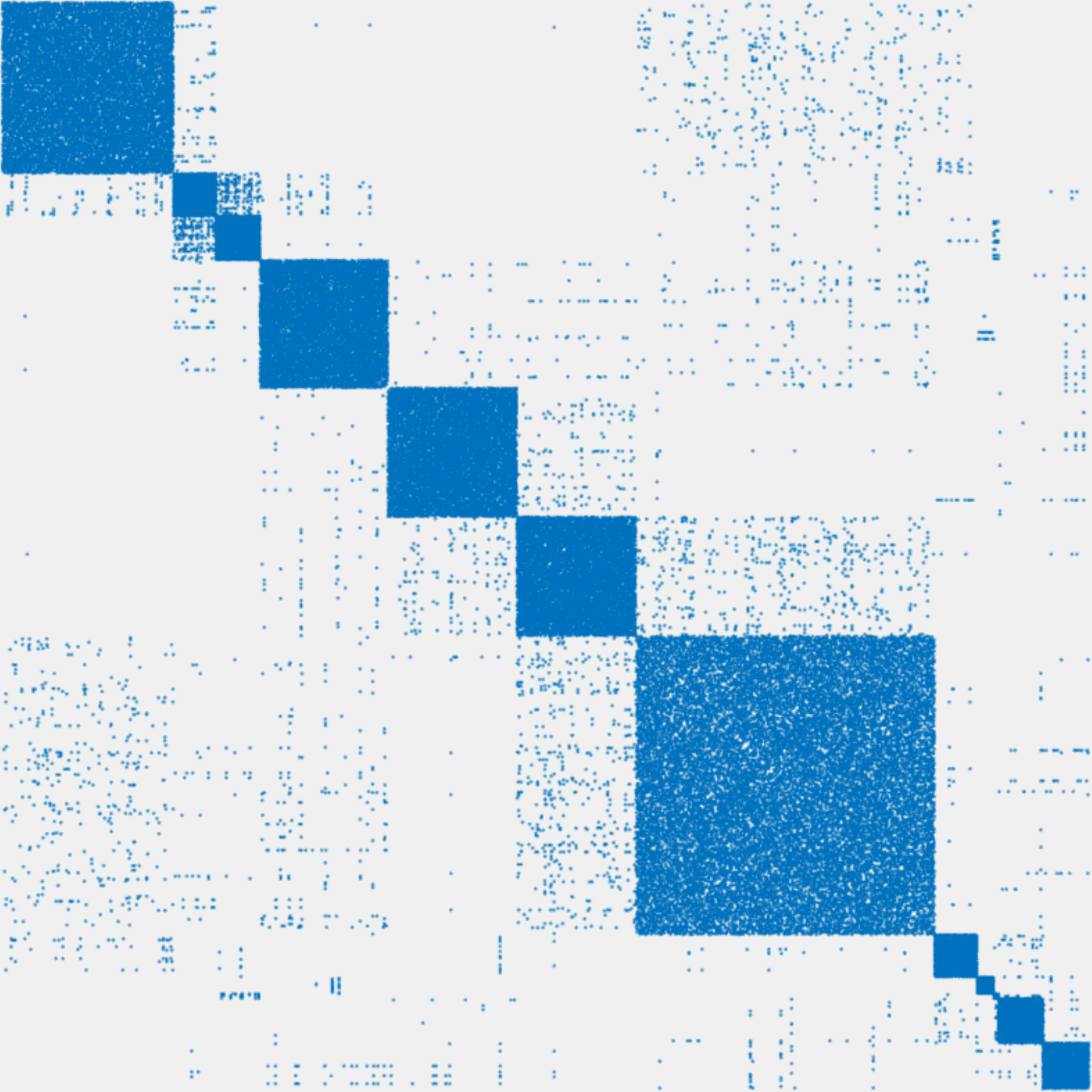}}
		\subcaptionbox{IJCNN}
		{\includegraphics[width=\spyfigwidth,height=\spyfigwidth]{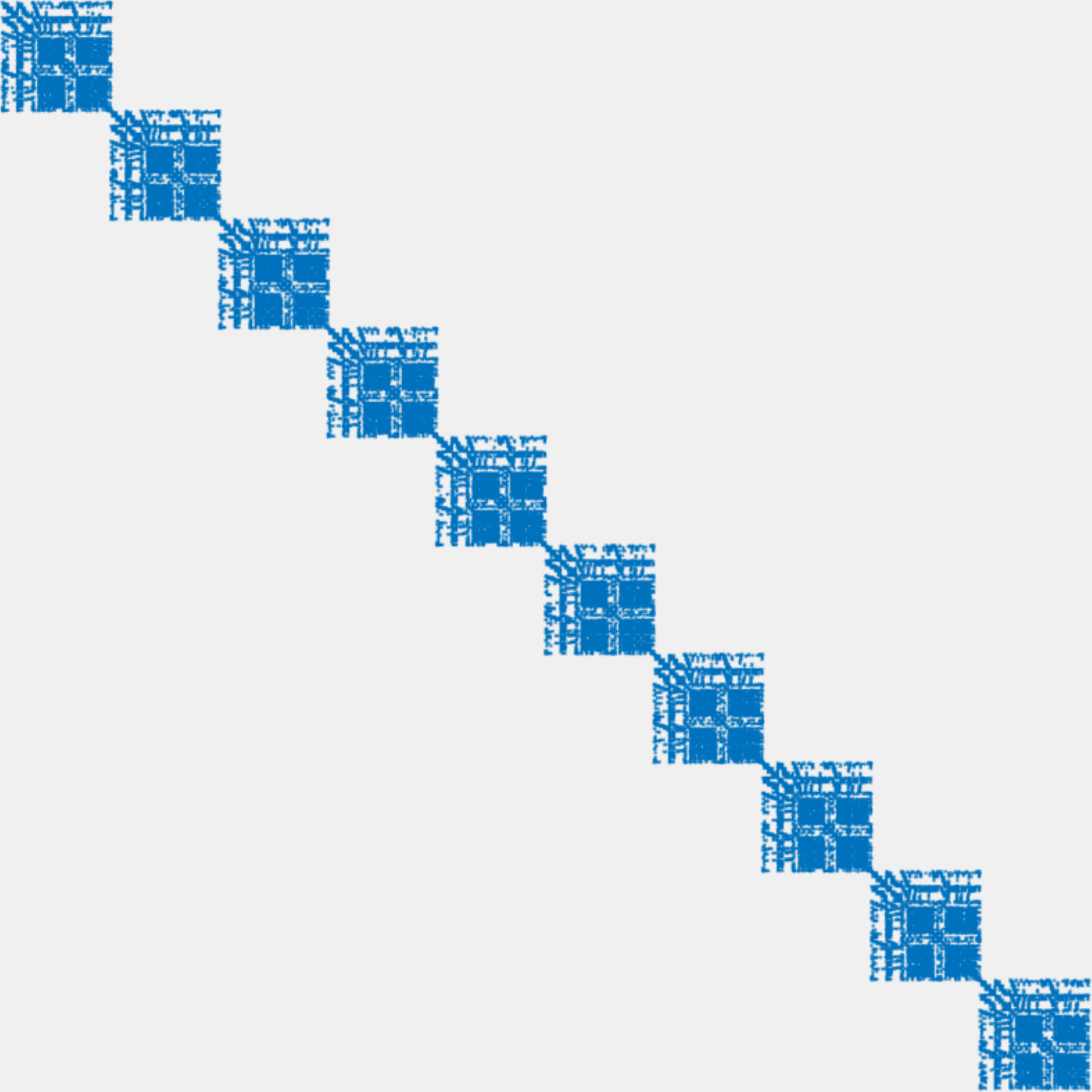}}
		
\vspace{\baselineskip}
		
		\subcaptionbox{TOMORADAR}
		{\includegraphics[width=\spyfigwidth,height=\spyfigwidth]{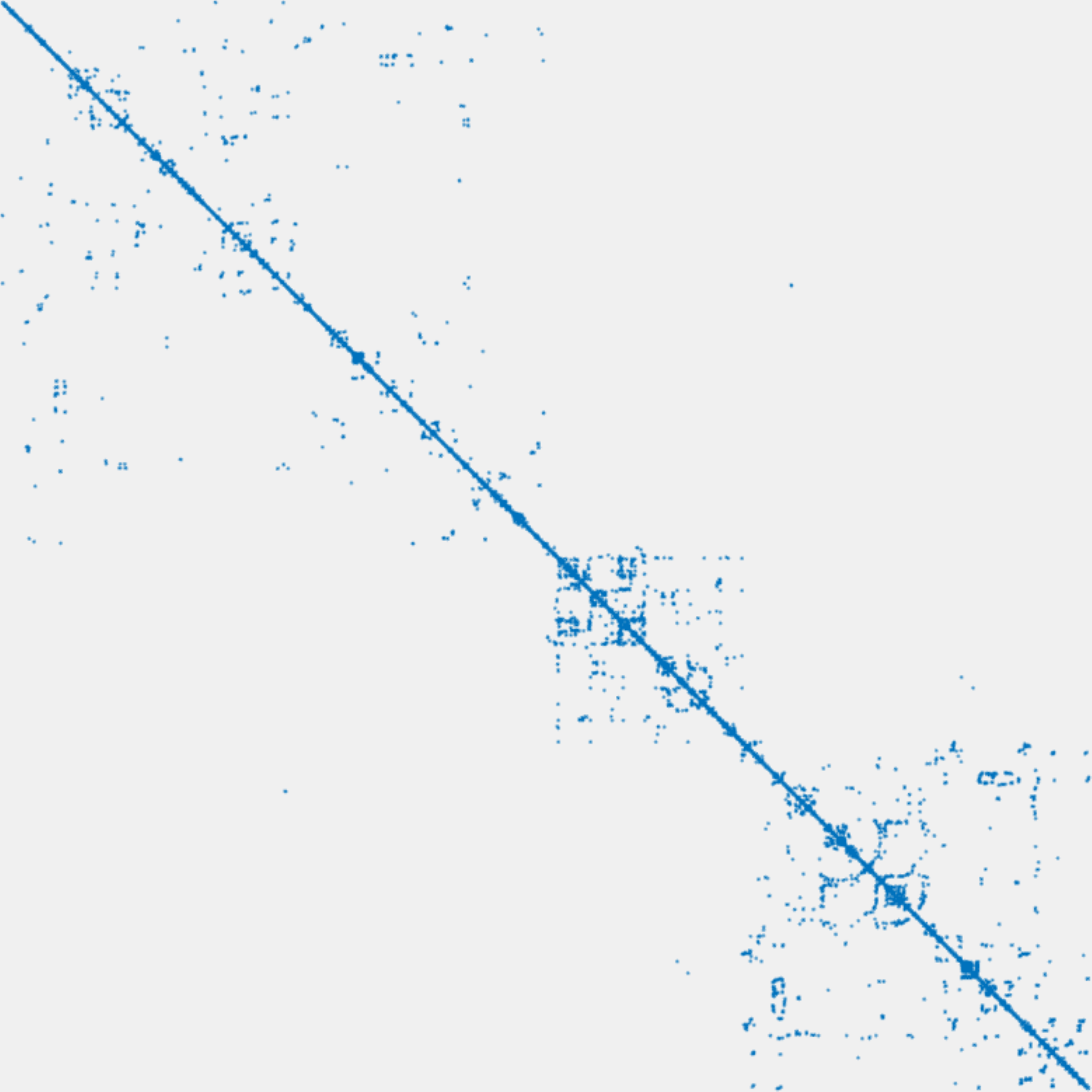}}
		\subcaptionbox{FLOW-CYTOMETRY}
		{\includegraphics[width=\spyfigwidth,height=\spyfigwidth]{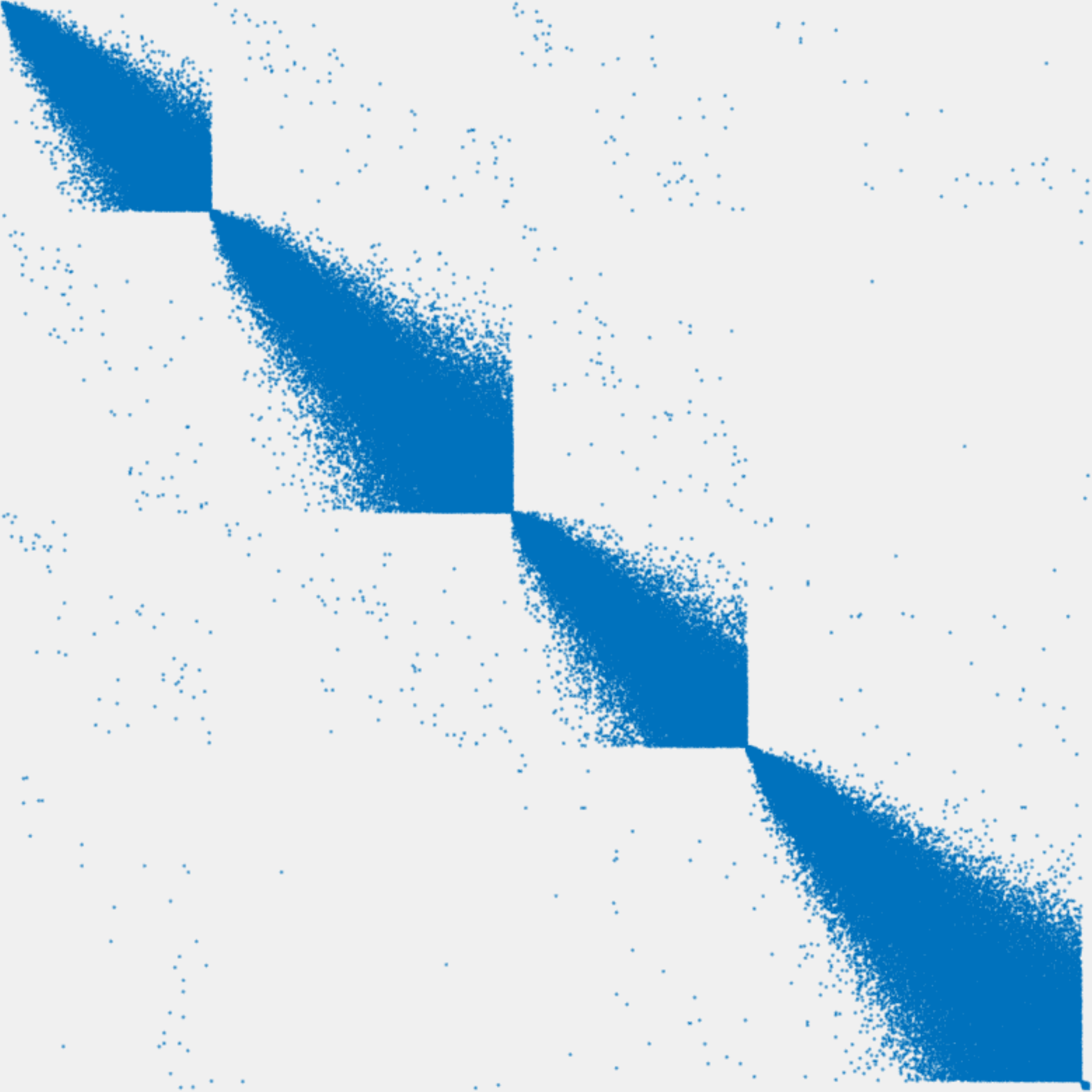}}
		\subcaptionbox{HIGGS}
		{\includegraphics[width=\spyfigwidth,height=\spyfigwidth]{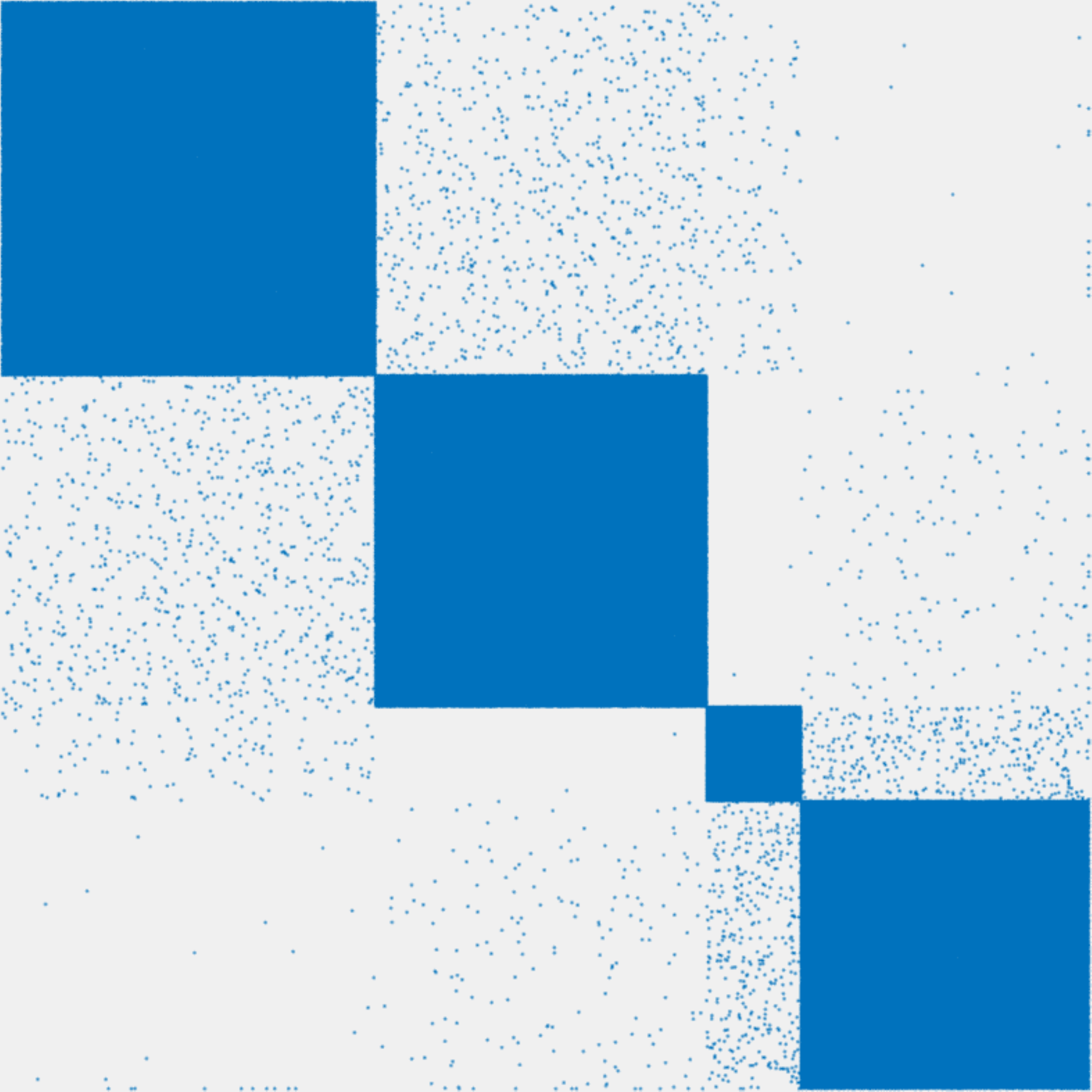}}
	\end{center}
	\caption{Visualization of the similarity matrix $P$ of the experimented data sets using Matlab \texttt{spy} function, where the rows and columns are sorted by the manual cluster labels. Blue dots show the 1's in the matrix and white dots show the 0's. Due to limited resolution, the figures shows a uniform subsample 10\% data points.}
	\label{fig:blockvis}
\end{figure}

\setlength{\fboxsep}{0pt}
\newcommand{\comparedfigwidth}{5.2cm}
\begin{figure}[t]
	\begin{center}
		\subcaptionbox{\texttt{SHUTTLE}}
		{\includegraphics[width=\comparedfigwidth,height=\comparedfigwidth]{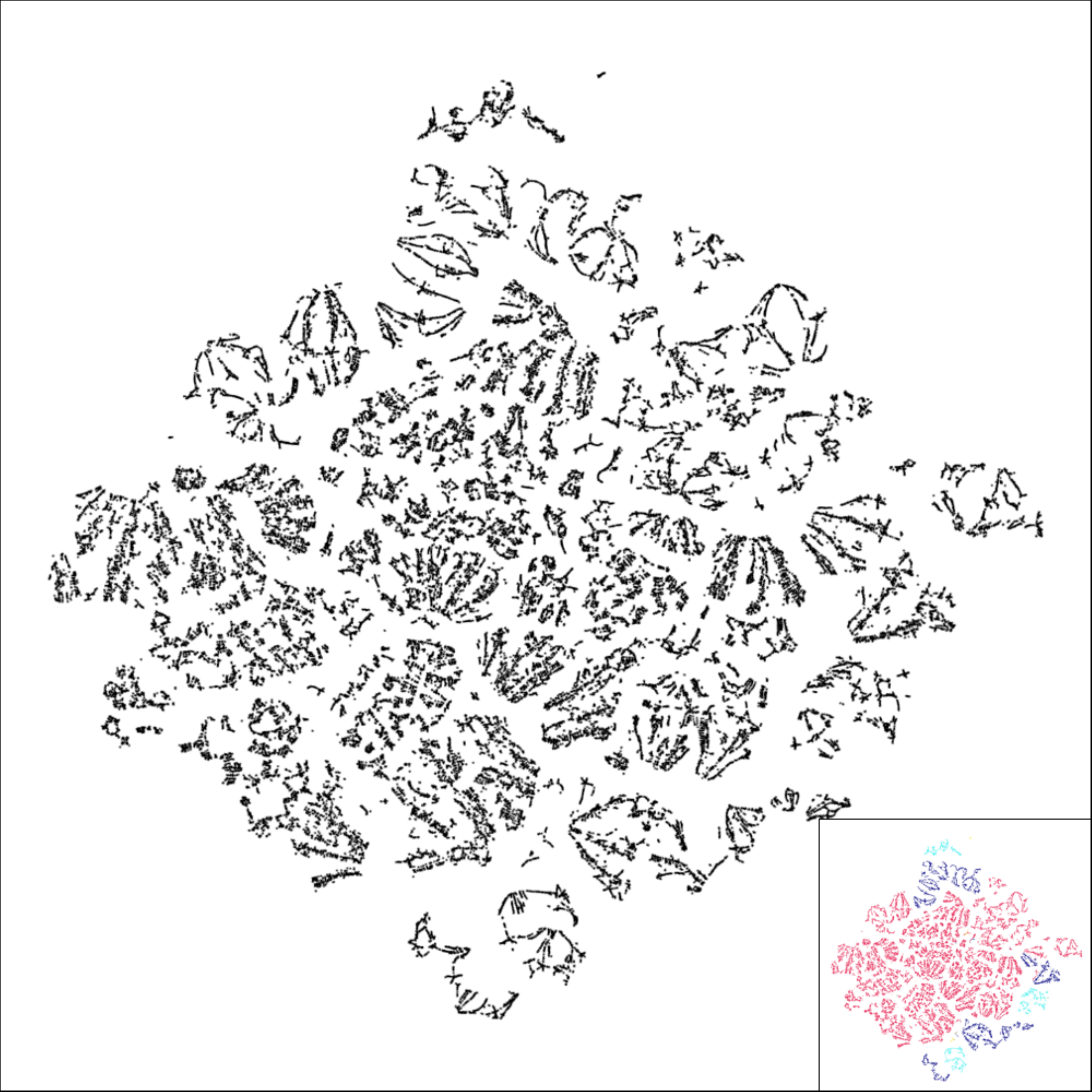}}
		\subcaptionbox{\texttt{IJCNN}}
		{\includegraphics[width=\comparedfigwidth,height=\comparedfigwidth]{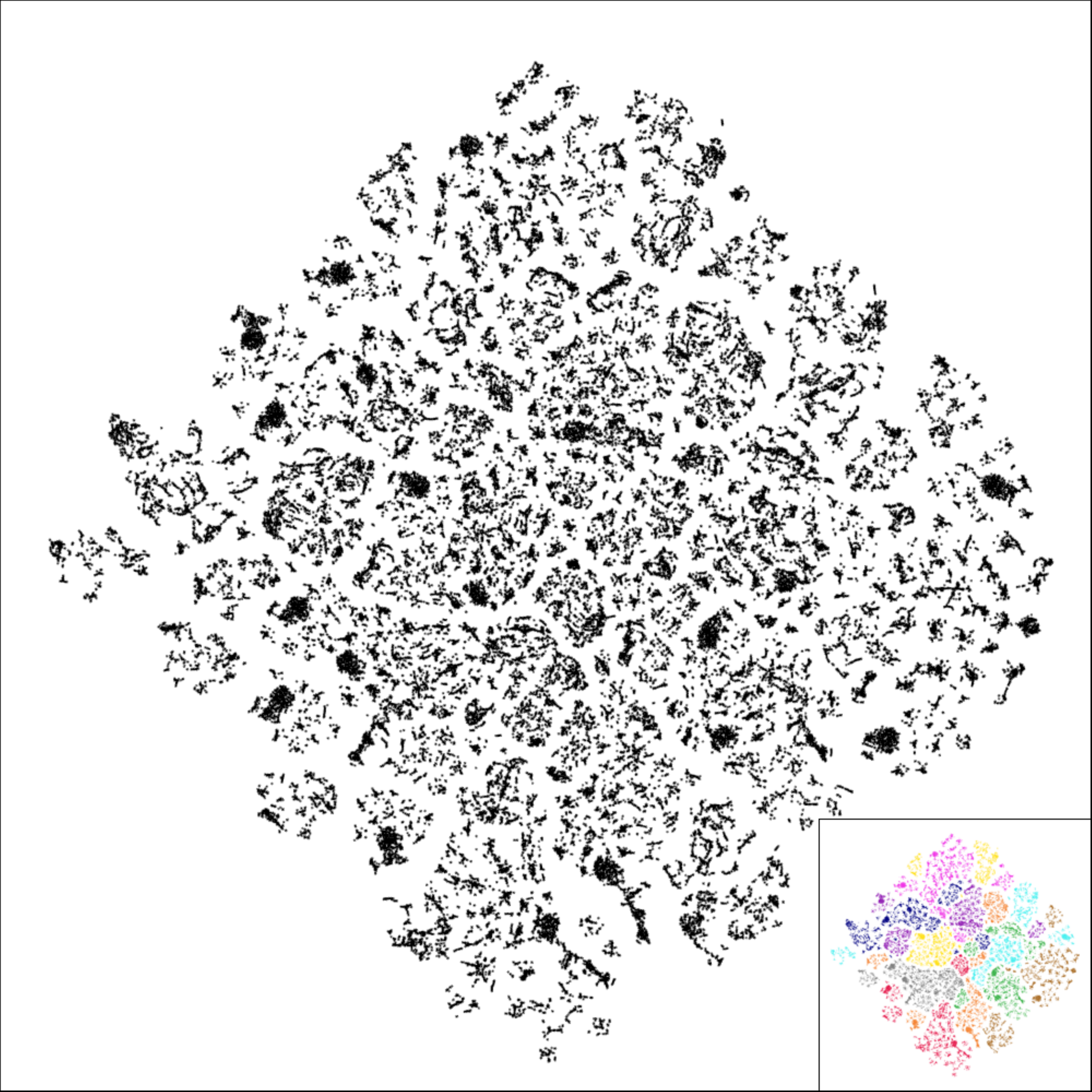}}
		
		\vspace{\baselineskip}
		
		\subcaptionbox{\texttt{TOMORADAR}}
		{\includegraphics[width=\comparedfigwidth,height=\comparedfigwidth]{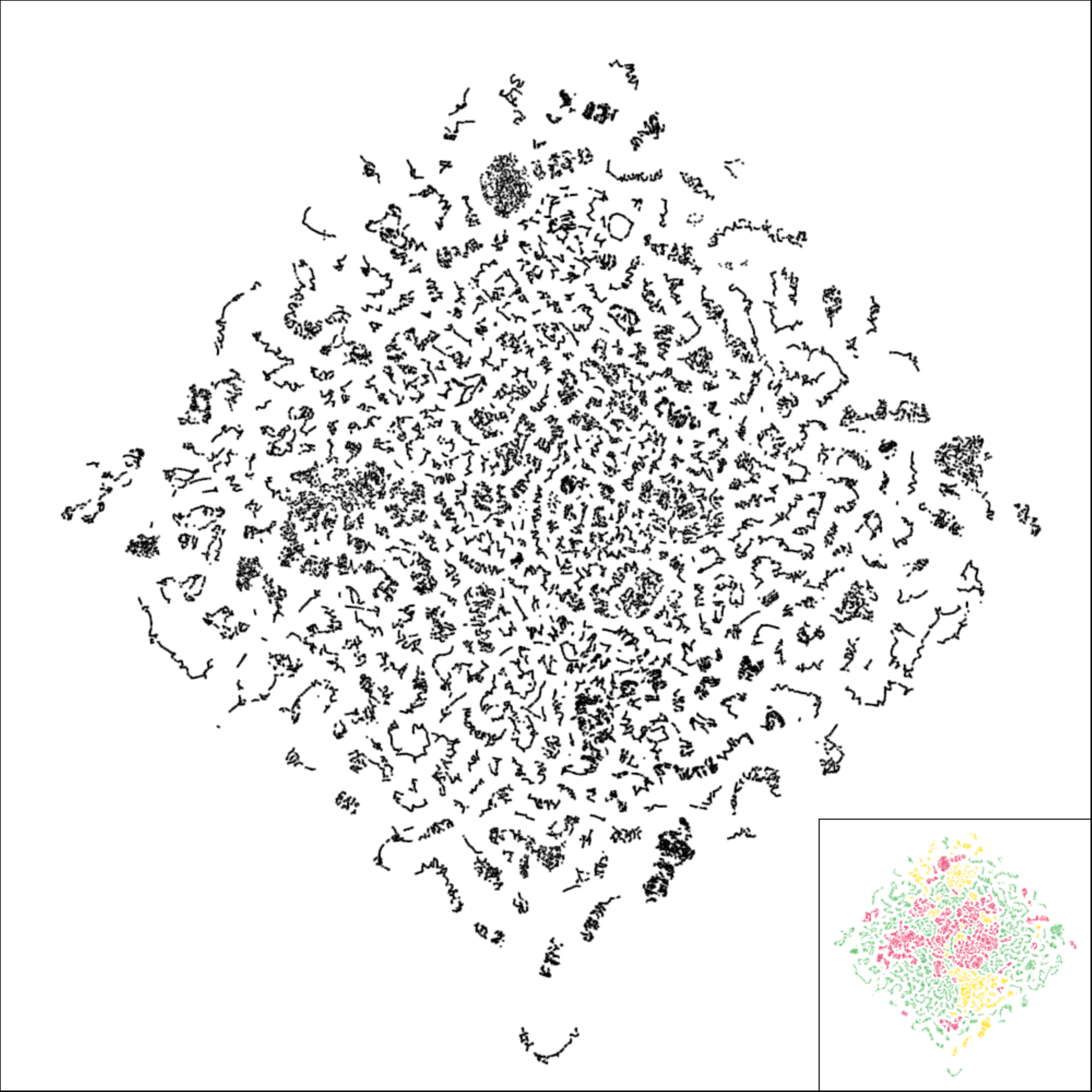}}
		\subcaptionbox{\texttt{FLOW-CYTOMETRY}}
		{\fbox{\includegraphics[width=\comparedfigwidth,height=\comparedfigwidth]{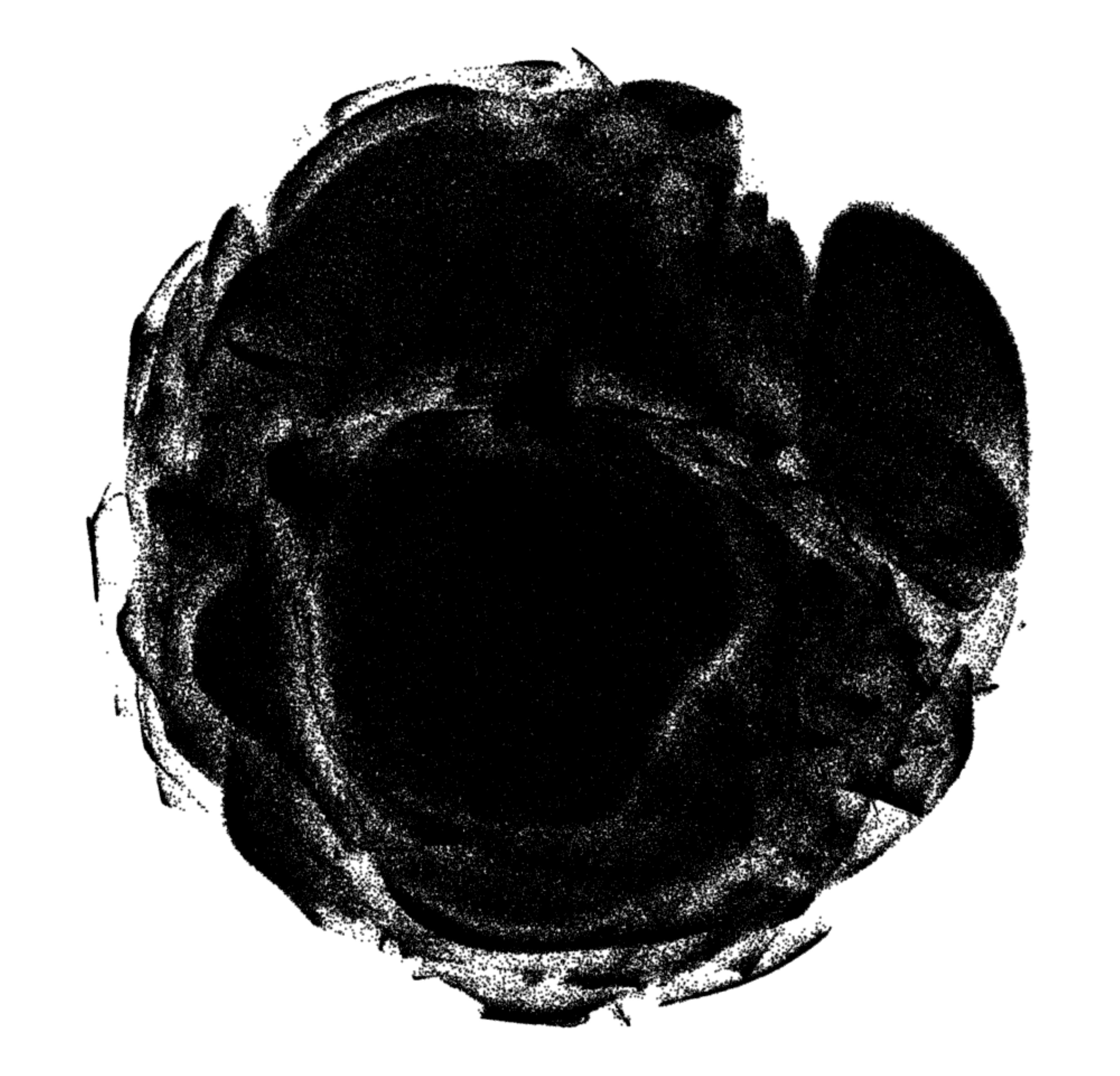}}}
		\subcaptionbox{\texttt{HIGGS}}
		{\fbox{\includegraphics[width=\comparedfigwidth,height=\comparedfigwidth]{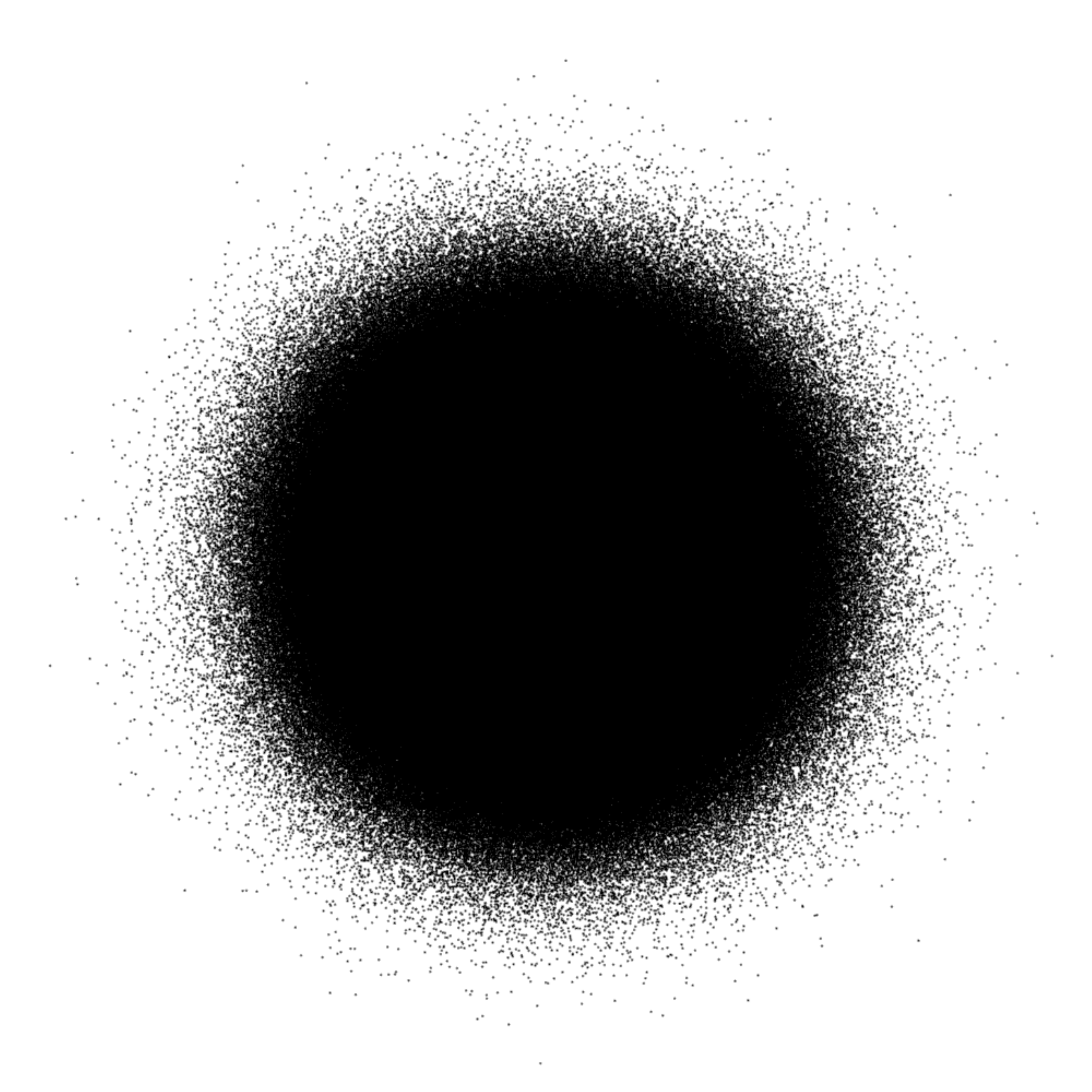}}}
	\end{center}
	\caption{Visualizations of the data sets by using t-SNE. If there exist class labels, they are shown by colors in the small sub-figures.}
	\label{fig:tsnevis}
\end{figure}

\newcommand{\perpfigwidth}{5.2cm}
\begin{figure}[p]
	\centering
	\subcaptionbox{perplexity=10}{\fbox{\includegraphics[width=\perpfigwidth,height=\perpfigwidth]{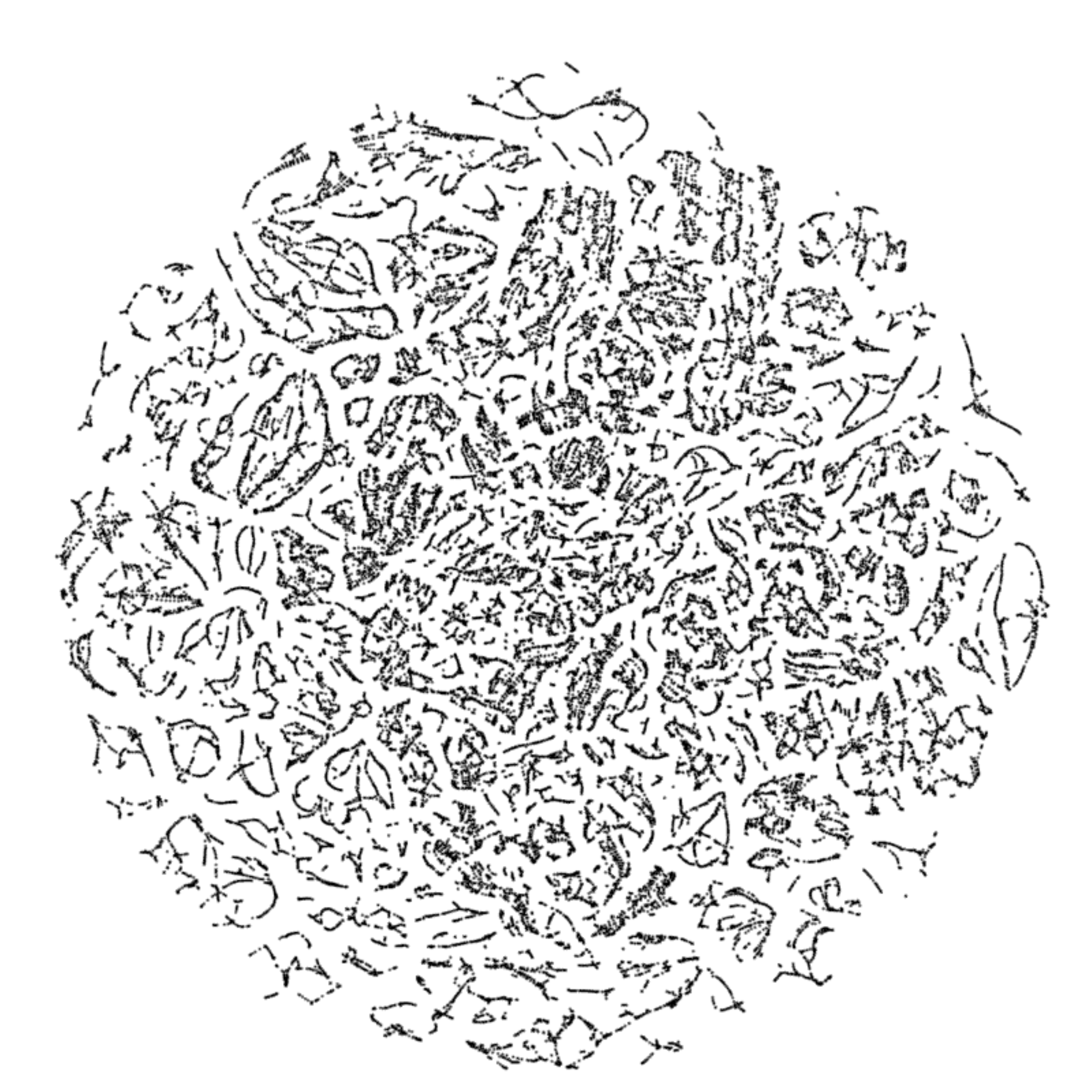}}}
	\subcaptionbox{perplexity=20}{\fbox{\includegraphics[width=\perpfigwidth,height=\perpfigwidth]{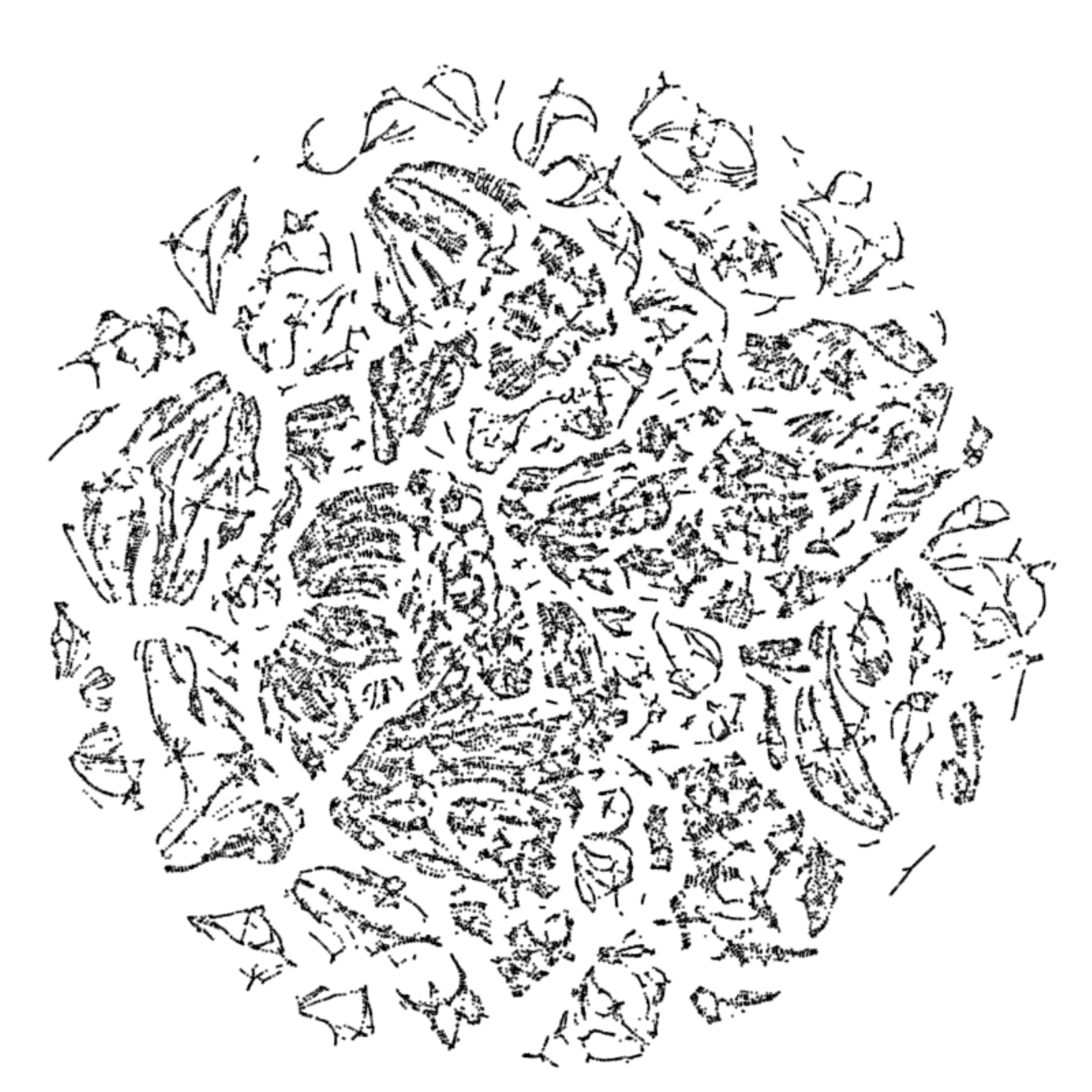}}}
	\subcaptionbox{perplexity=30}{\fbox{\includegraphics[width=\perpfigwidth,height=\perpfigwidth]{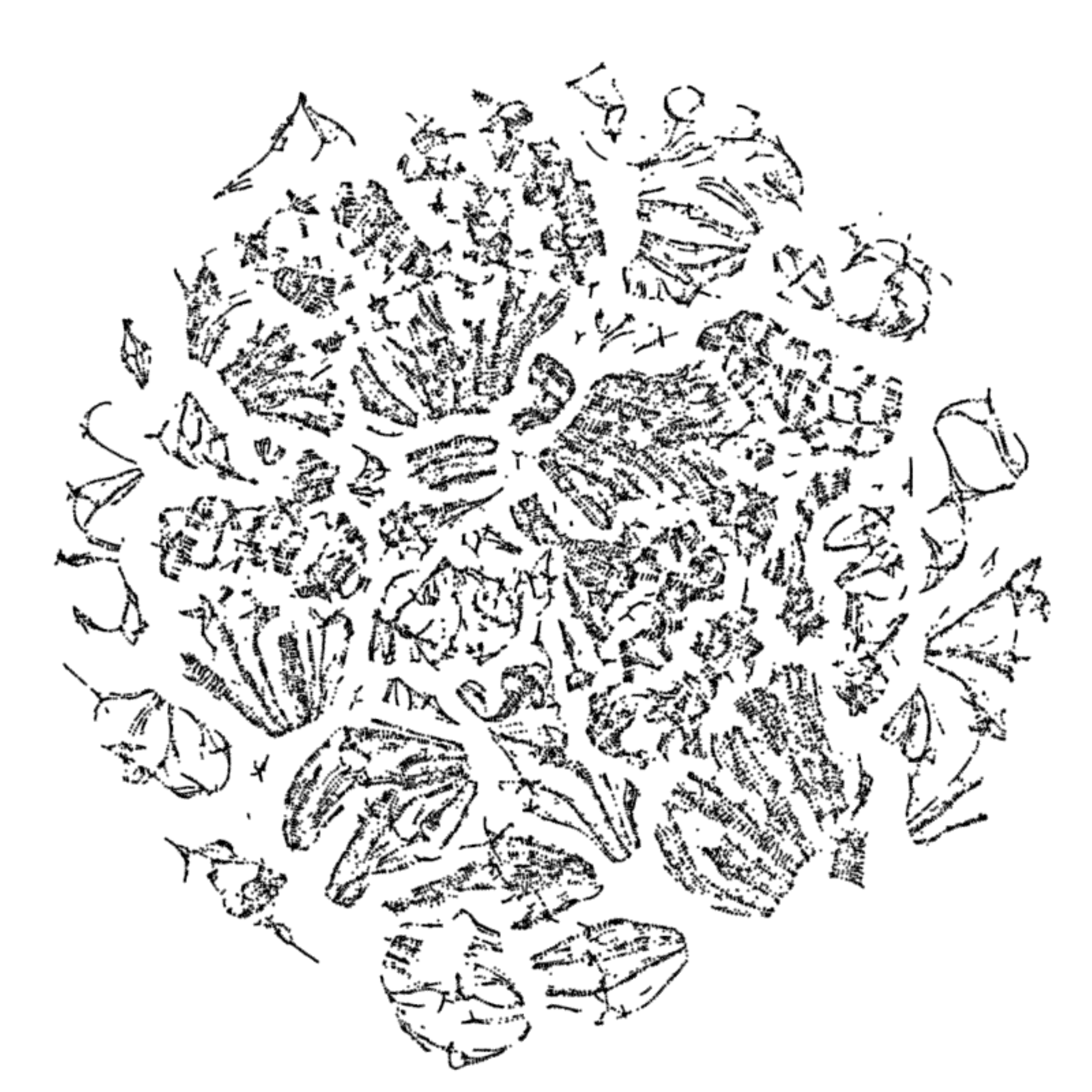}}}
	
	\vspace{\baselineskip}
	
	\subcaptionbox{perplexity=50}{\fbox{\includegraphics[width=\perpfigwidth,height=\perpfigwidth]{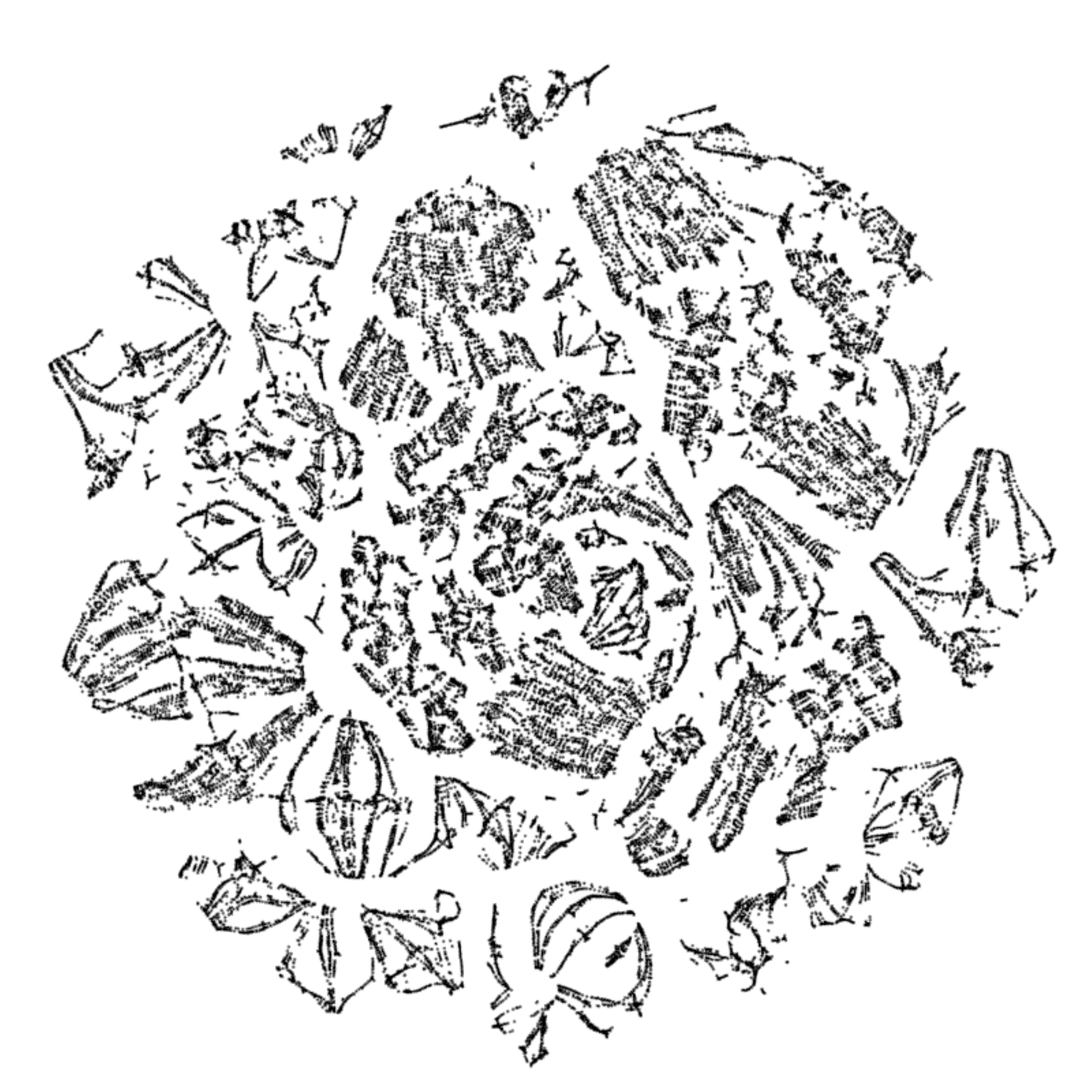}}}
	\subcaptionbox{perplexity=70}{\fbox{\includegraphics[width=\perpfigwidth,height=\perpfigwidth]{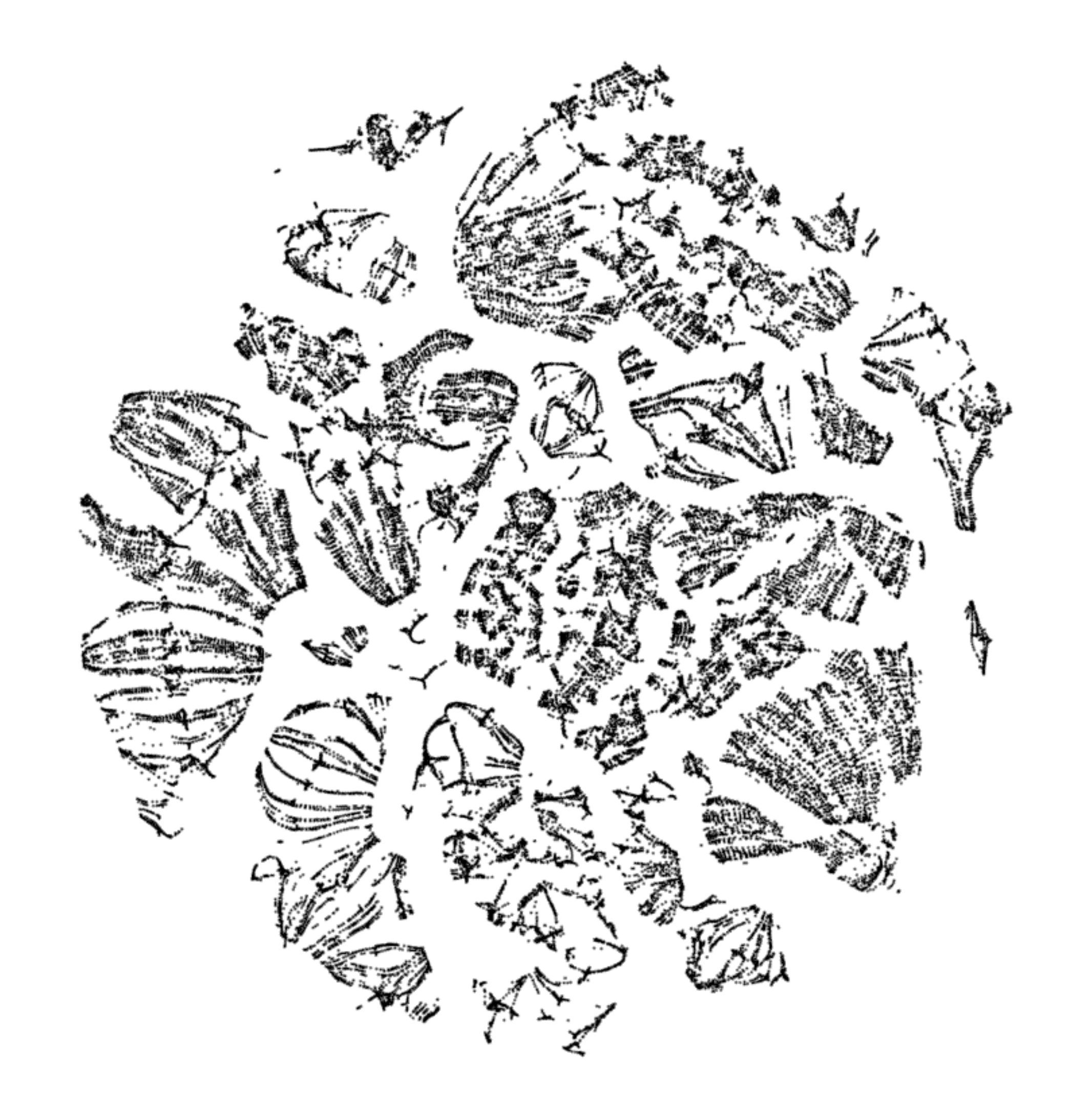}}}
	\subcaptionbox{perplexity=90}{\fbox{\includegraphics[width=\perpfigwidth,height=\perpfigwidth]{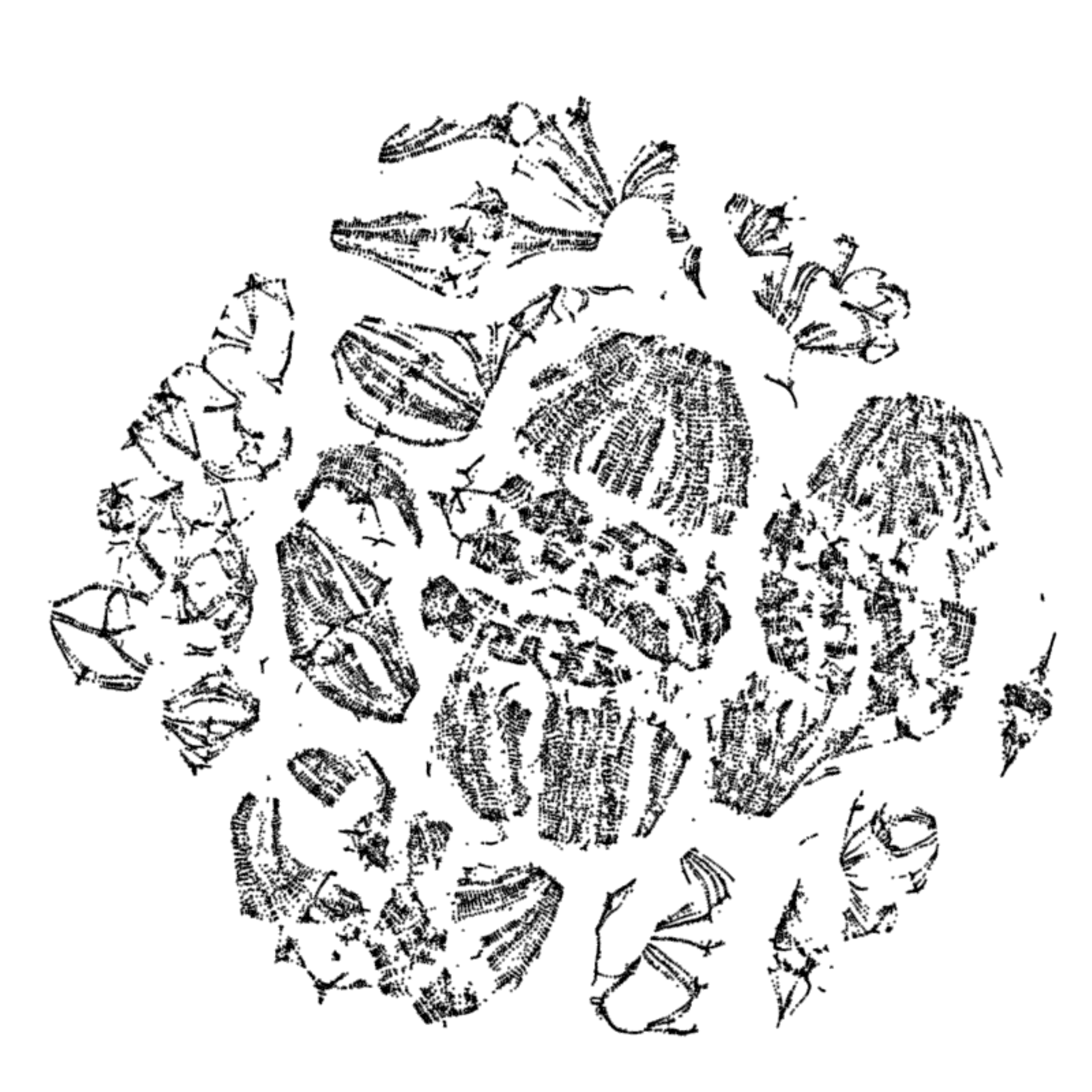}}}
	
	\vspace{\baselineskip}
	
	\subcaptionbox{perplexity=110}{\fbox{\includegraphics[width=\perpfigwidth,height=\perpfigwidth]{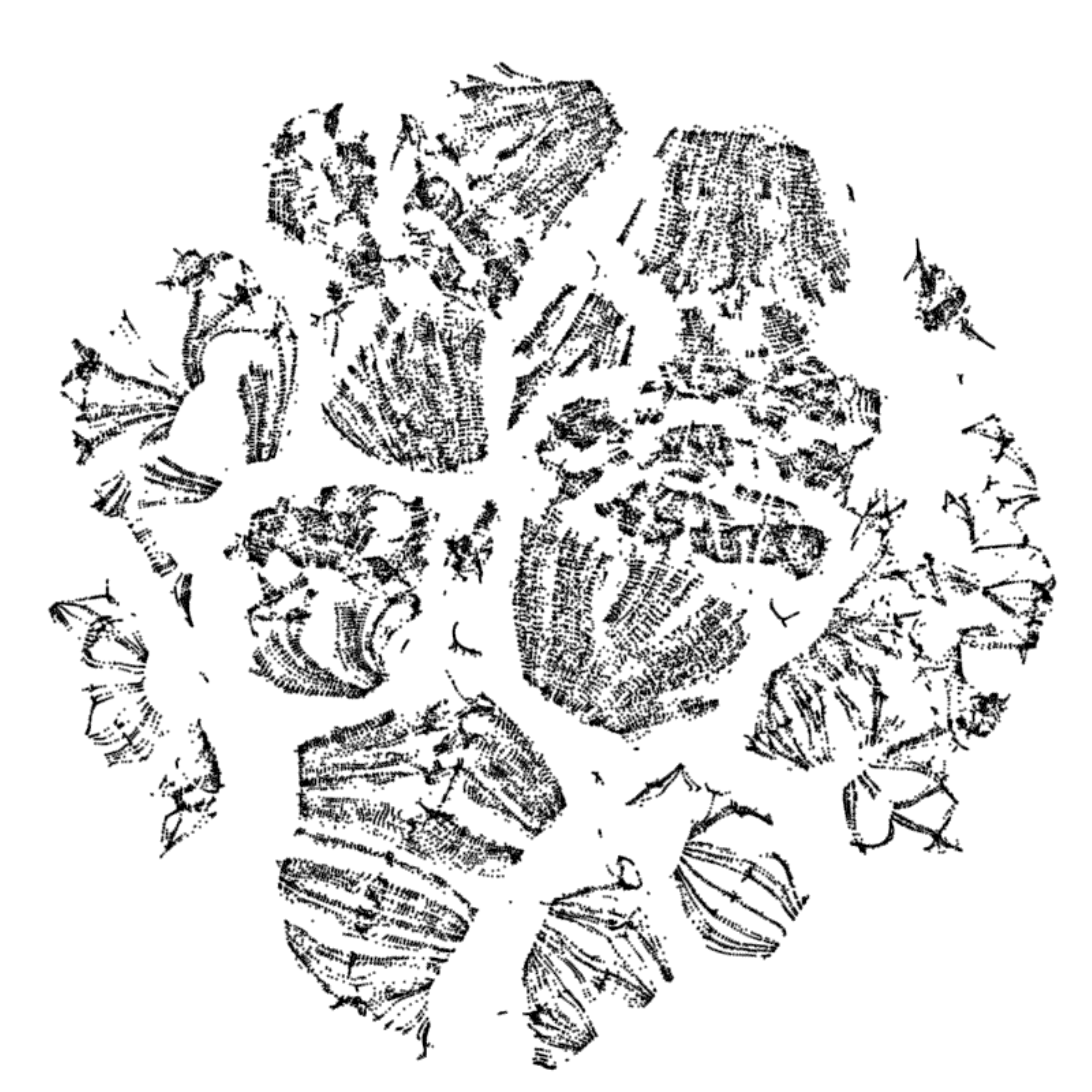}}}
	\subcaptionbox{perplexity=130}{\fbox{\includegraphics[width=\perpfigwidth,height=\perpfigwidth]{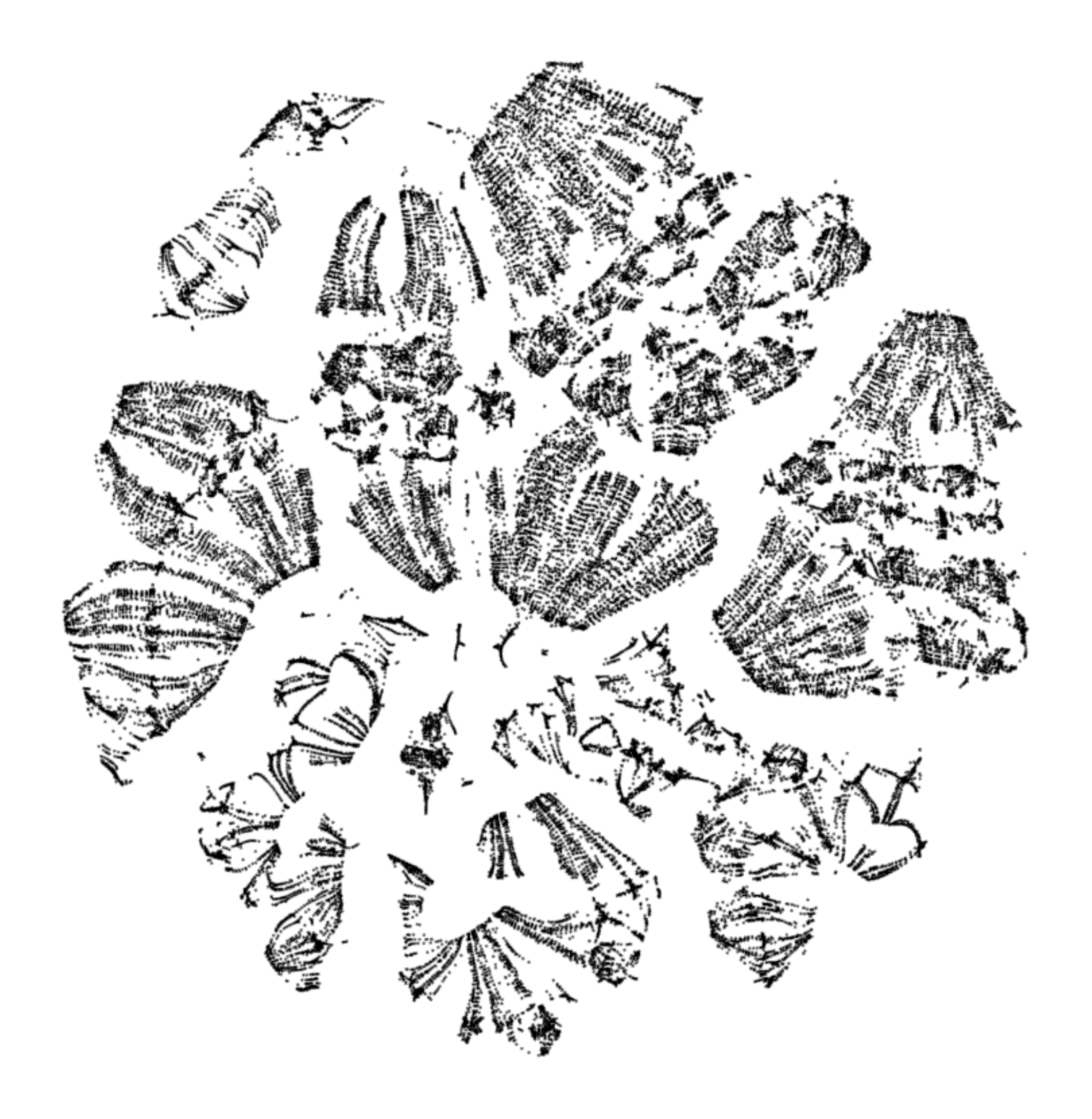}}}
	\subcaptionbox{perplexity=150}{\fbox{\includegraphics[width=\perpfigwidth,height=\perpfigwidth]{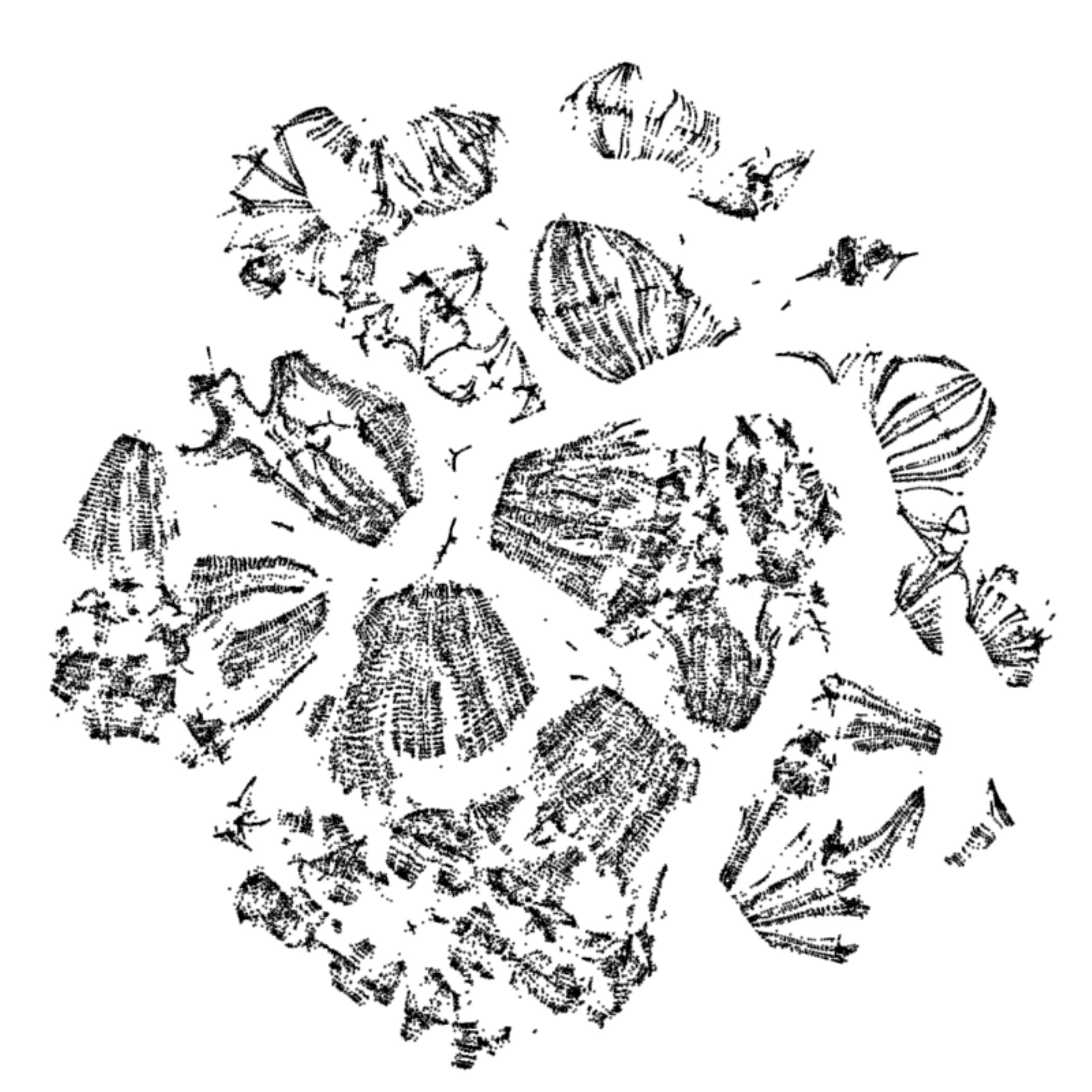}}}
	\caption{Visualizations of the \texttt{SHUTTLE} data set using t-SNE with various perplexities.}
	\label{fig:perpsshuttle}
\end{figure}

\begin{figure}[p]
	\centering
	\subcaptionbox{perplexity=10}{\fbox{\includegraphics[width=\perpfigwidth,height=\perpfigwidth]{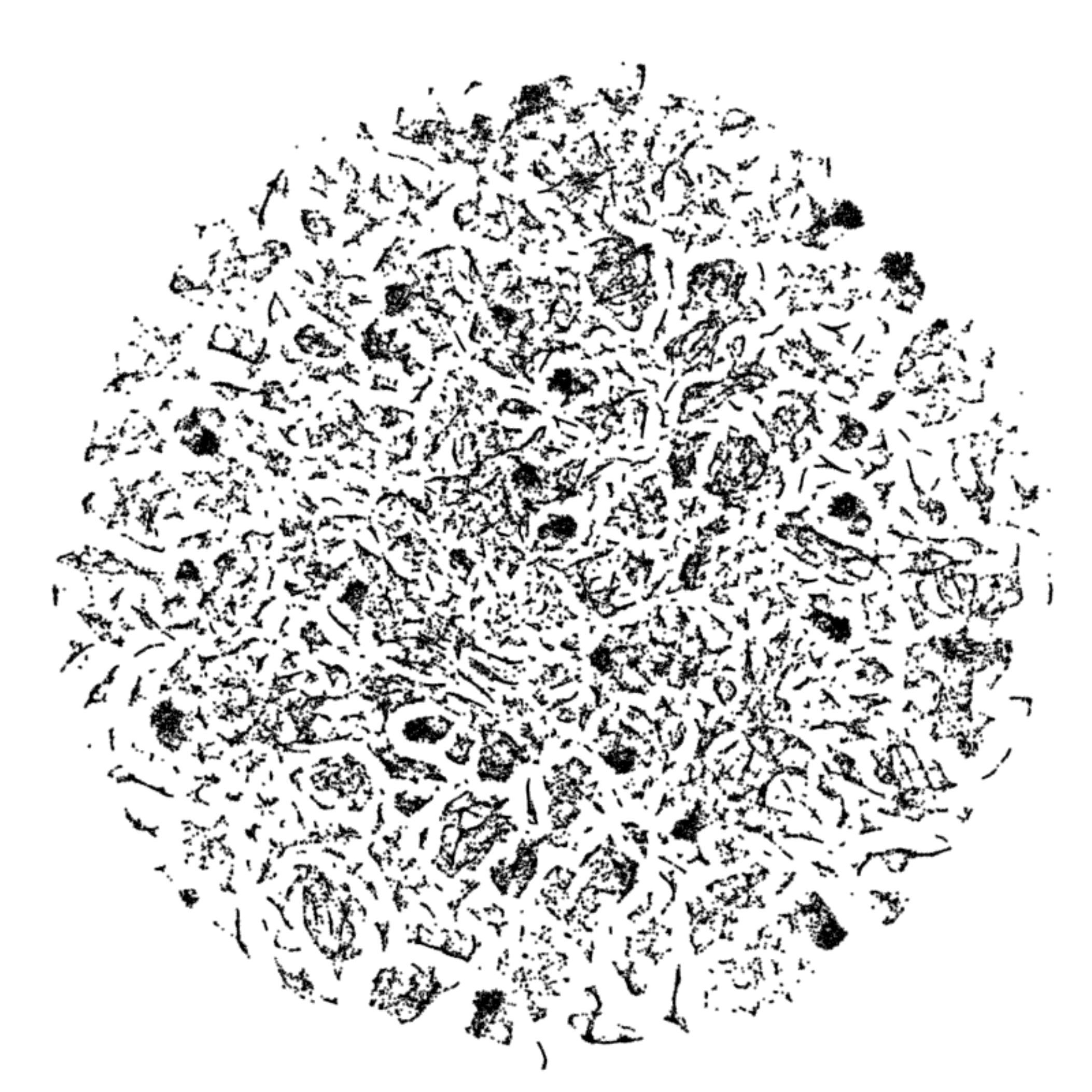}}}
	\subcaptionbox{perplexity=20}{\fbox{\includegraphics[width=\perpfigwidth,height=\perpfigwidth]{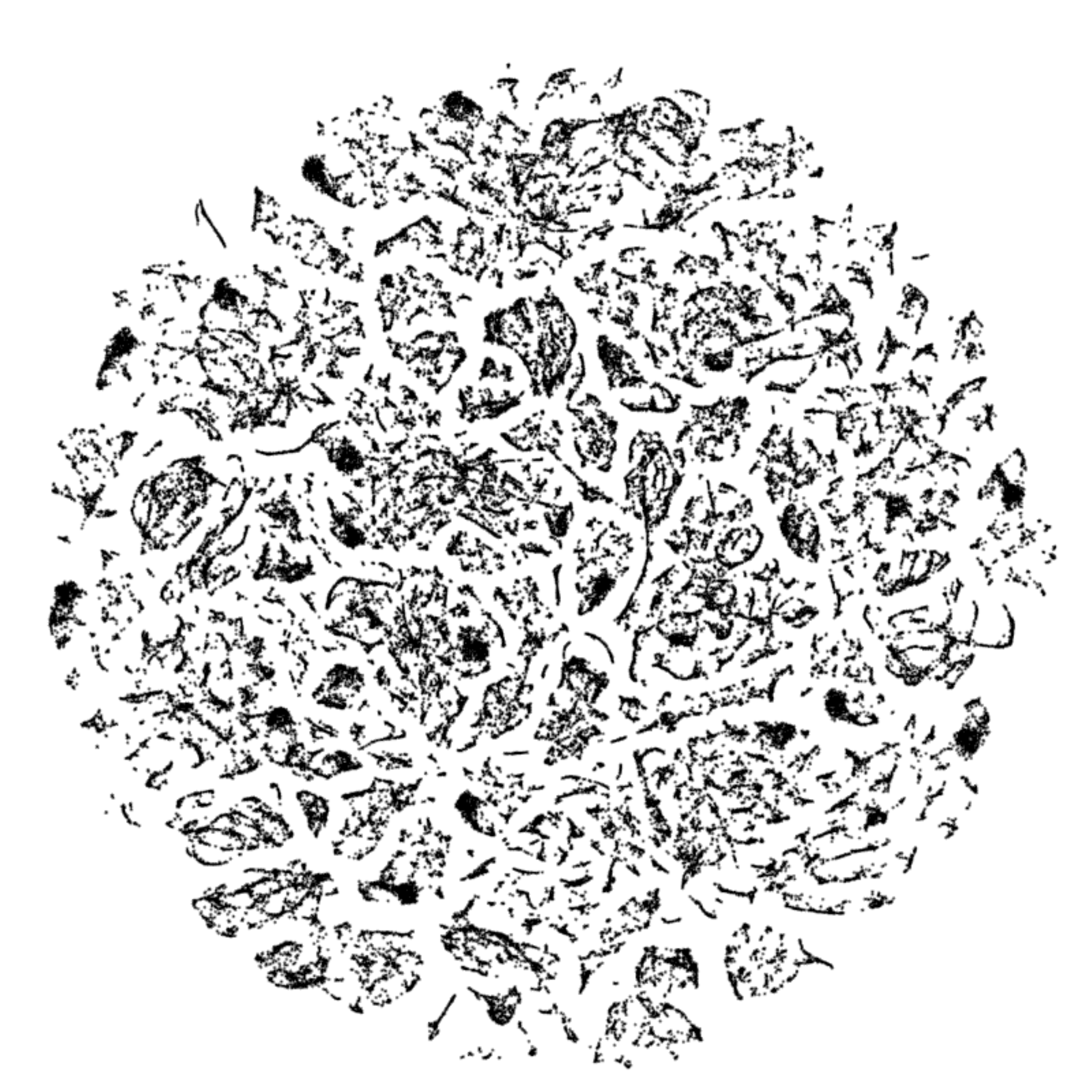}}}
	\subcaptionbox{perplexity=30}{\fbox{\includegraphics[width=\perpfigwidth,height=\perpfigwidth]{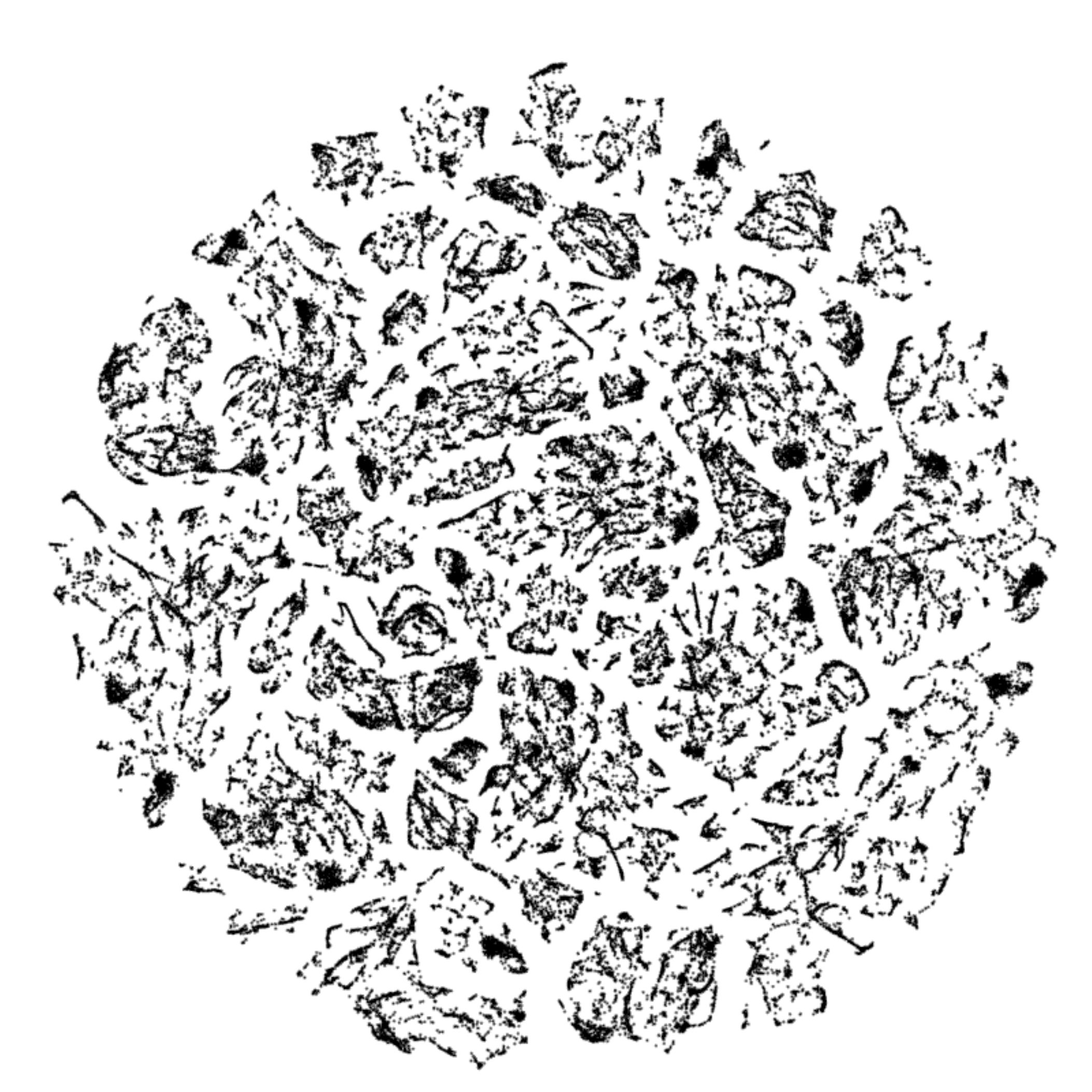}}}
	
	\vspace{\baselineskip}
	
	\subcaptionbox{perplexity=50}{\fbox{\includegraphics[width=\perpfigwidth,height=\perpfigwidth]{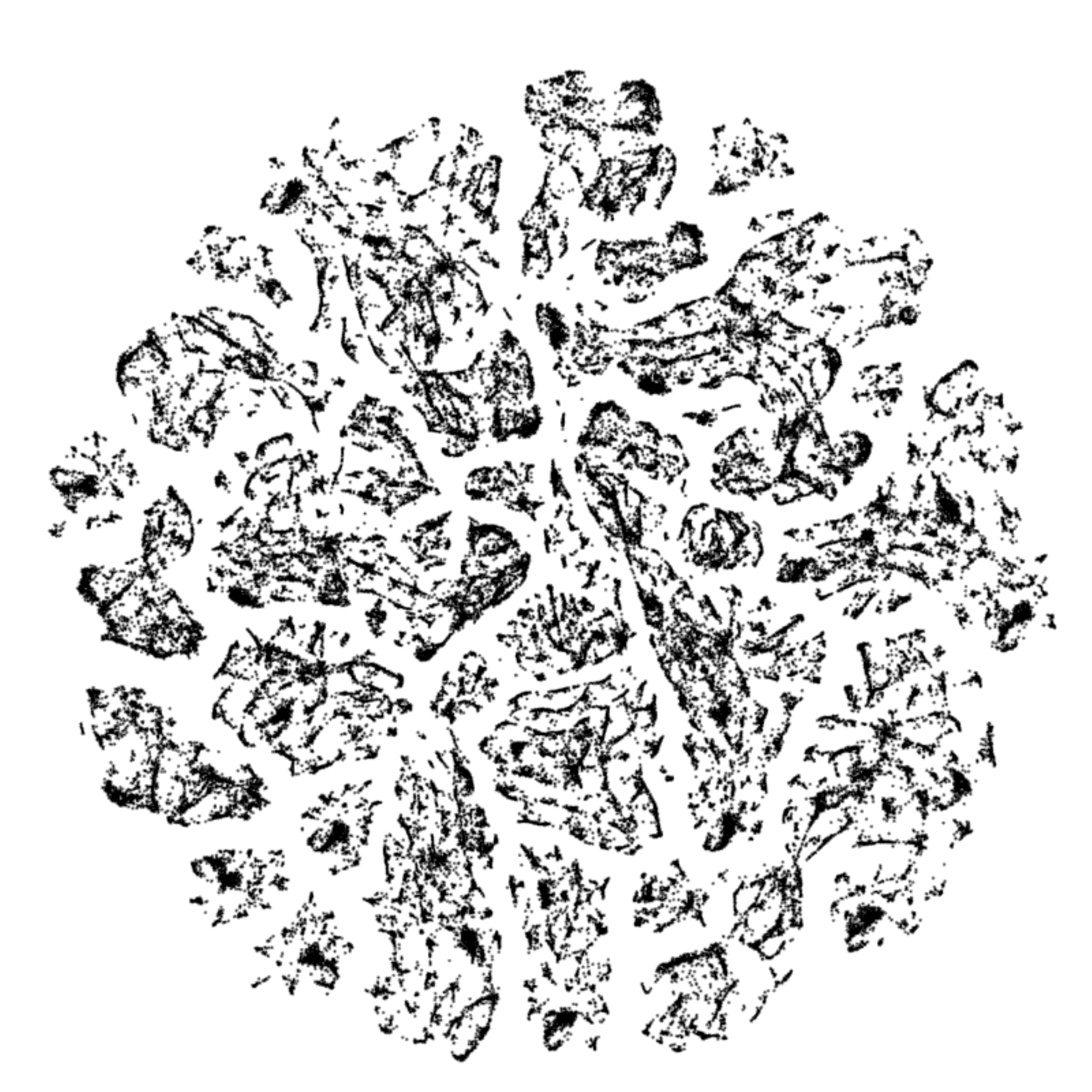}}}
	\subcaptionbox{perplexity=70}{\fbox{\includegraphics[width=\perpfigwidth,height=\perpfigwidth]{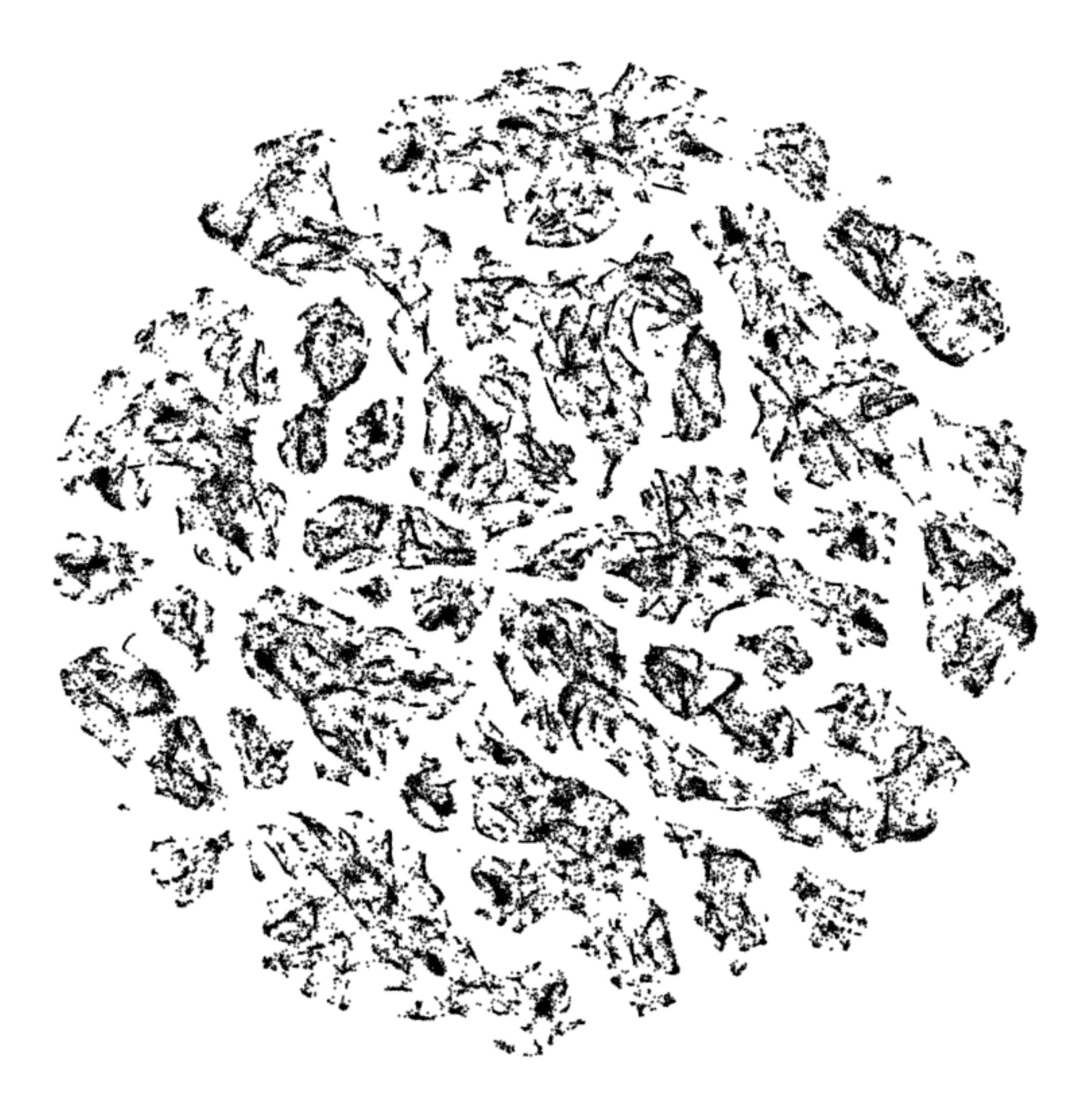}}}
	\subcaptionbox{perplexity=90}{\fbox{\includegraphics[width=\perpfigwidth,height=\perpfigwidth]{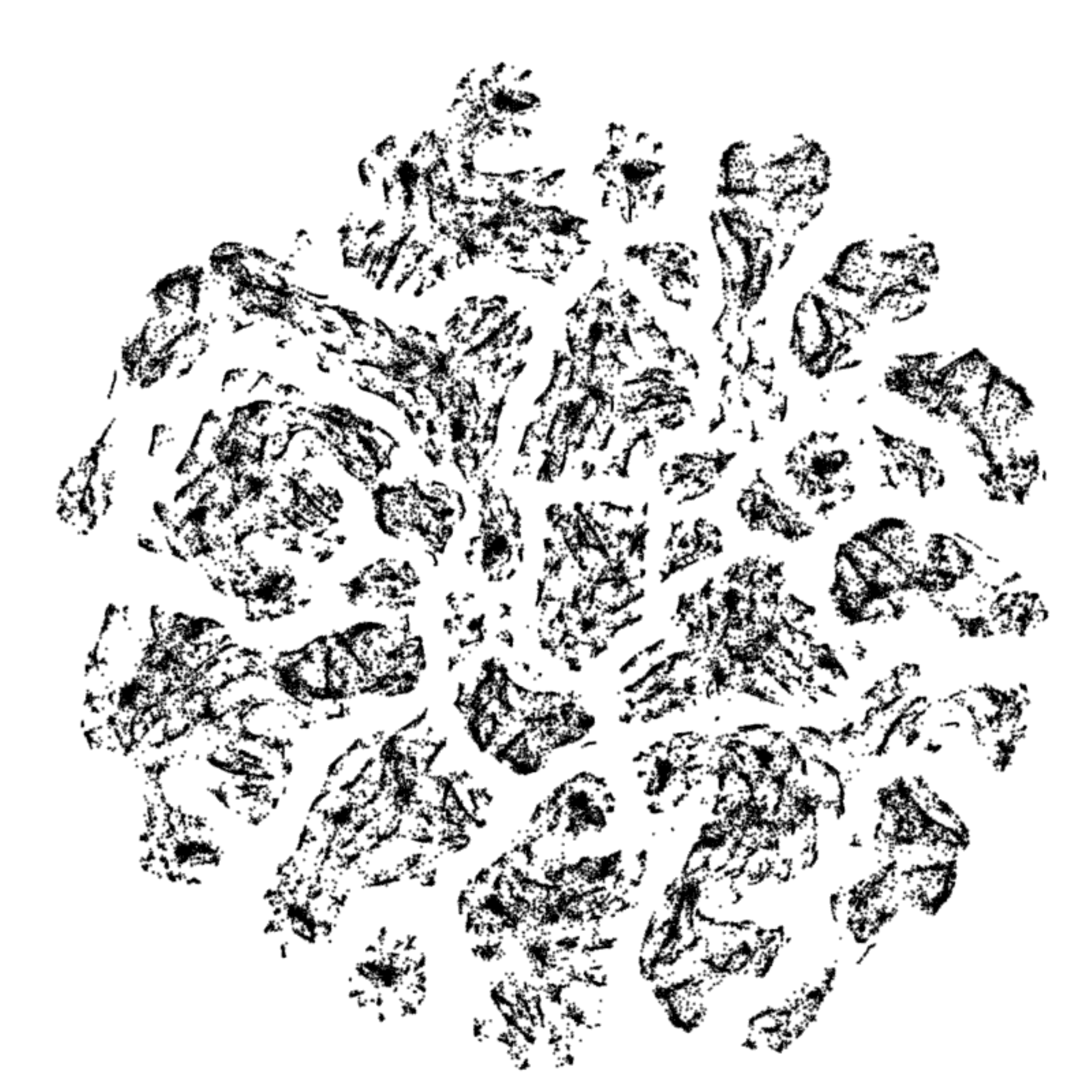}}}
	
	\vspace{\baselineskip}
	
	\subcaptionbox{perplexity=110}{\fbox{\includegraphics[width=\perpfigwidth,height=\perpfigwidth]{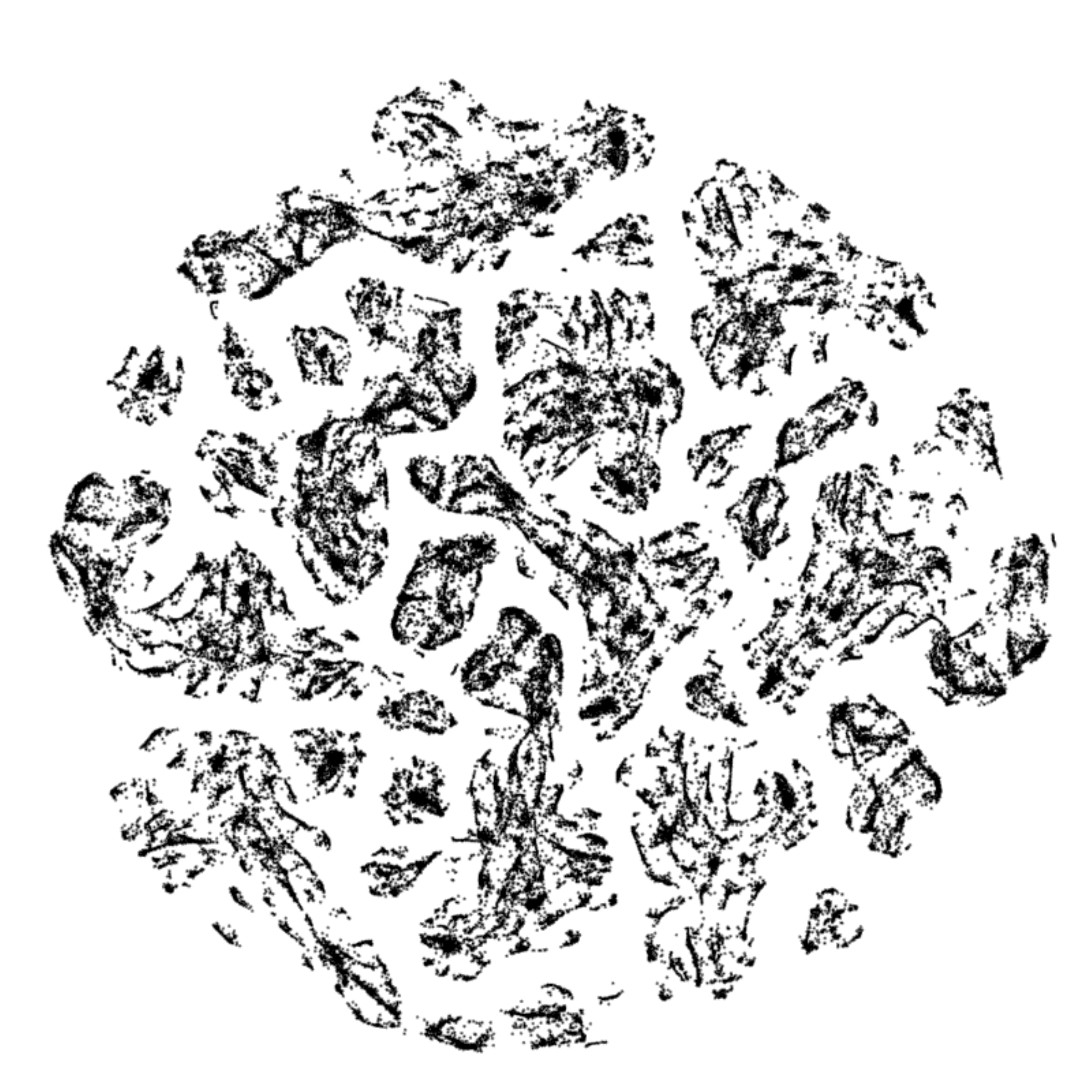}}}
	\subcaptionbox{perplexity=130}{\fbox{\includegraphics[width=\perpfigwidth,height=\perpfigwidth]{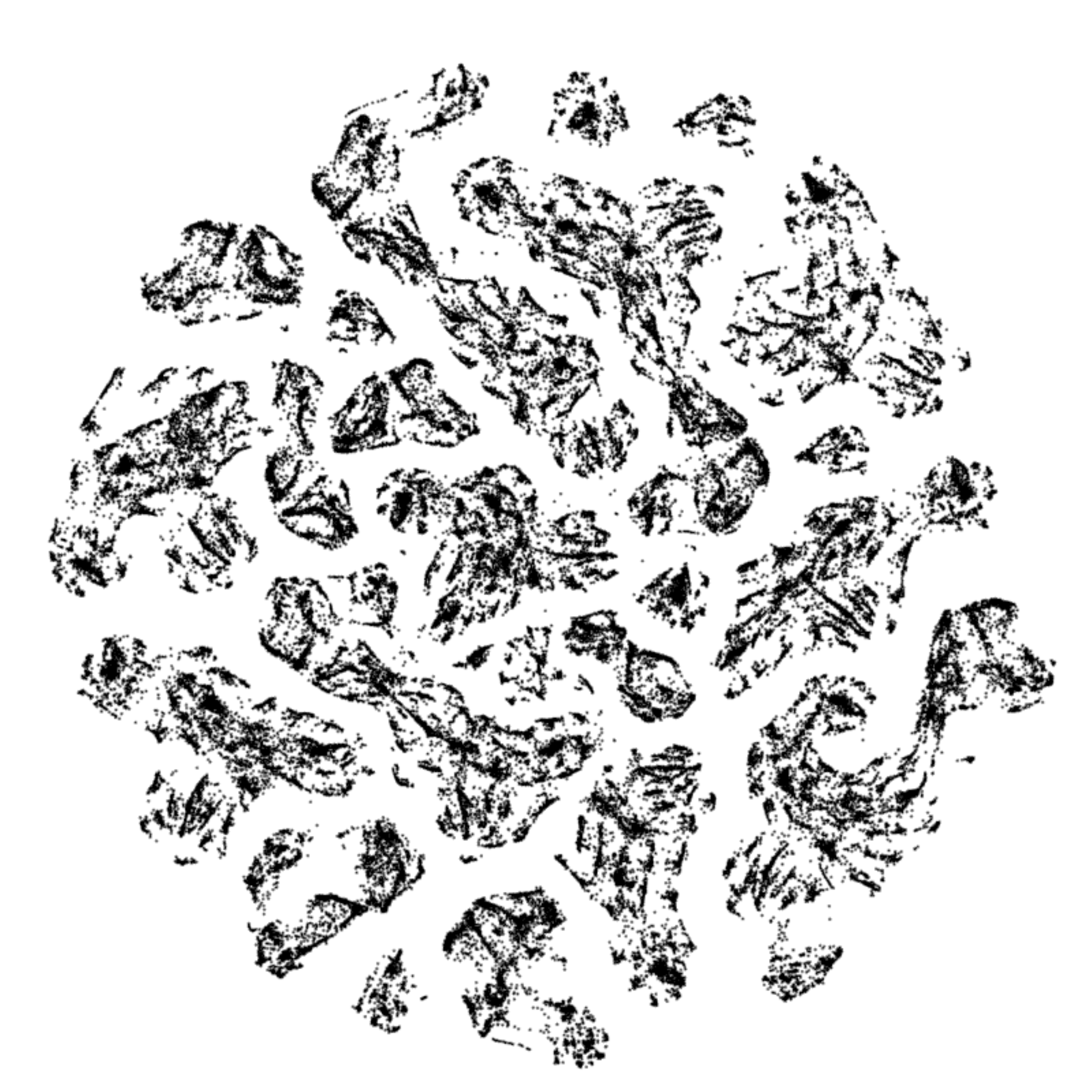}}}
	\subcaptionbox{perplexity=150}{\fbox{\includegraphics[width=\perpfigwidth,height=\perpfigwidth]{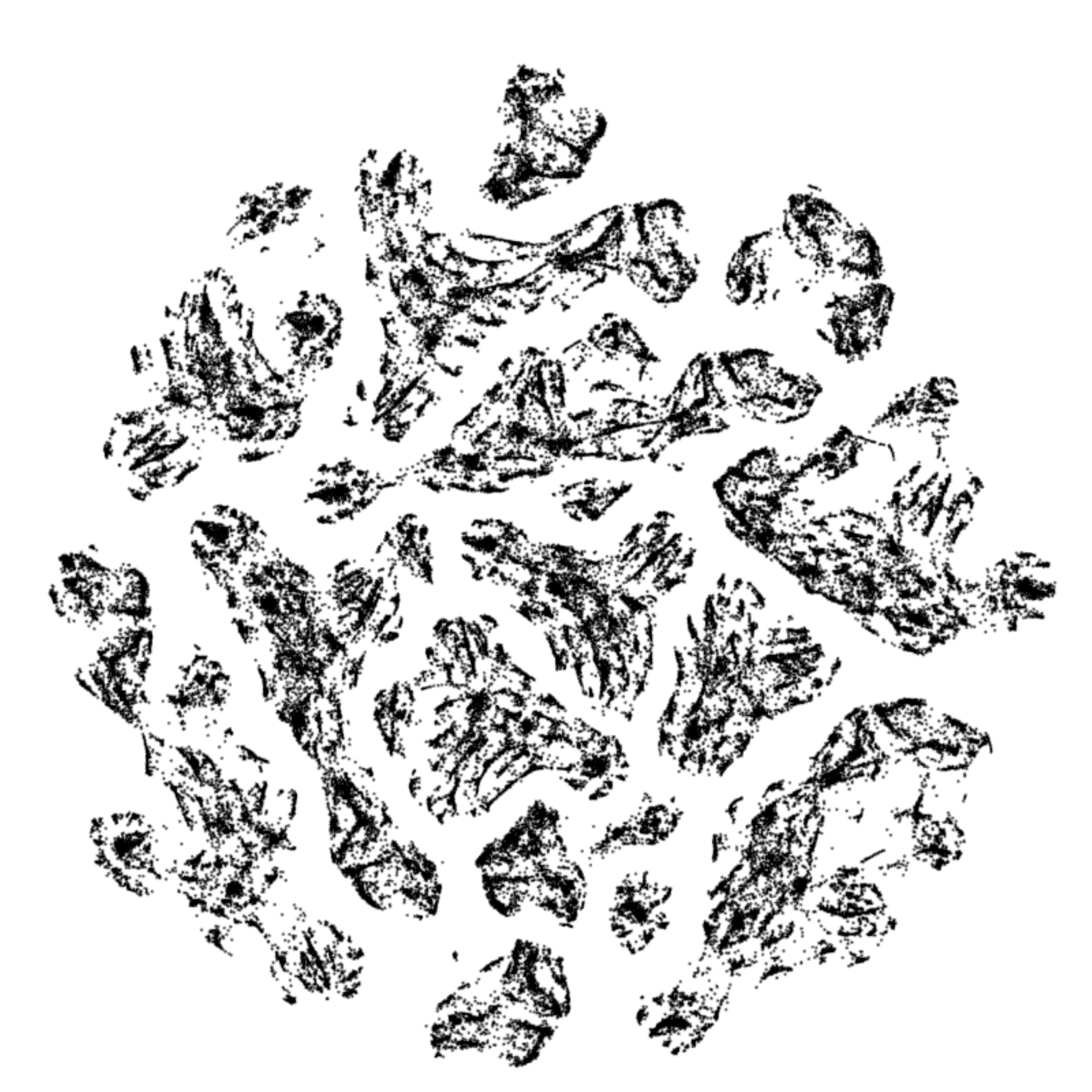}}}
	\caption{Visualizations of the \texttt{IJCNN} data set using t-SNE with various perplexities.}
	\label{fig:perpsijcnn}
\end{figure}

\begin{figure}[p]
	\centering
	\subcaptionbox{perplexity=10}{\fbox{\includegraphics[width=\perpfigwidth,height=\perpfigwidth]{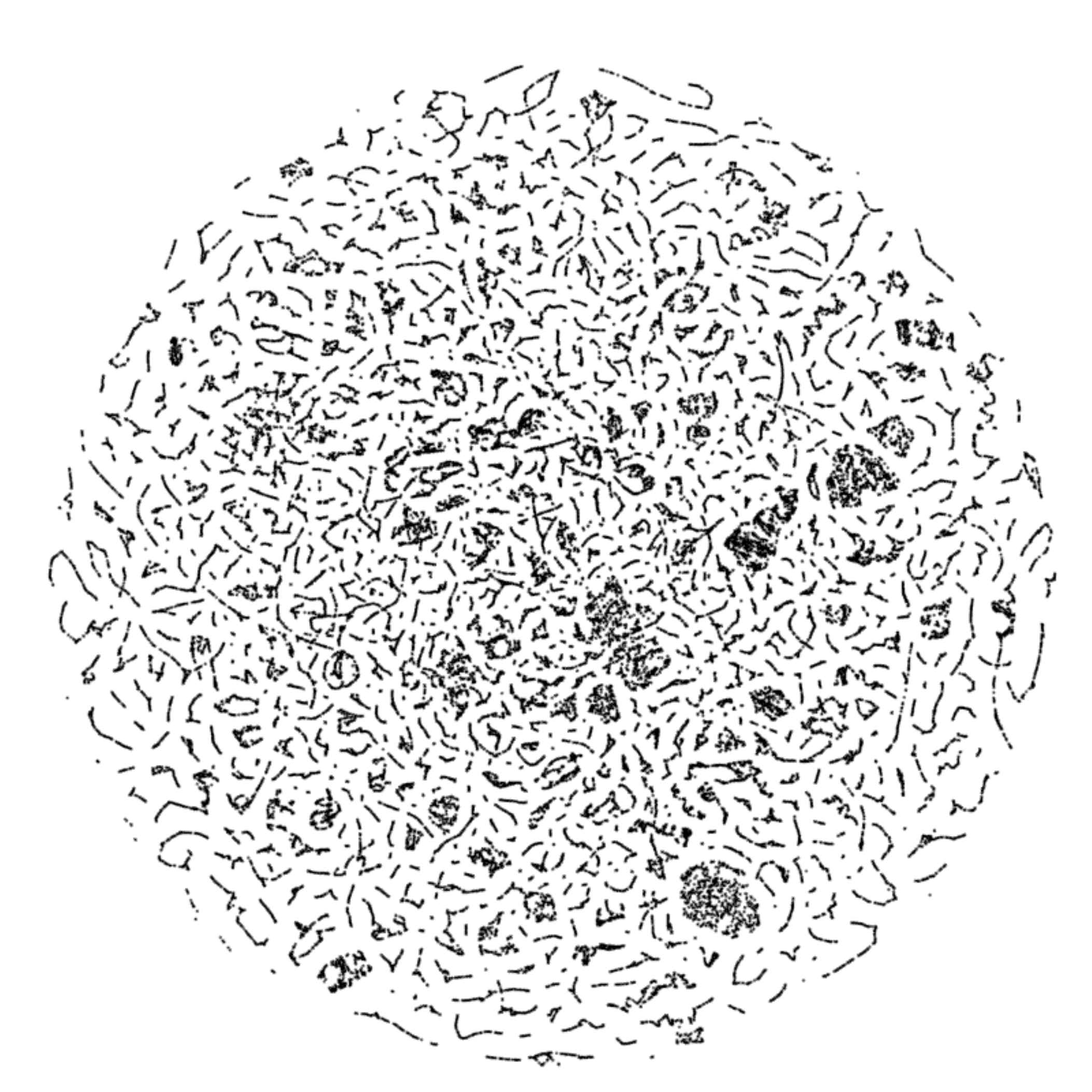}}}
	\subcaptionbox{perplexity=20}{\fbox{\includegraphics[width=\perpfigwidth,height=\perpfigwidth]{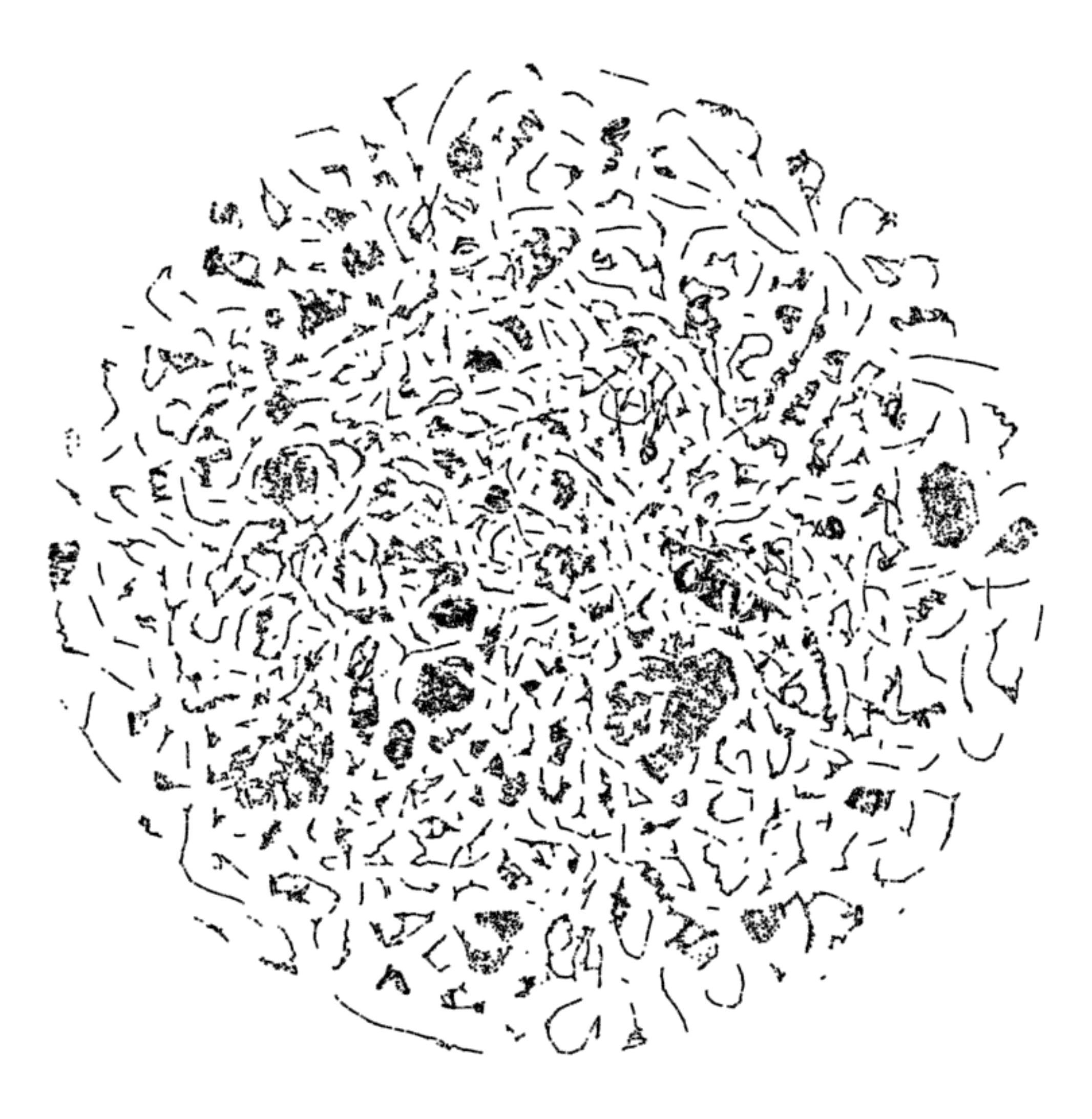}}}
	\subcaptionbox{perplexity=30}{\fbox{\includegraphics[width=\perpfigwidth,height=\perpfigwidth]{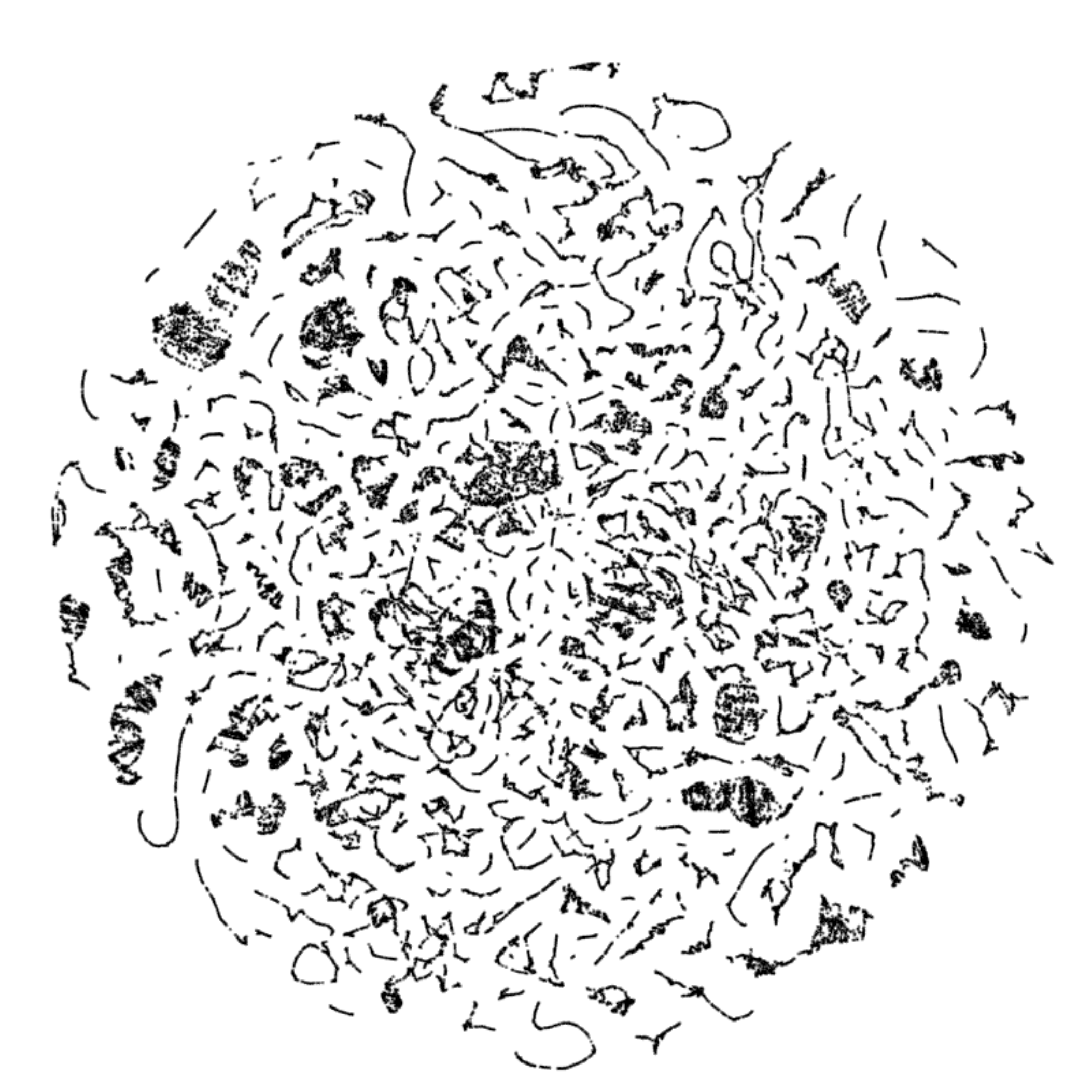}}}
	
	\vspace{\baselineskip}
	
	\subcaptionbox{perplexity=50}{\fbox{\includegraphics[width=\perpfigwidth,height=\perpfigwidth]{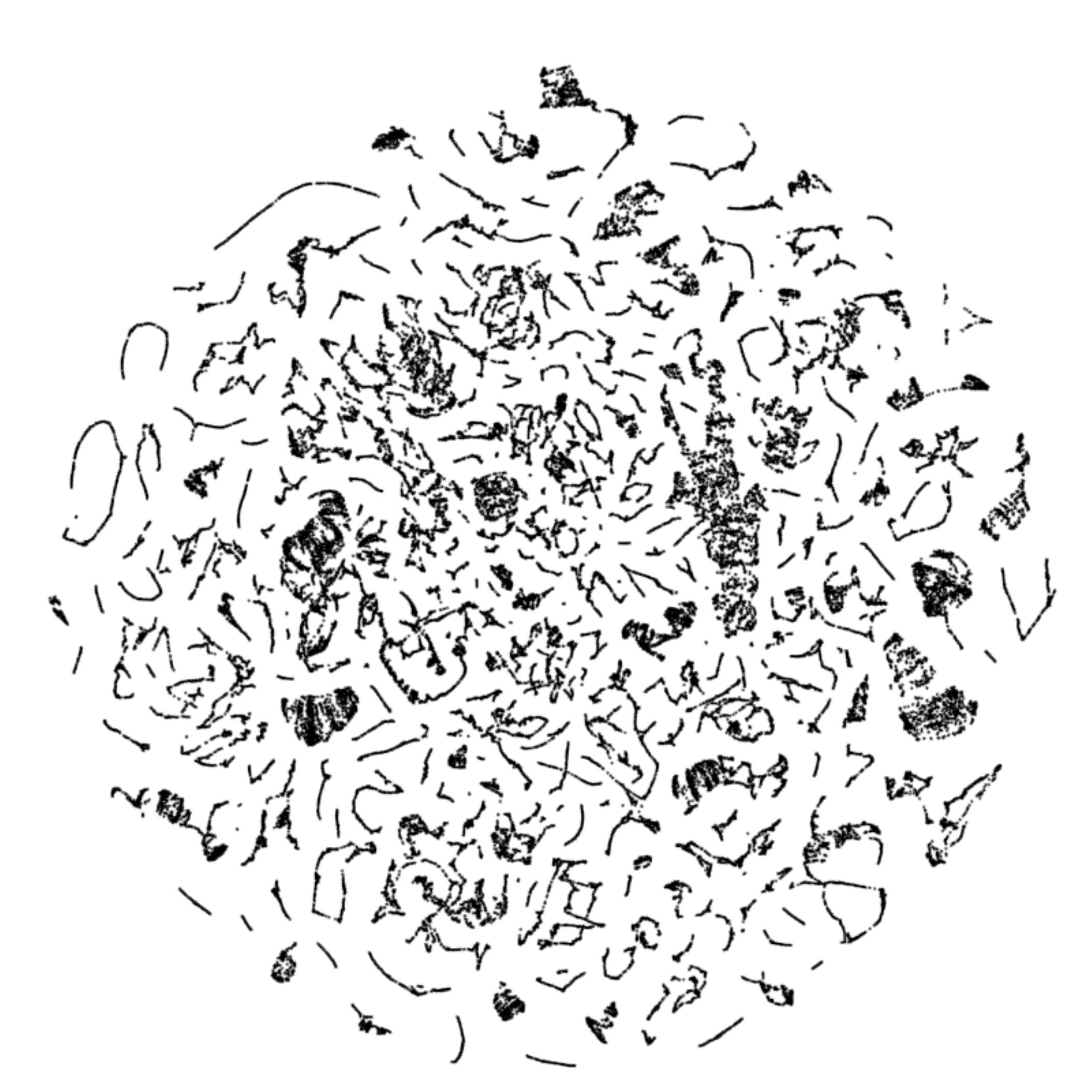}}}
	\subcaptionbox{perplexity=70}{\fbox{\includegraphics[width=\perpfigwidth,height=\perpfigwidth]{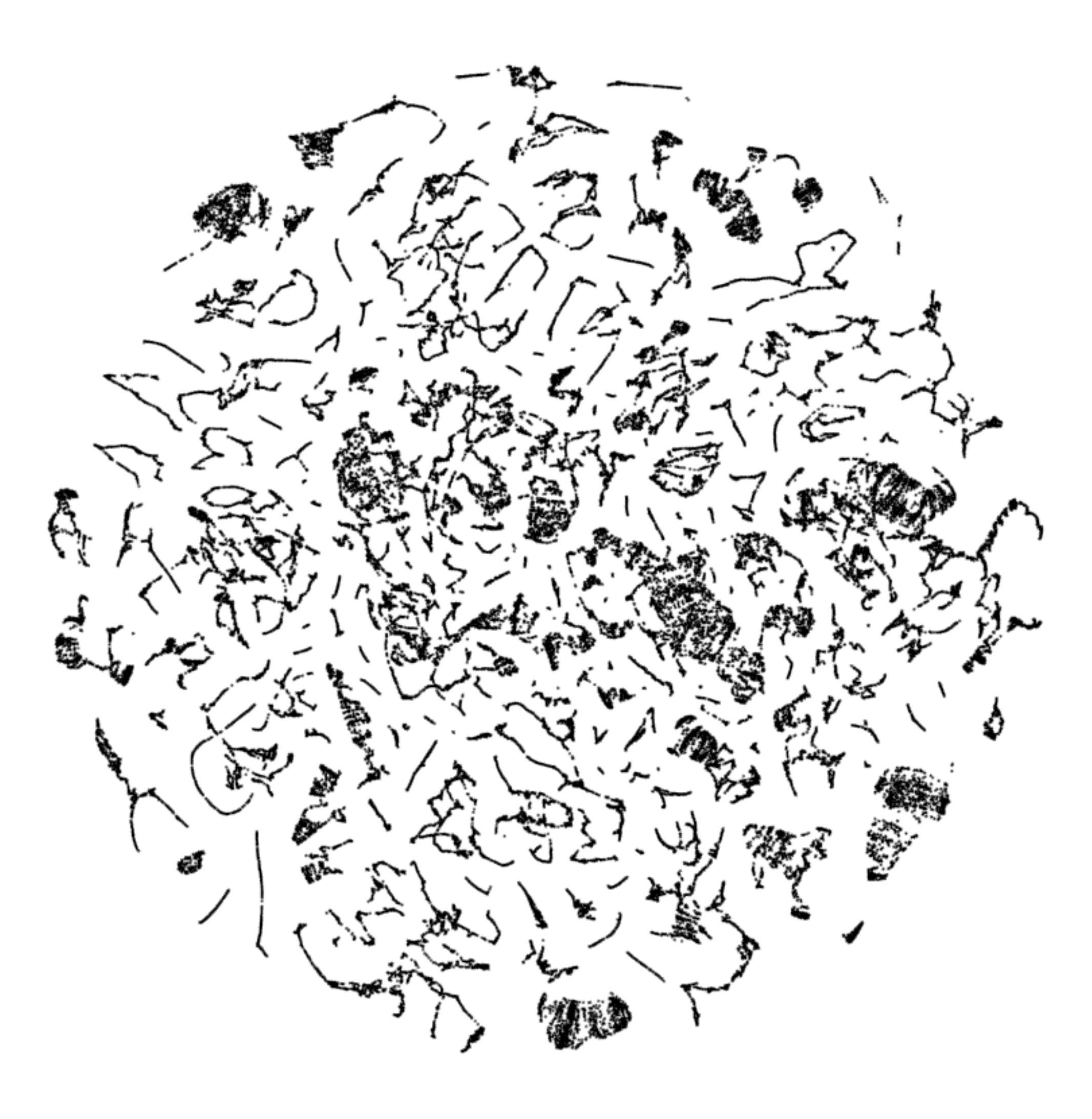}}}
	\subcaptionbox{perplexity=90}{\fbox{\includegraphics[width=\perpfigwidth,height=\perpfigwidth]{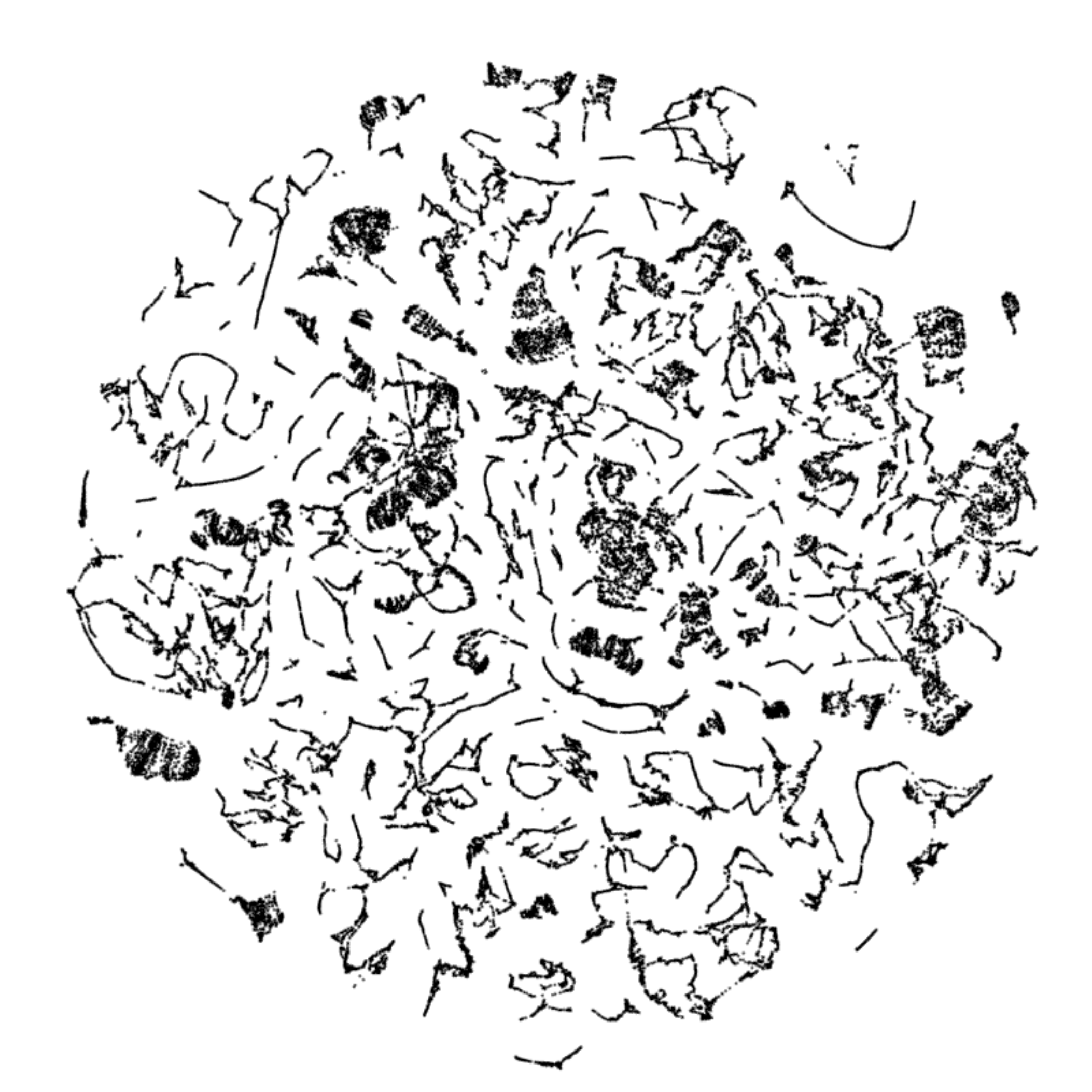}}}
	
	\vspace{\baselineskip}
	
	\subcaptionbox{perplexity=110}{\fbox{\includegraphics[width=\perpfigwidth,height=\perpfigwidth]{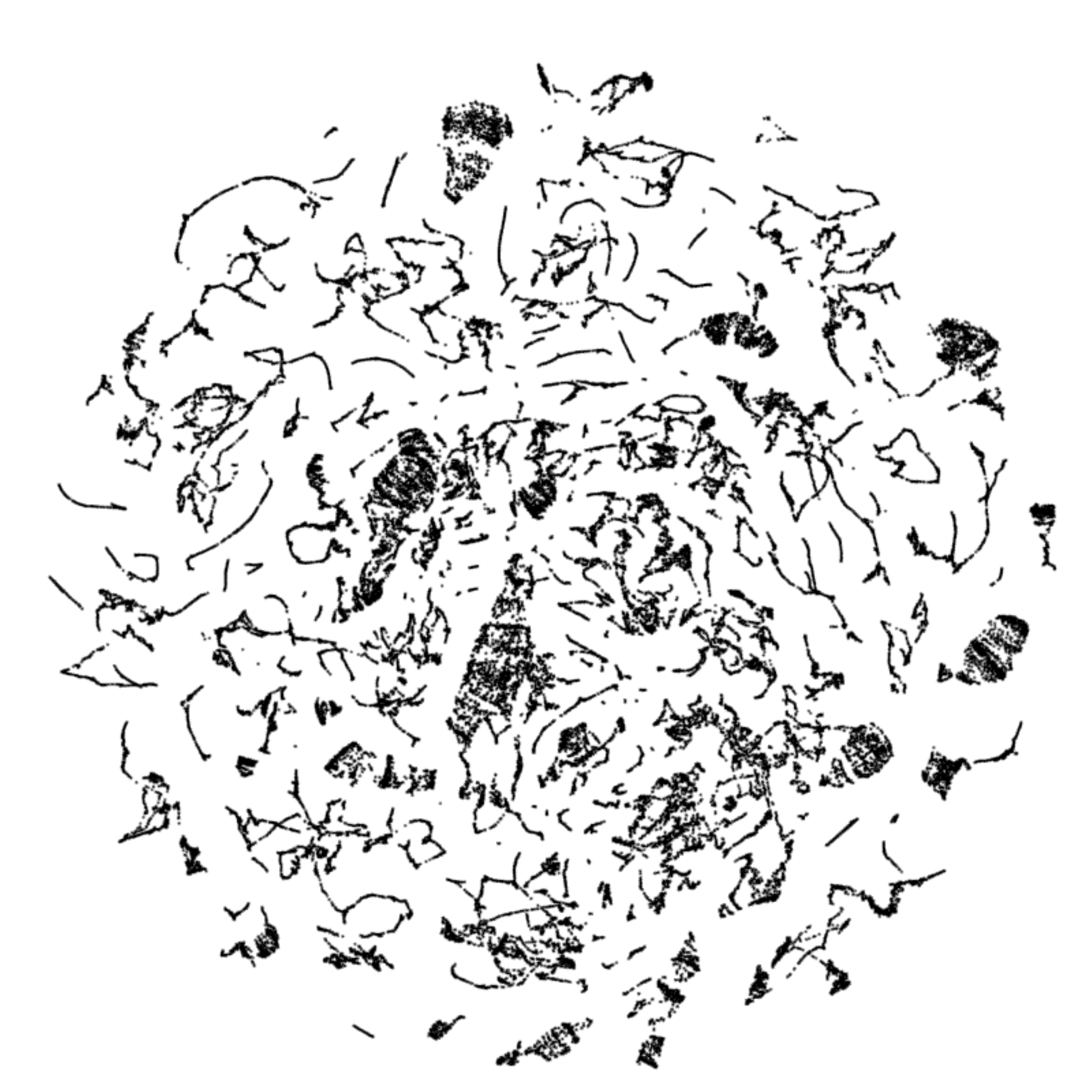}}}
	\subcaptionbox{perplexity=130}{\fbox{\includegraphics[width=\perpfigwidth,height=\perpfigwidth]{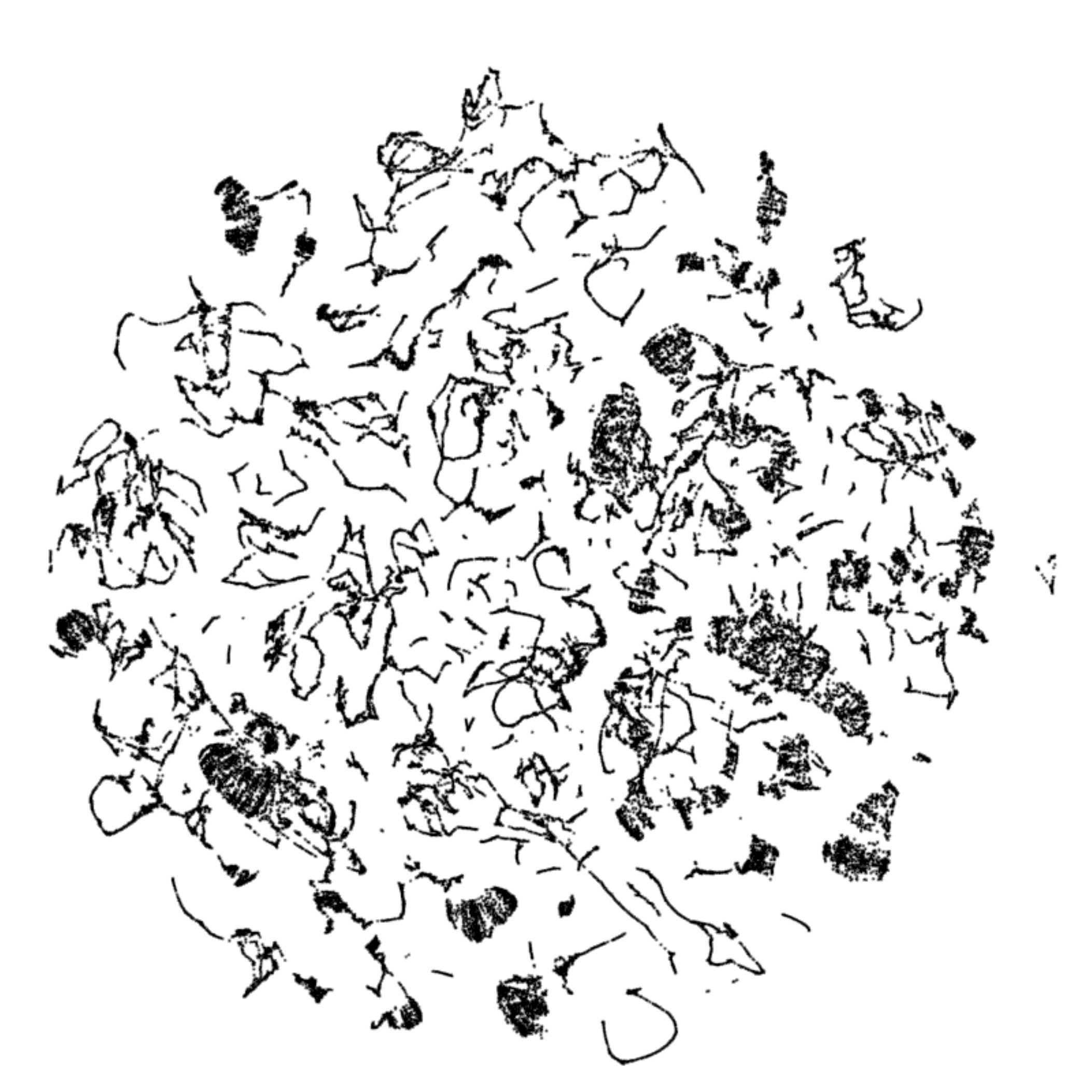}}}
	\subcaptionbox{perplexity=150}{\fbox{\includegraphics[width=\perpfigwidth,height=\perpfigwidth]{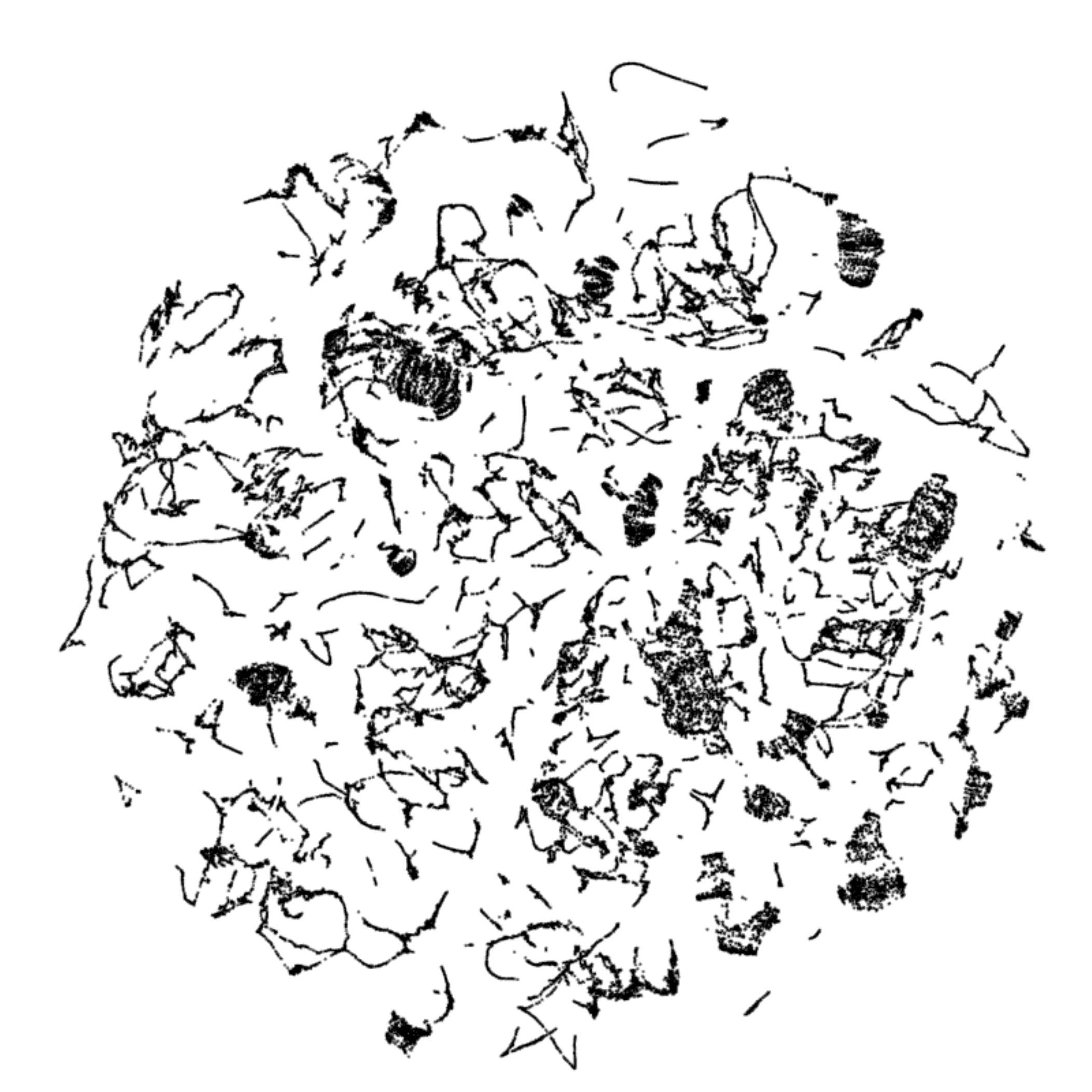}}}
	\caption{Visualizations of the \texttt{TOMORADAR} data set using t-SNE with various perplexities.}
	\label{fig:perpstomoradar}
\end{figure}

%===================== various degrees ===============================

\newcommand{\degreefigwidth}{5.2cm}
\begin{figure}[p]
	\centering
	\subcaptionbox{degree=2}{\fbox{\includegraphics[width=\degreefigwidth,height=\degreefigwidth]{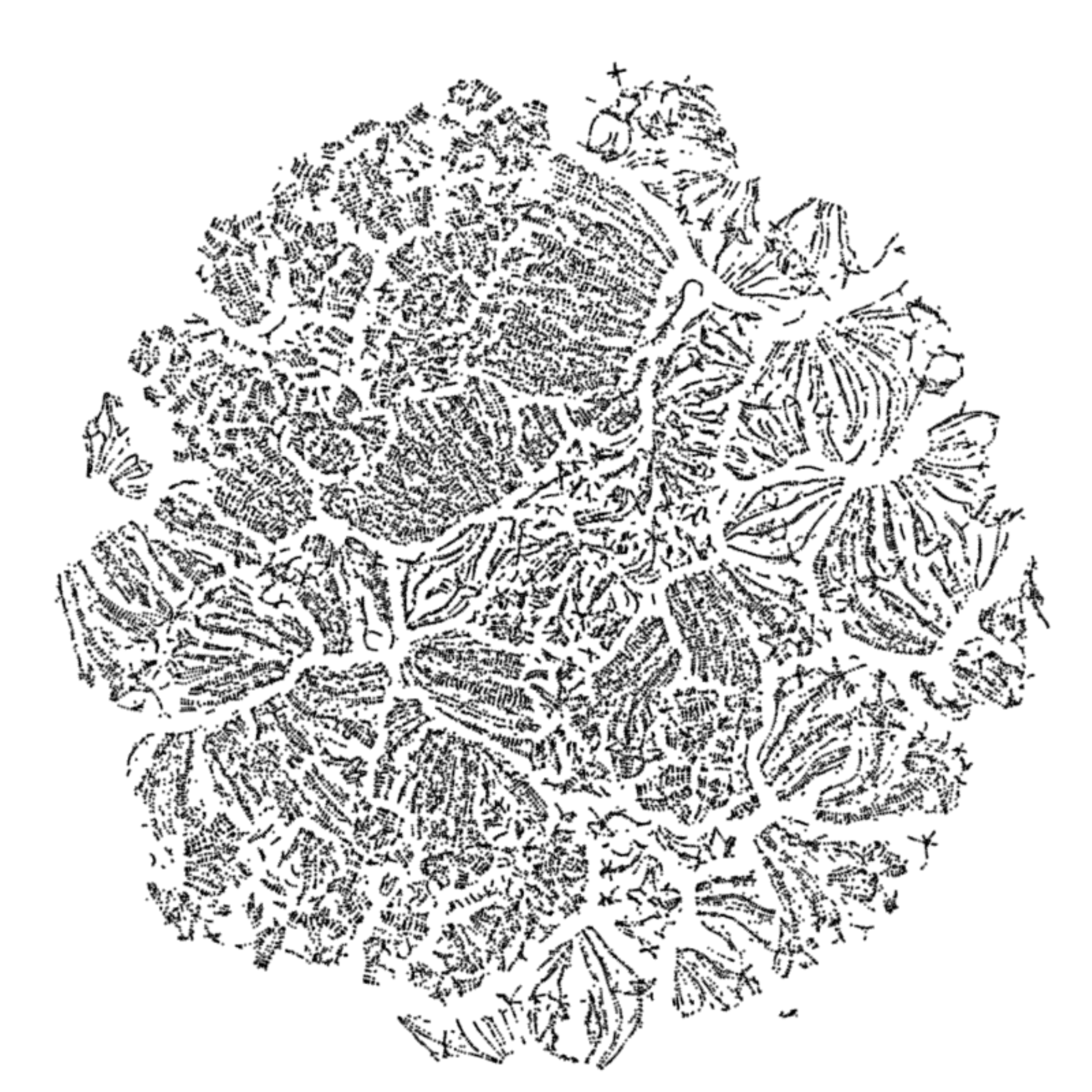}}}
	\subcaptionbox{degree=3}{\fbox{\includegraphics[width=\degreefigwidth,height=\degreefigwidth]{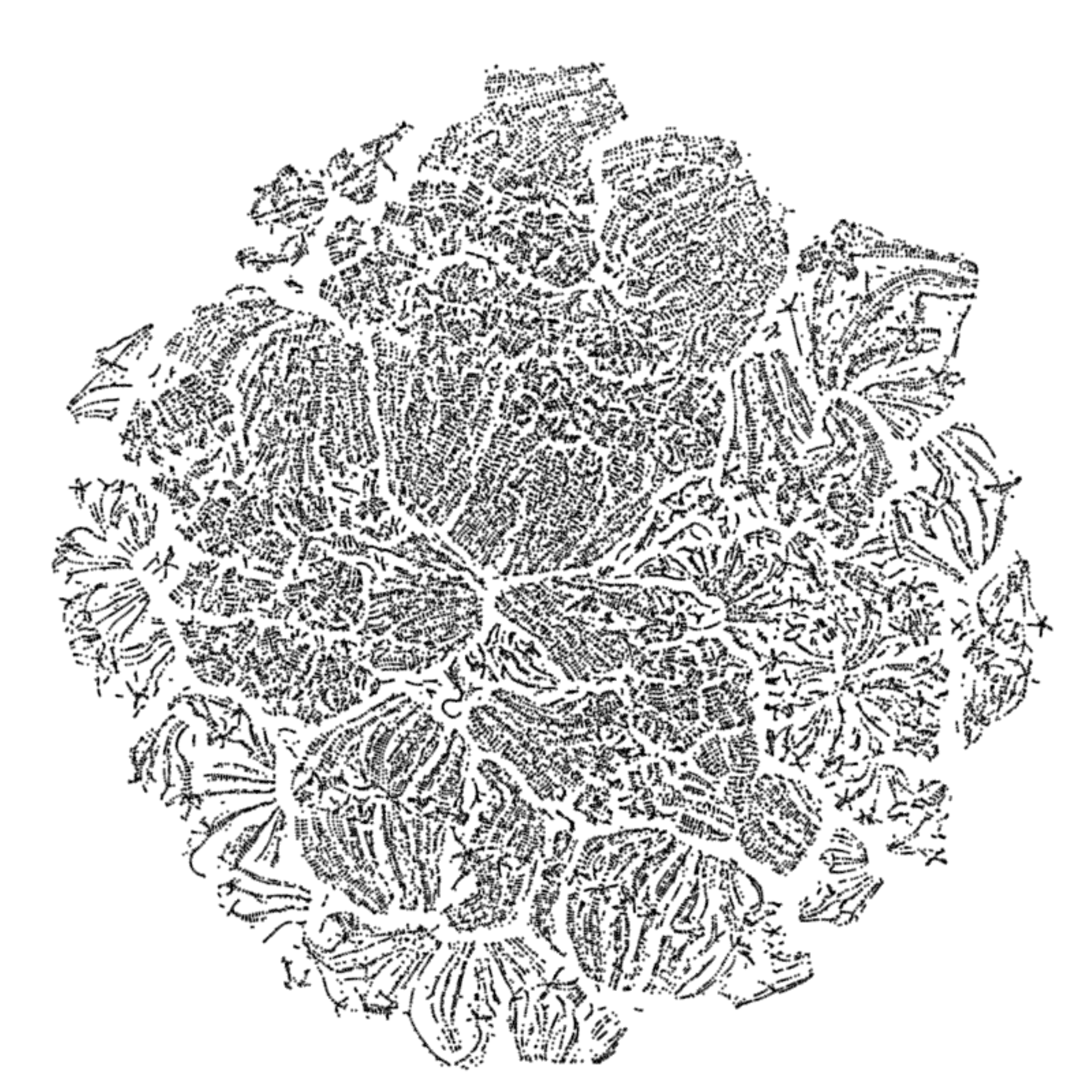}}}
	\subcaptionbox{degree=4}{\fbox{\includegraphics[width=\degreefigwidth,height=\degreefigwidth]{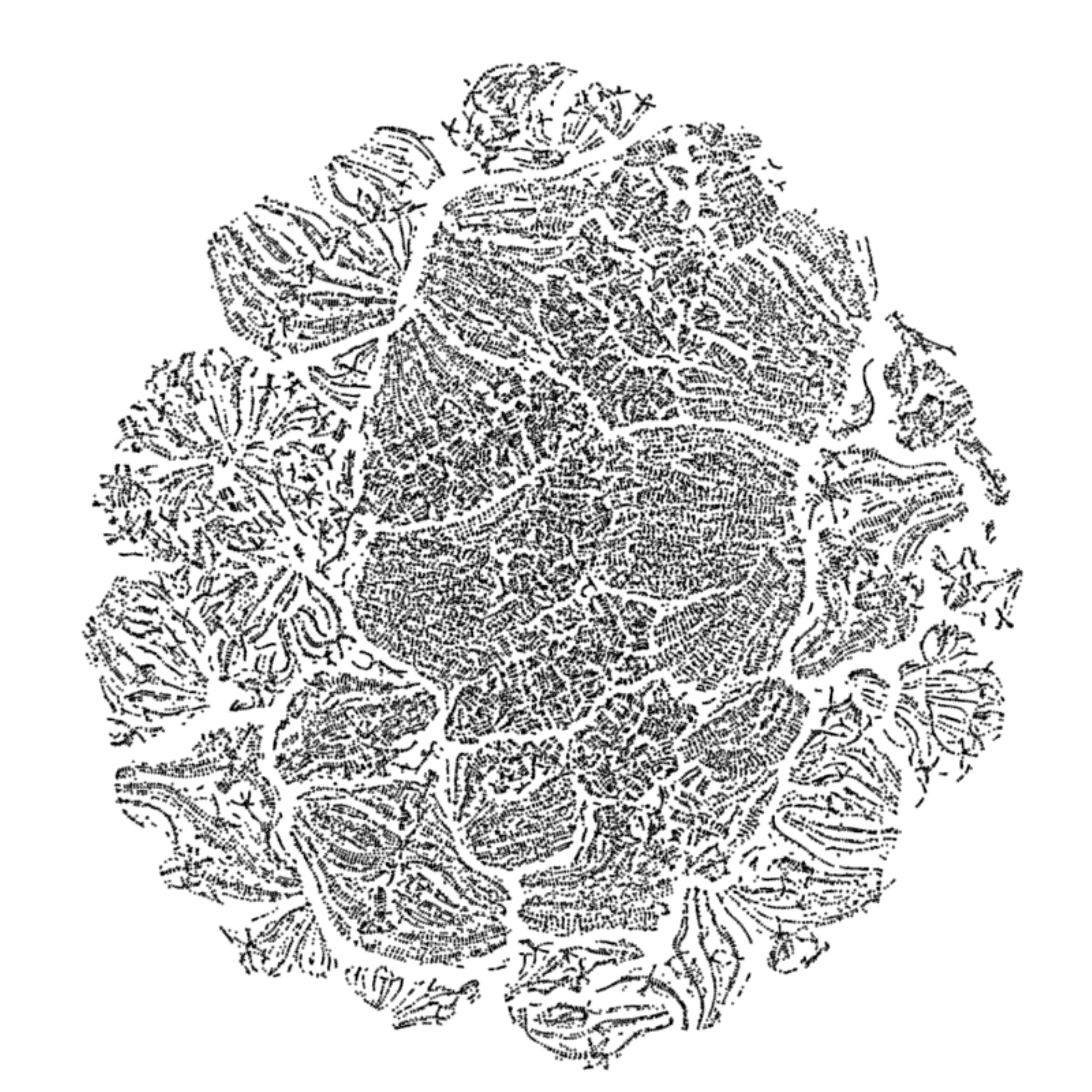}}}
	
	\vspace{\baselineskip}
	
	\subcaptionbox{degree=5}{\fbox{\includegraphics[width=\degreefigwidth,height=\degreefigwidth]{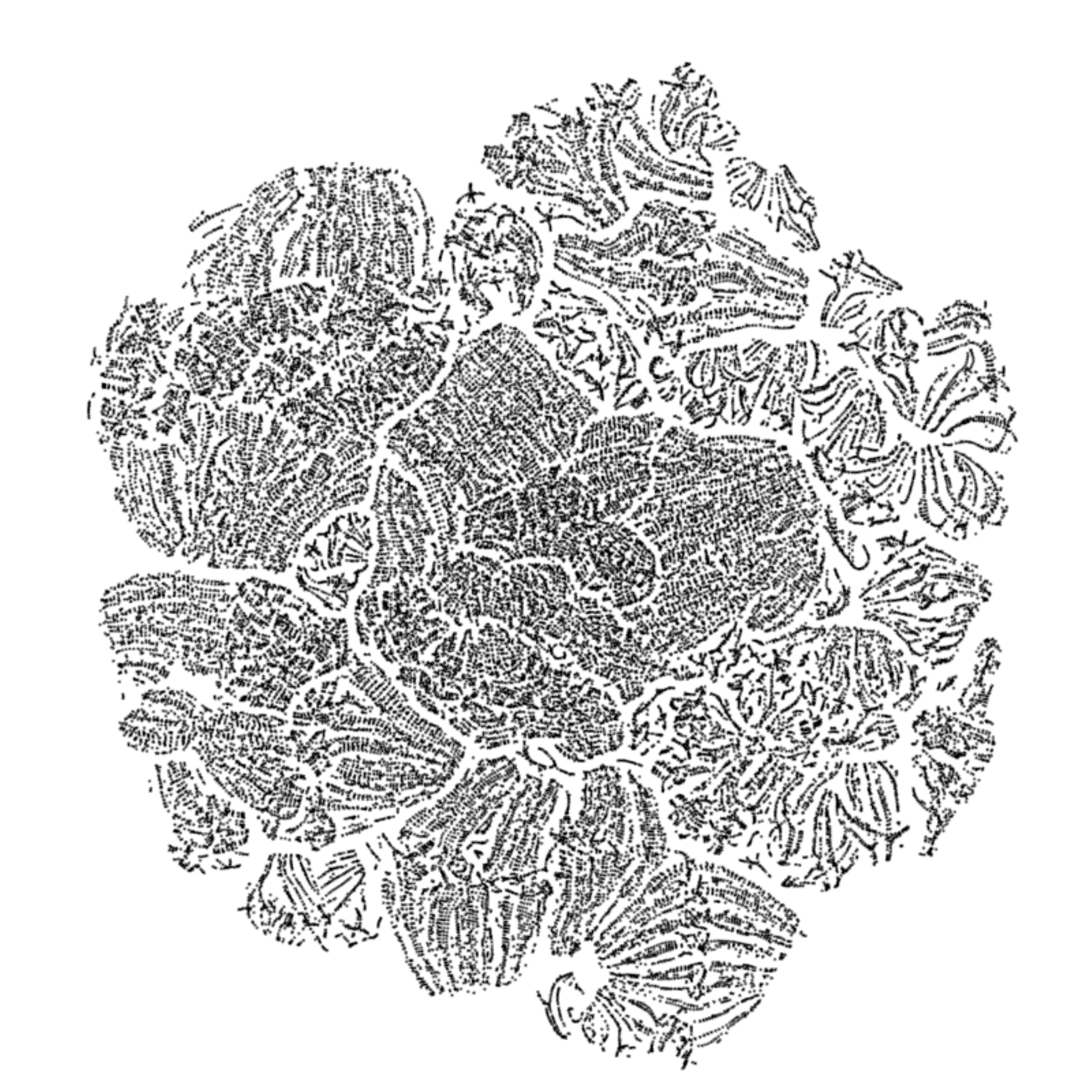}}}
	\subcaptionbox{degree=7}{\fbox{\includegraphics[width=\degreefigwidth,height=\degreefigwidth]{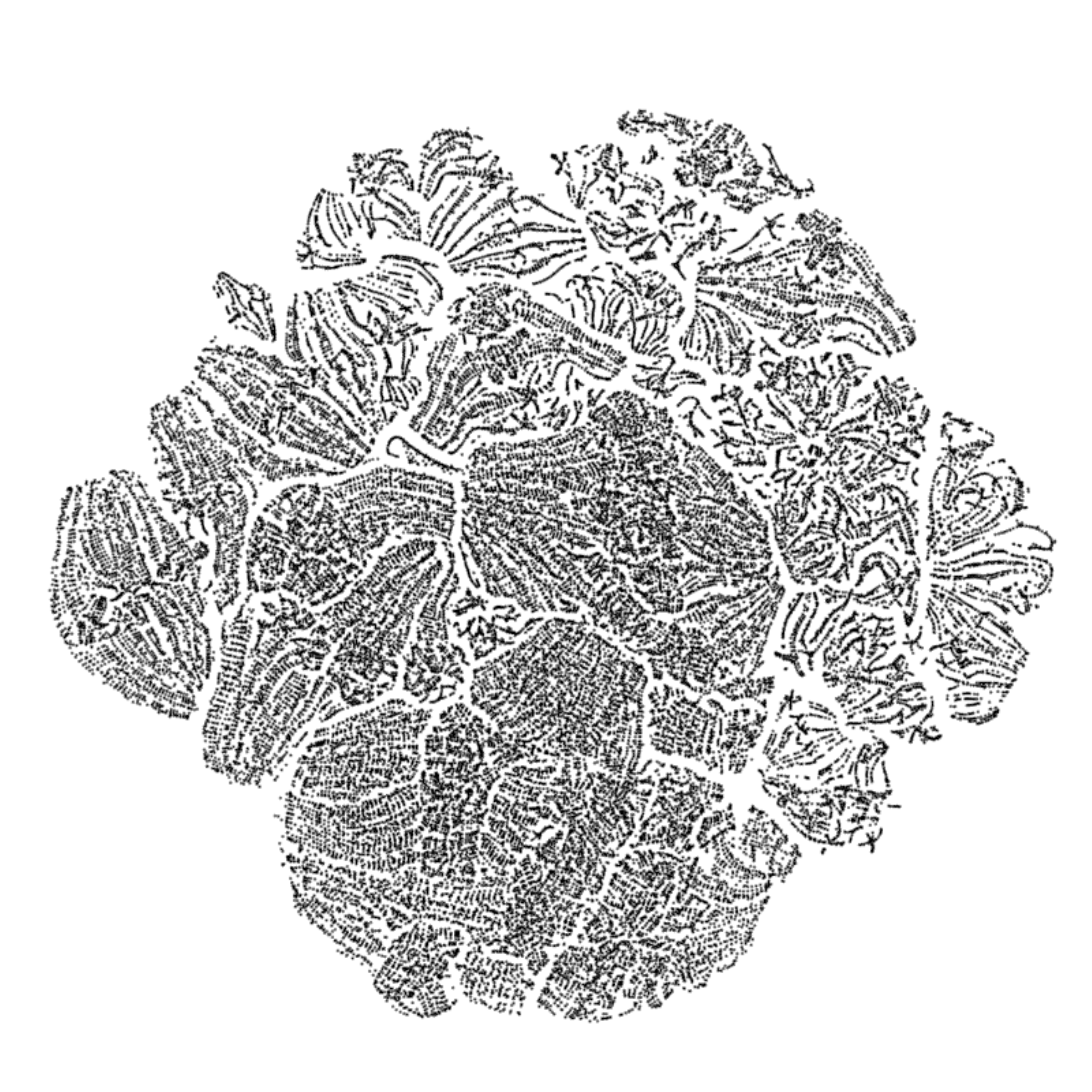}}}
	\subcaptionbox{degree=9}{\fbox{\includegraphics[width=\degreefigwidth,height=\degreefigwidth]{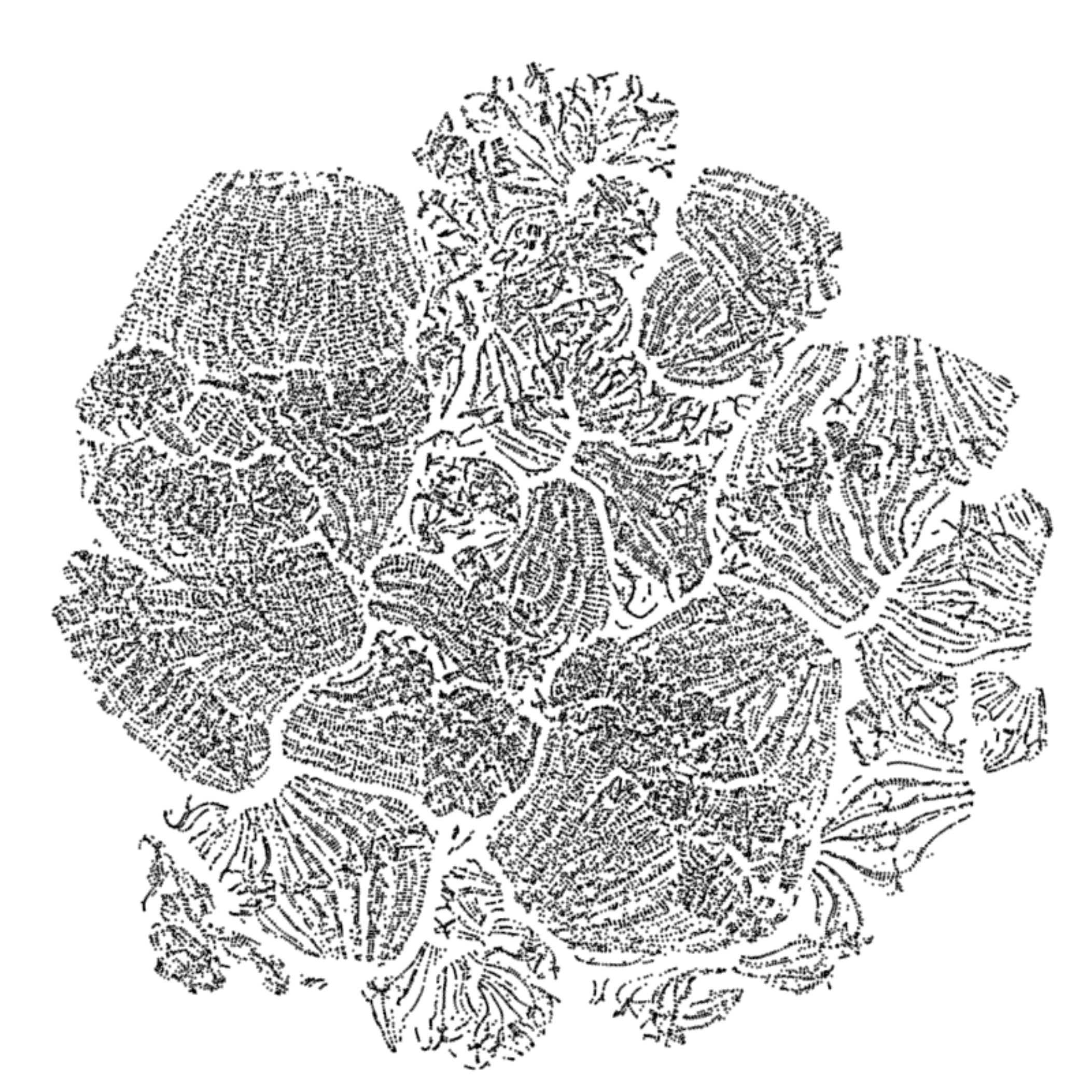}}}
	
	\vspace{\baselineskip}
	
	\subcaptionbox{degree=11}{\fbox{\includegraphics[width=\degreefigwidth,height=\degreefigwidth]{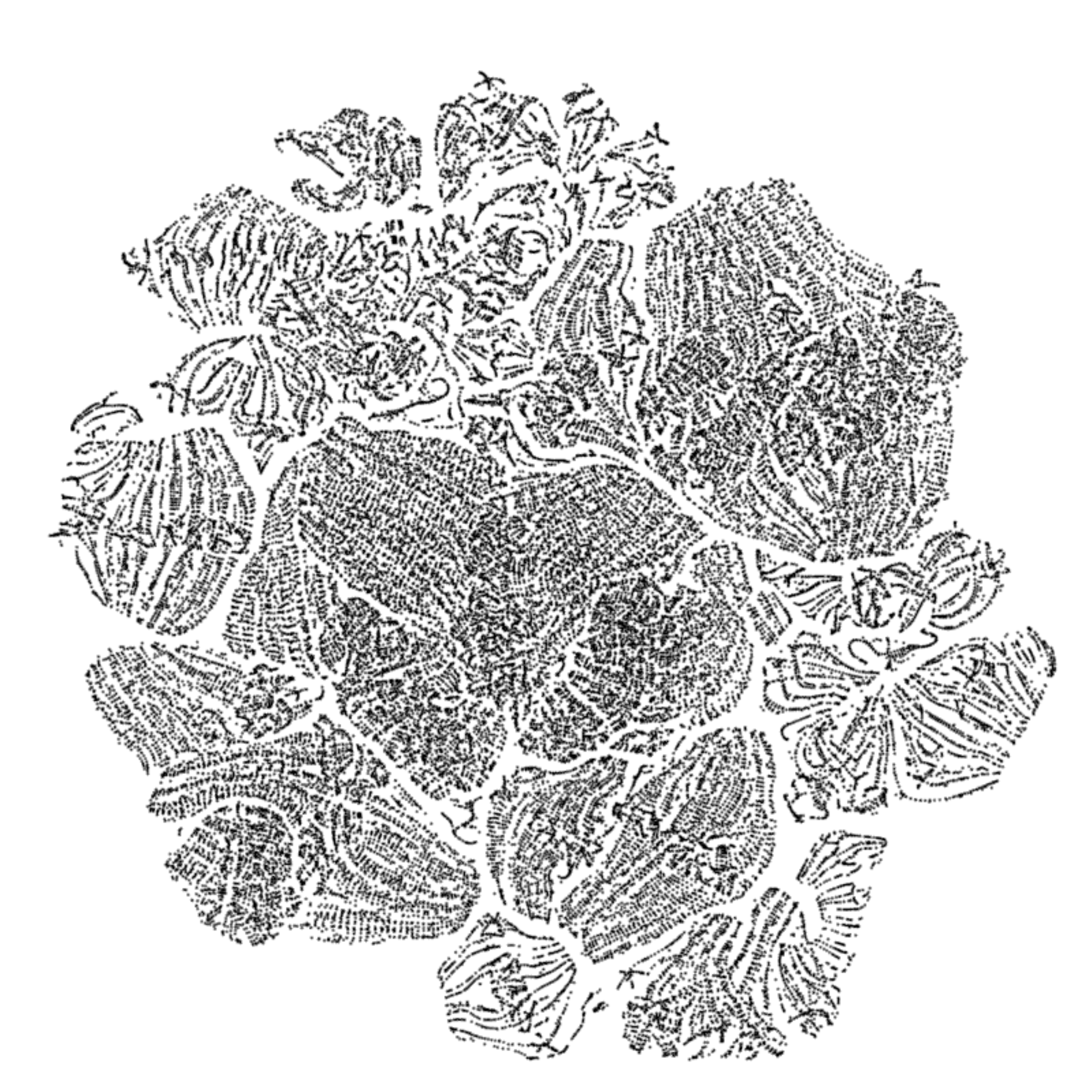}}}
	\subcaptionbox{degree=13}{\fbox{\includegraphics[width=\degreefigwidth,height=\degreefigwidth]{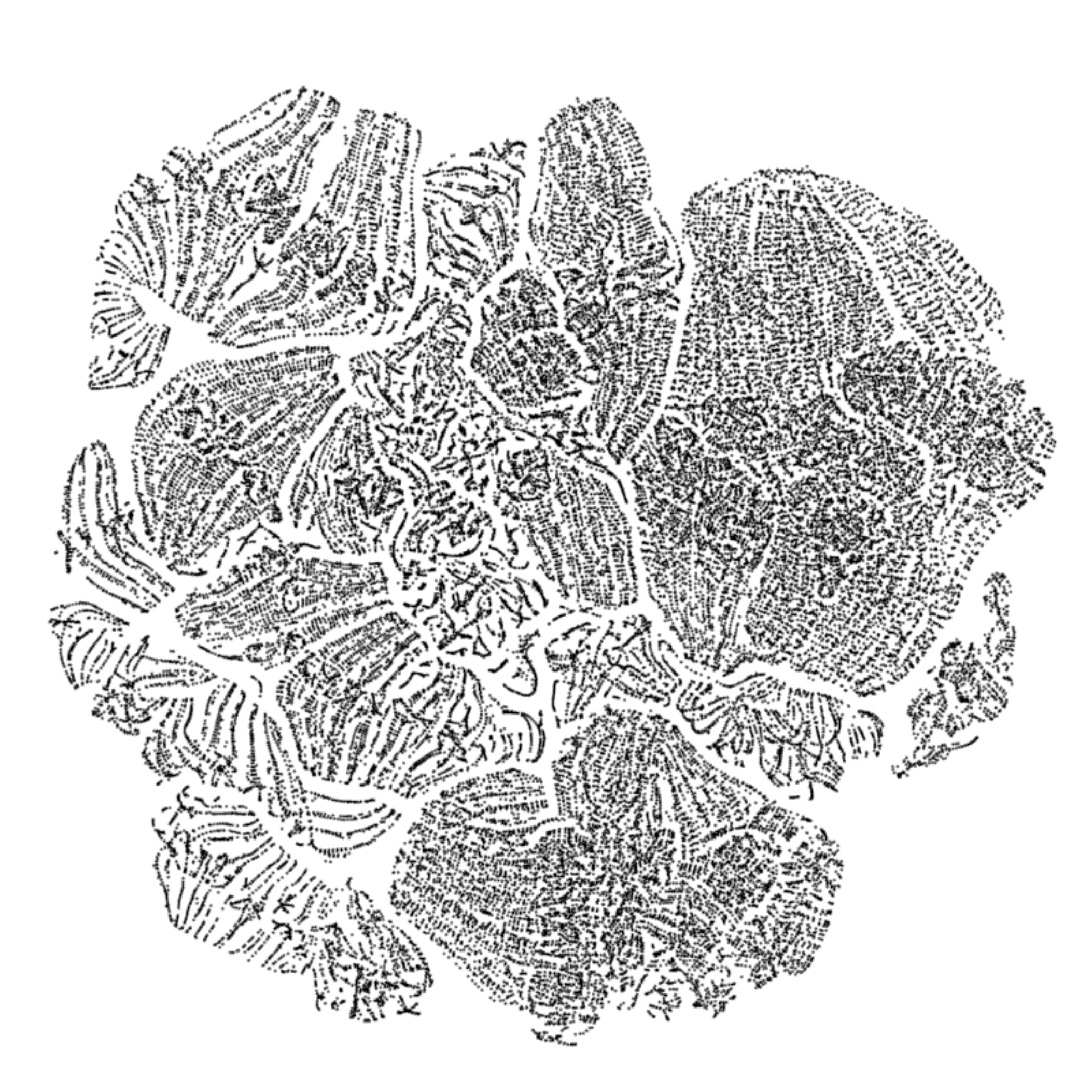}}}
	\subcaptionbox{degree=15}{\fbox{\includegraphics[width=\degreefigwidth,height=\degreefigwidth]{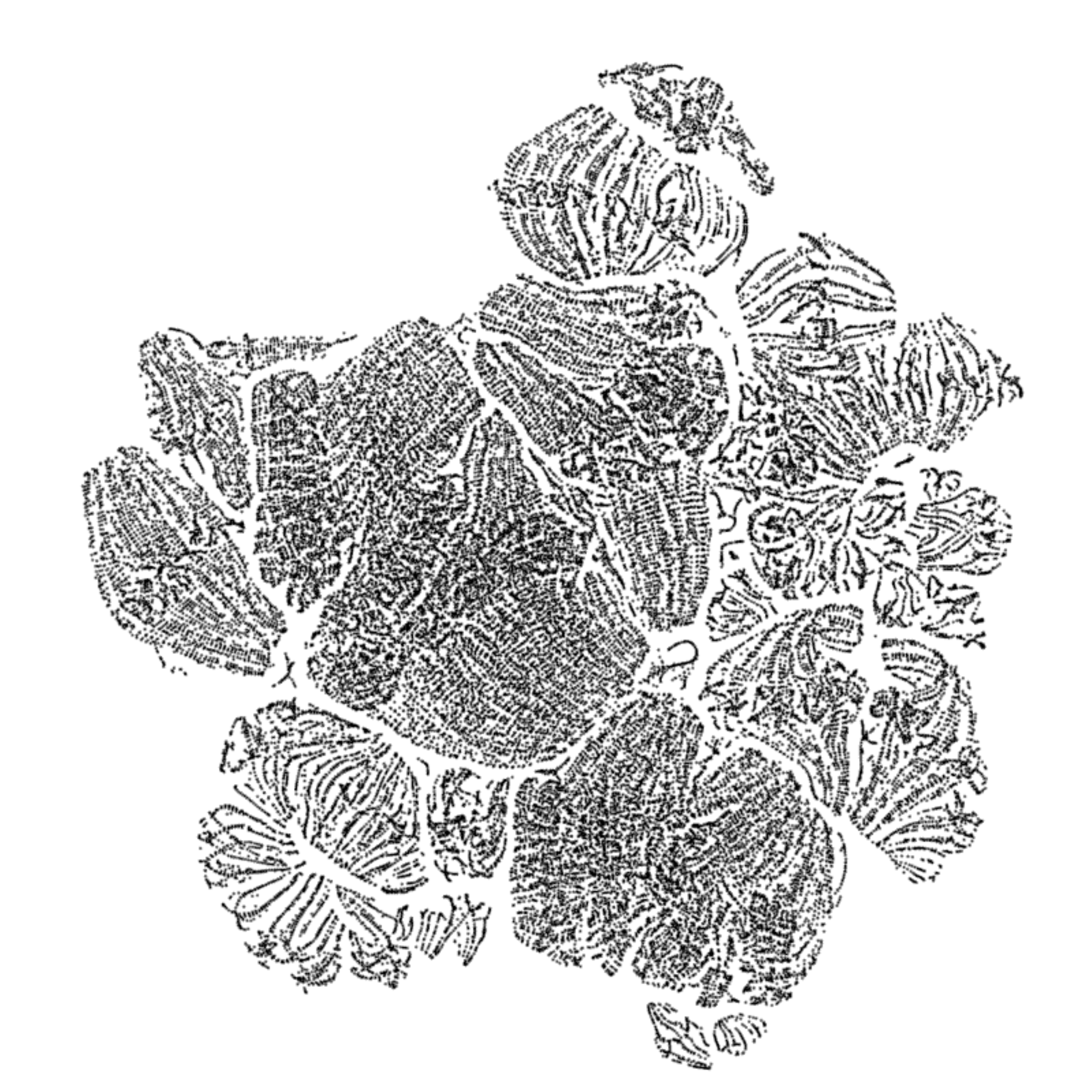}}}
	\caption{Visualizations of the \texttt{SHUTTLE} data set using t-SNE with various degree of freedom in the Student t-distribution. See Figure \ref{fig:tsnevis} for the standard t-SNE visualization (with degree=1).}
	\label{fig:degreesshuttle}
\end{figure}

\begin{figure}[p]
	\centering
	\subcaptionbox{degree=2}{\fbox{\includegraphics[width=\degreefigwidth,height=\degreefigwidth]{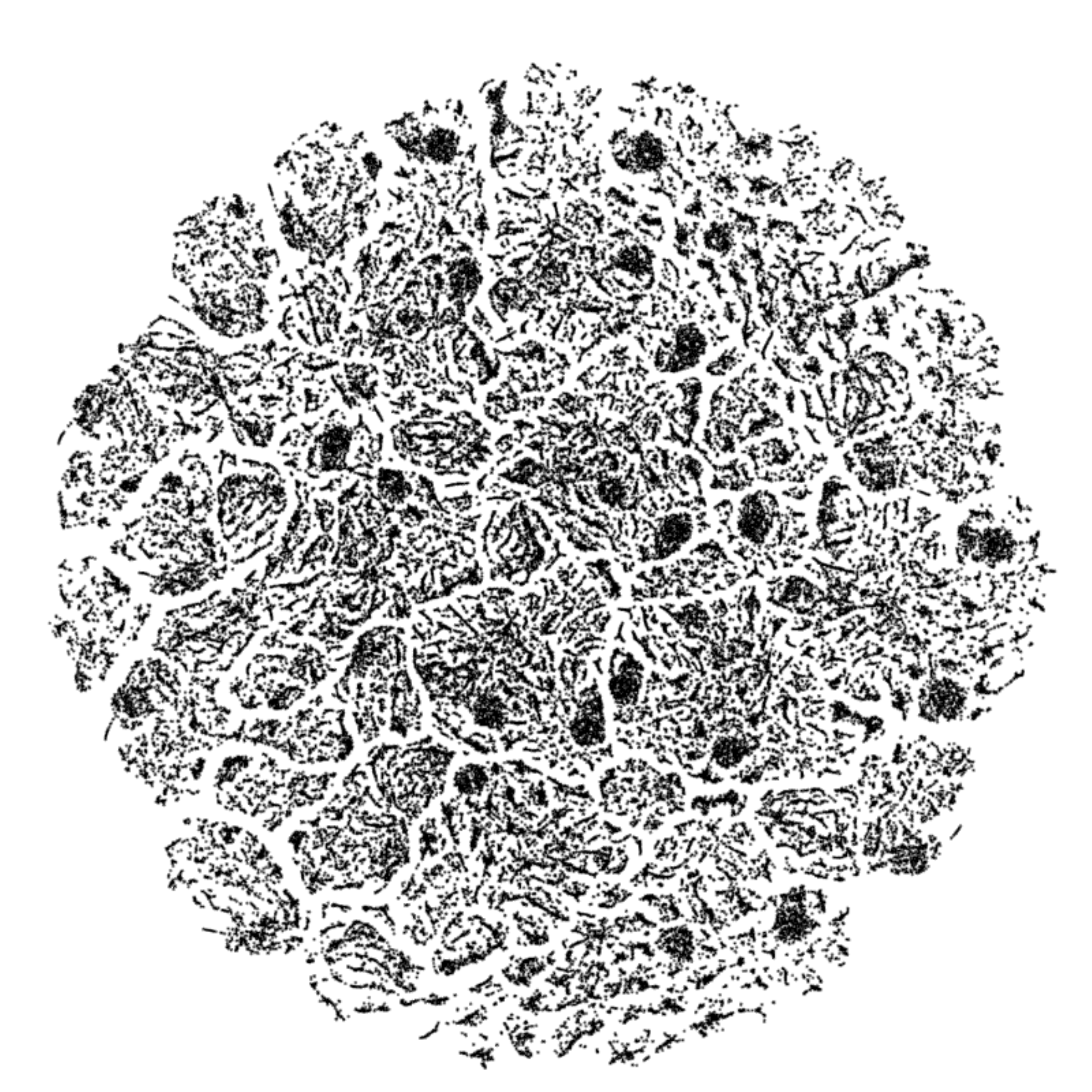}}}
	\subcaptionbox{degree=3}{\fbox{\includegraphics[width=\degreefigwidth,height=\degreefigwidth]{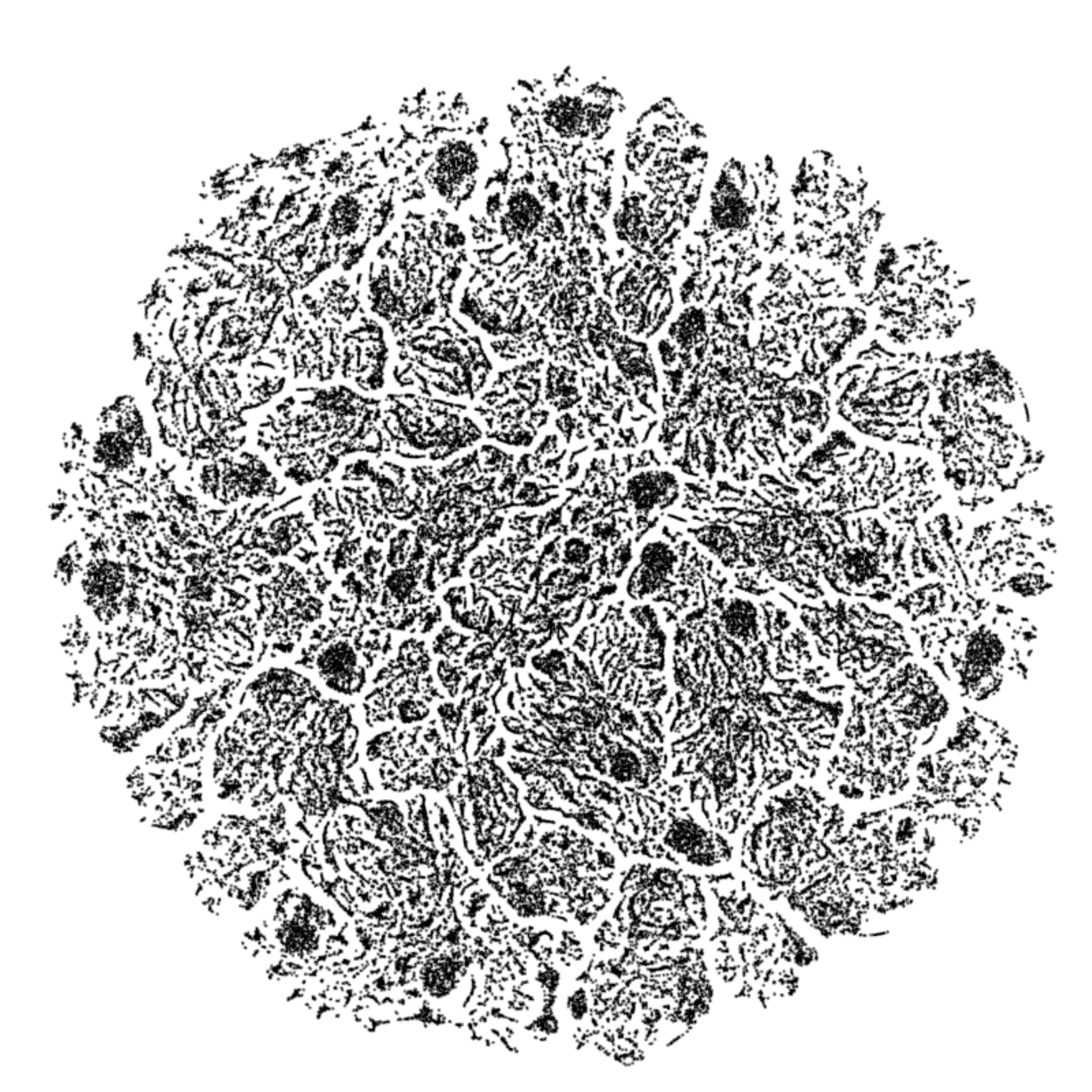}}}
	\subcaptionbox{degree=4}{\fbox{\includegraphics[width=\degreefigwidth,height=\degreefigwidth]{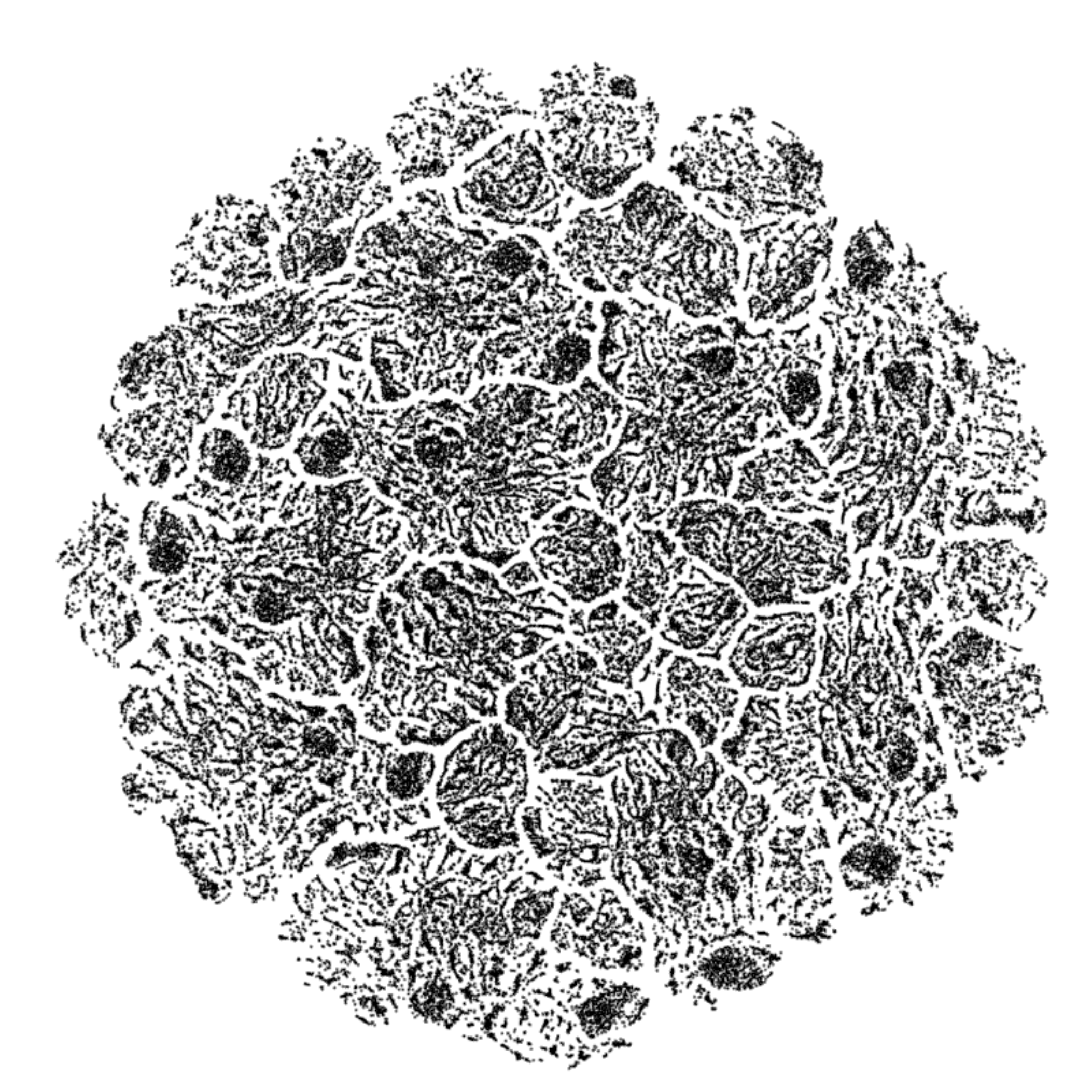}}}
	
	\vspace{\baselineskip}
	
	\subcaptionbox{degree=5}{\fbox{\includegraphics[width=\degreefigwidth,height=\degreefigwidth]{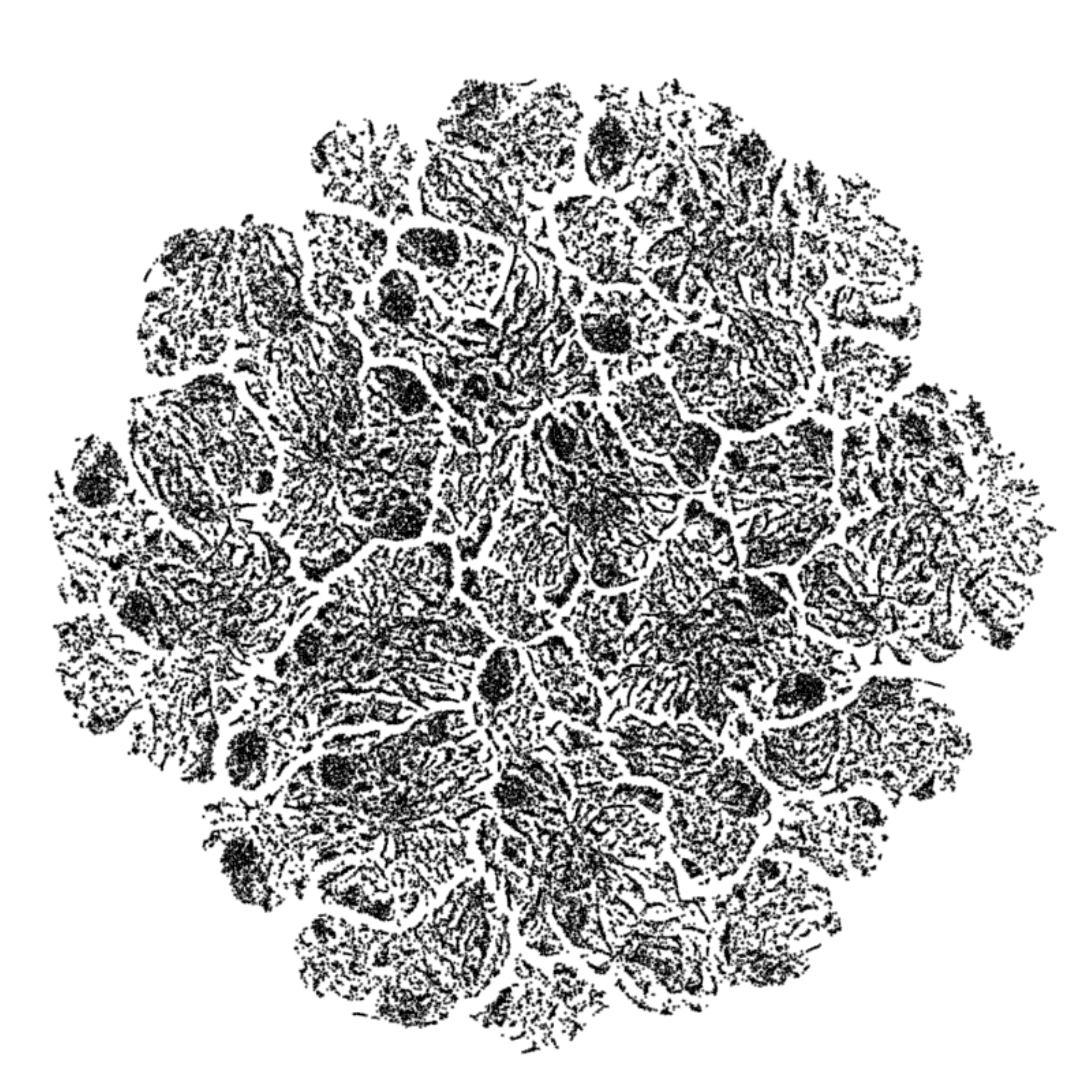}}}
	\subcaptionbox{degree=7}{\fbox{\includegraphics[width=\degreefigwidth,height=\degreefigwidth]{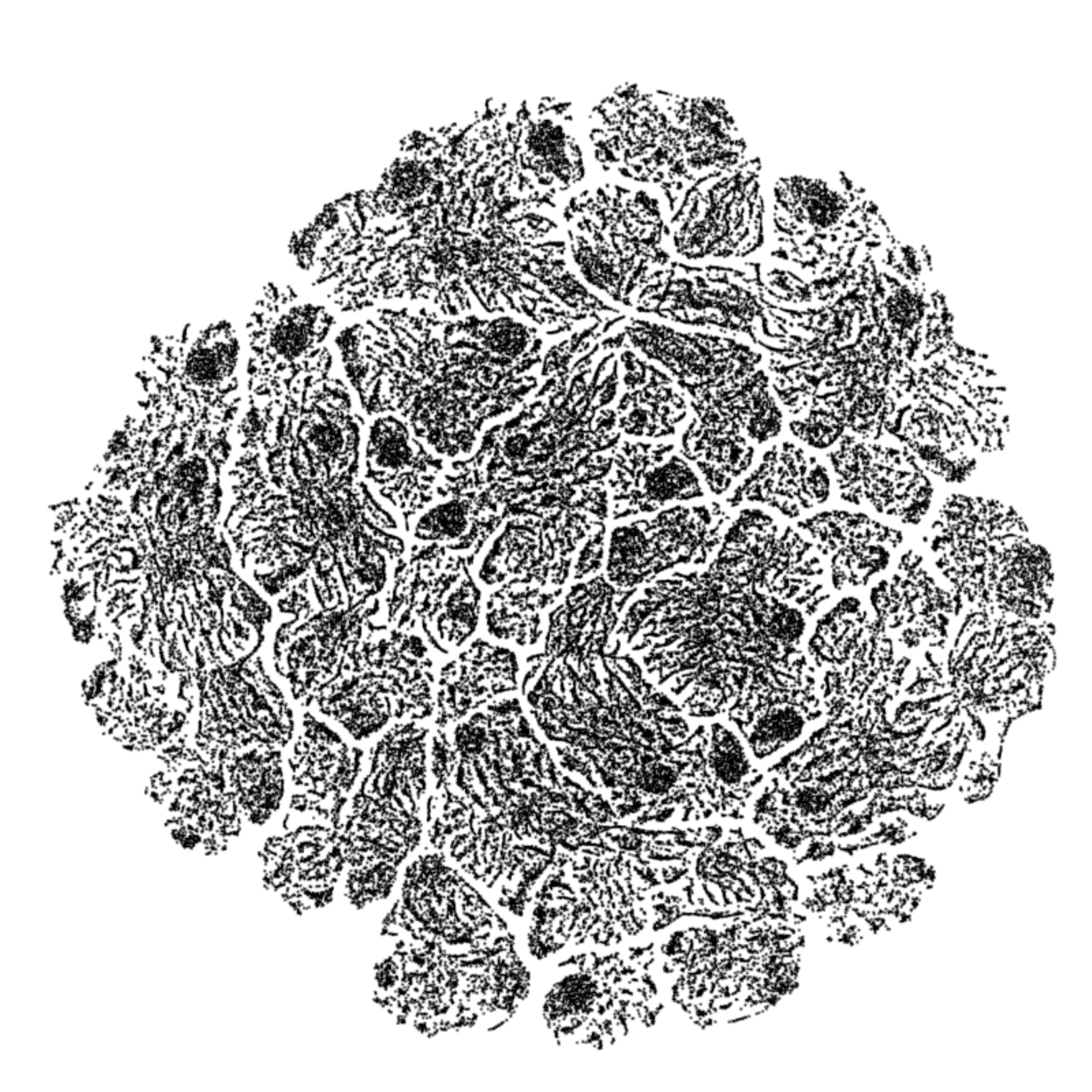}}}
	\subcaptionbox{degree=9}{\fbox{\includegraphics[width=\degreefigwidth,height=\degreefigwidth]{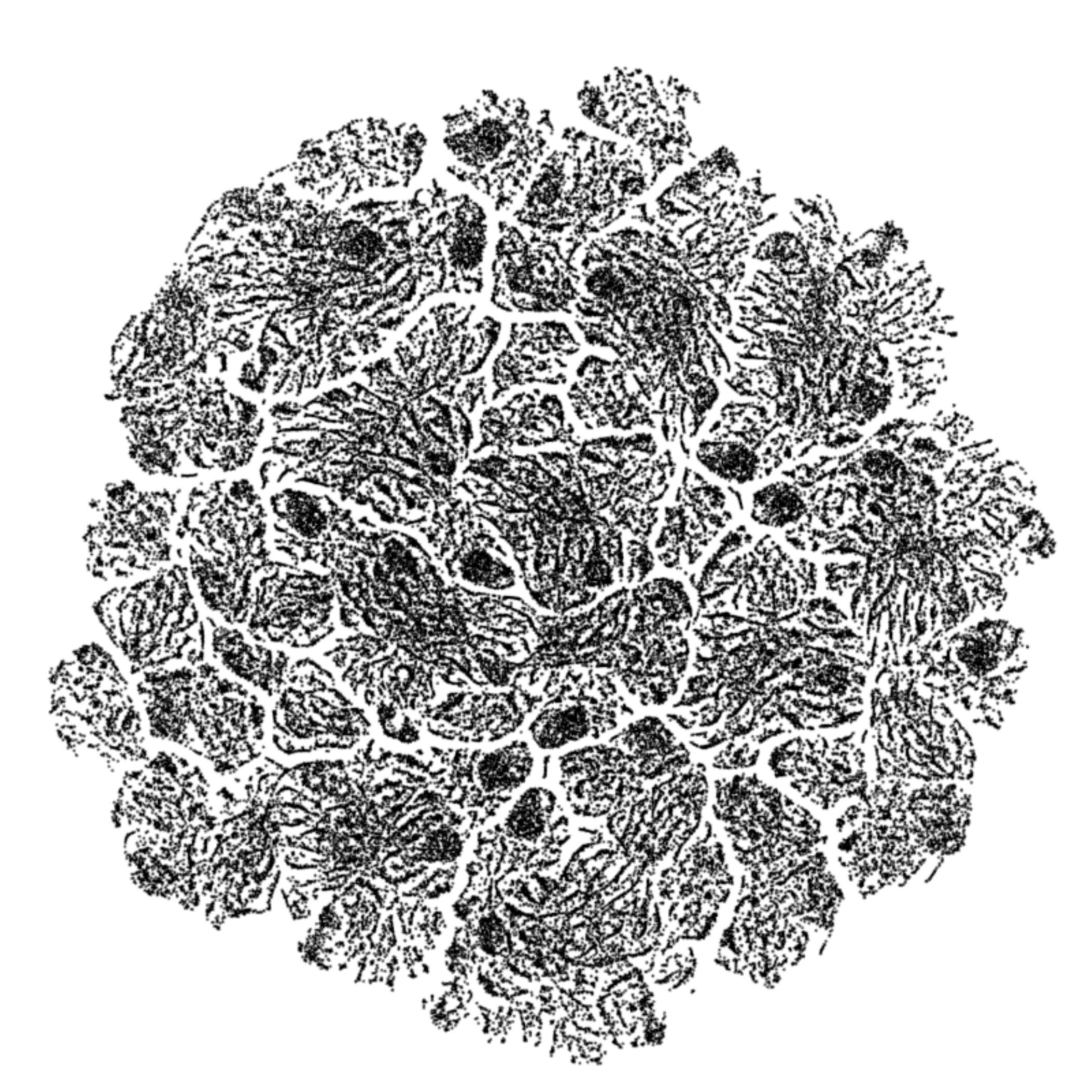}}}
	
	\vspace{\baselineskip}
	
	\subcaptionbox{degree=11}{\fbox{\includegraphics[width=\degreefigwidth,height=\degreefigwidth]{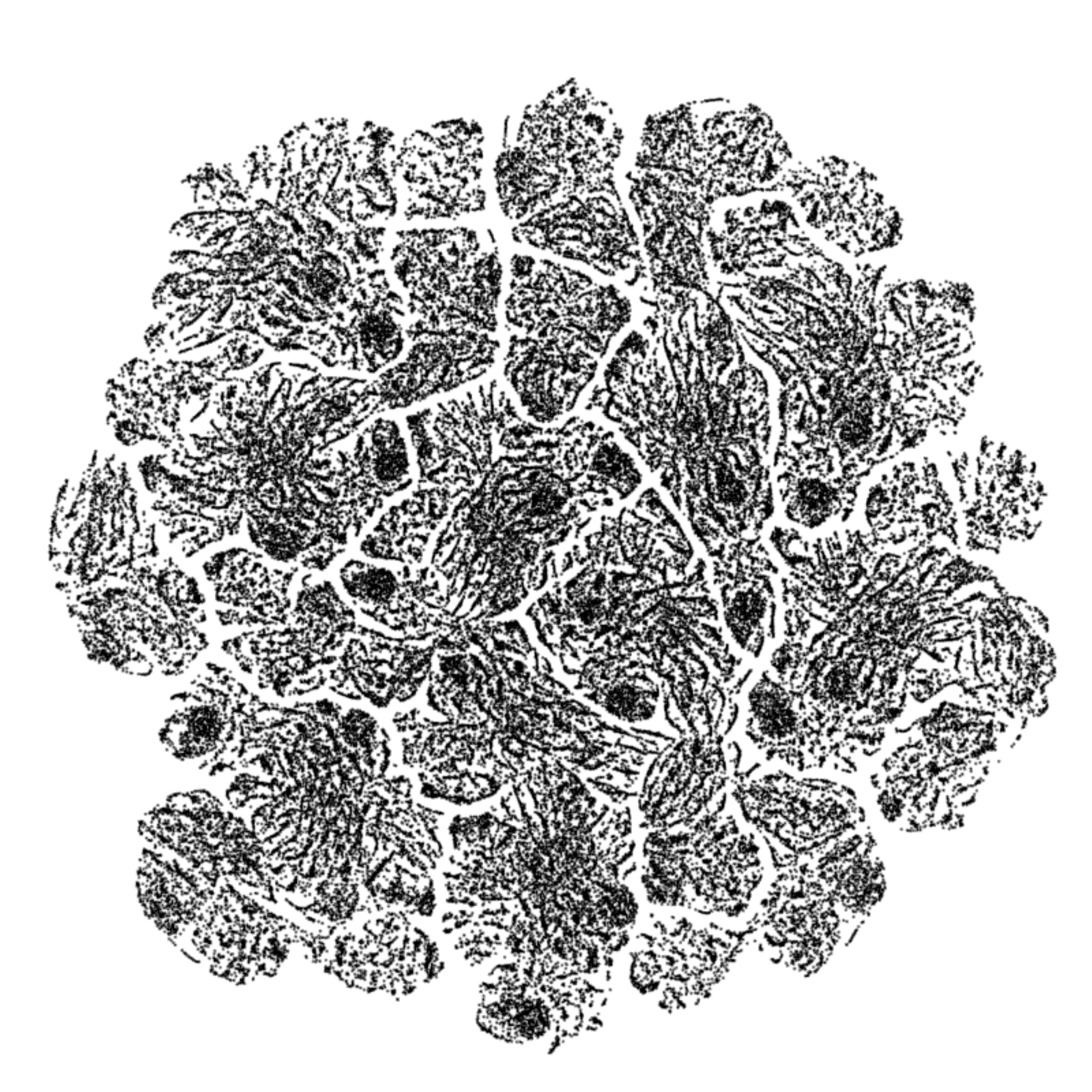}}}
	\subcaptionbox{degree=13}{\fbox{\includegraphics[width=\degreefigwidth,height=\degreefigwidth]{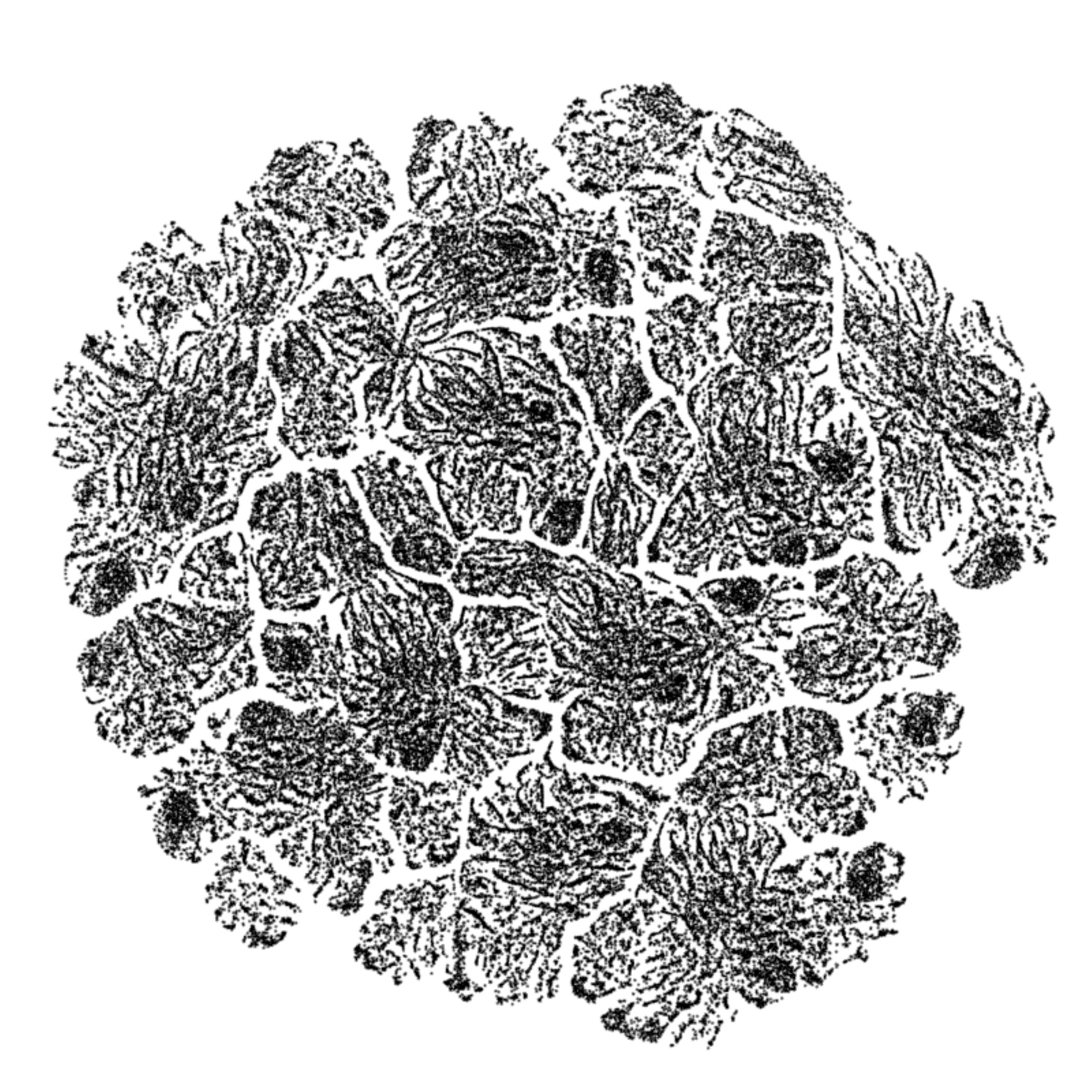}}}
	\subcaptionbox{degree=15}{\fbox{\includegraphics[width=\degreefigwidth,height=\degreefigwidth]{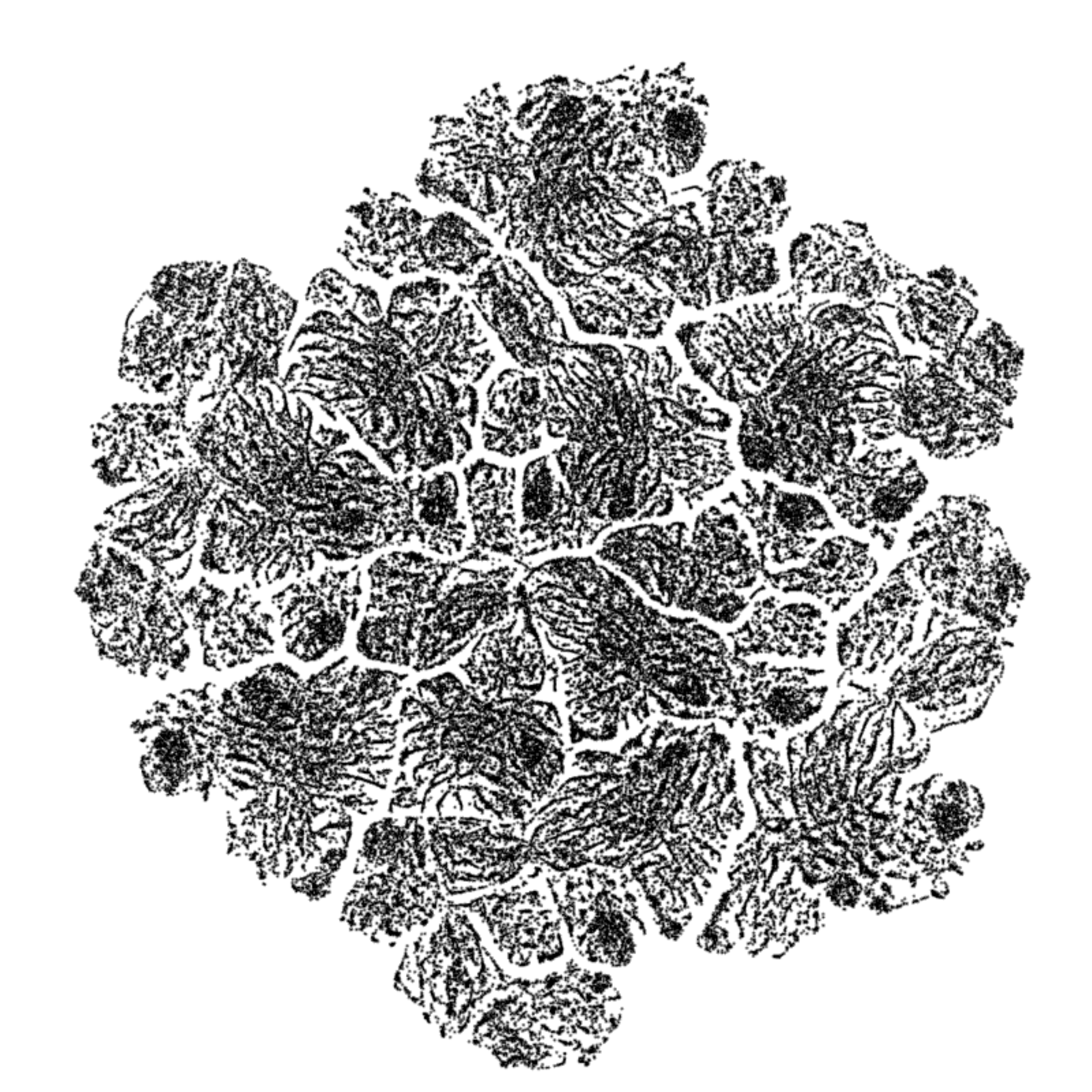}}}
	\caption{Visualizations of the \texttt{IJCNN} data set using t-SNE with various degree of freedom in the Student t-distribution. See Figure \ref{fig:tsnevis} for the standard t-SNE visualization (with degree=1).}
	\label{fig:degreesijcnn}
\end{figure}

\begin{figure}[p]
	\centering
	\subcaptionbox{degree=2}{\fbox{\includegraphics[width=\degreefigwidth,height=\degreefigwidth]{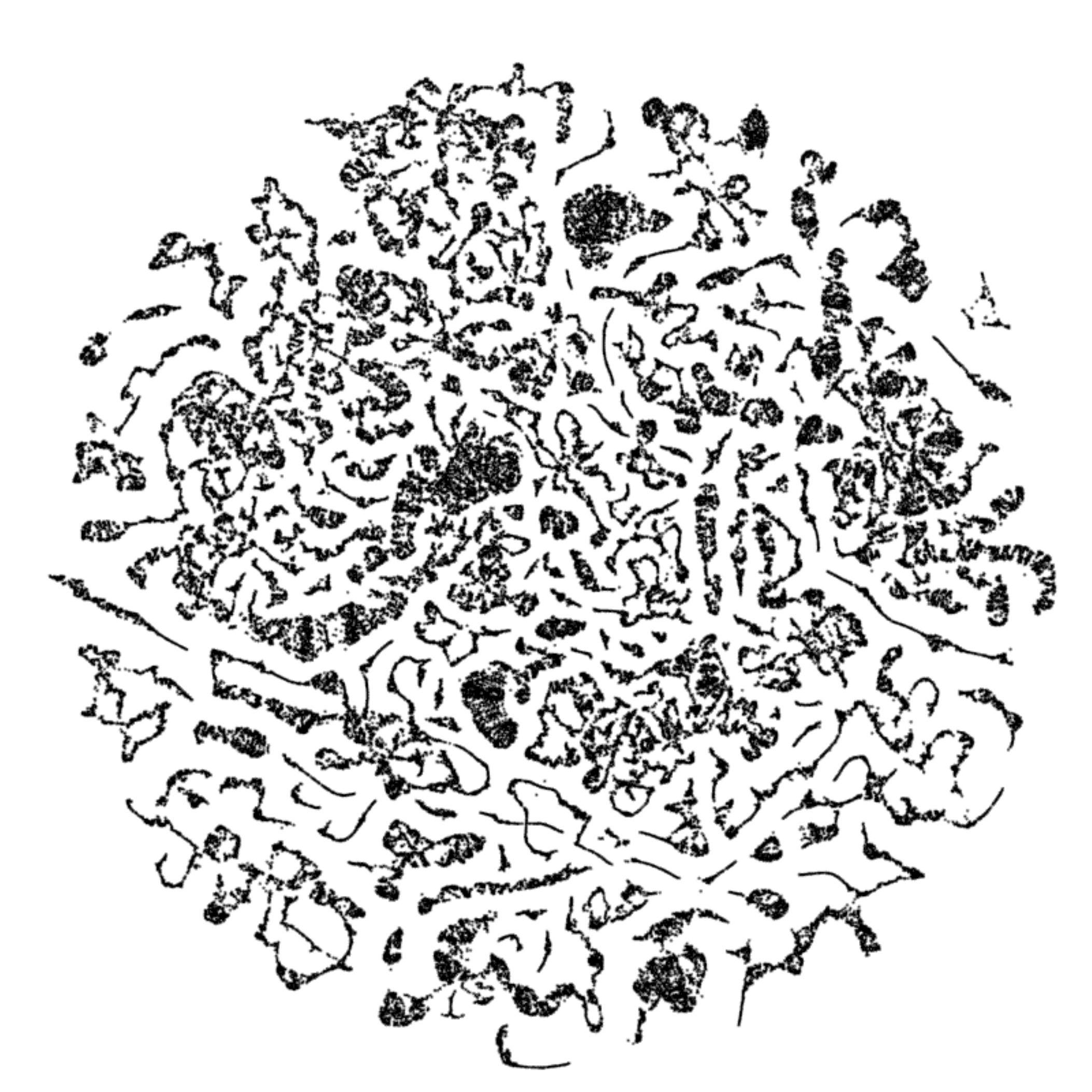}}}
	\subcaptionbox{degree=3}{\fbox{\includegraphics[width=\degreefigwidth,height=\degreefigwidth]{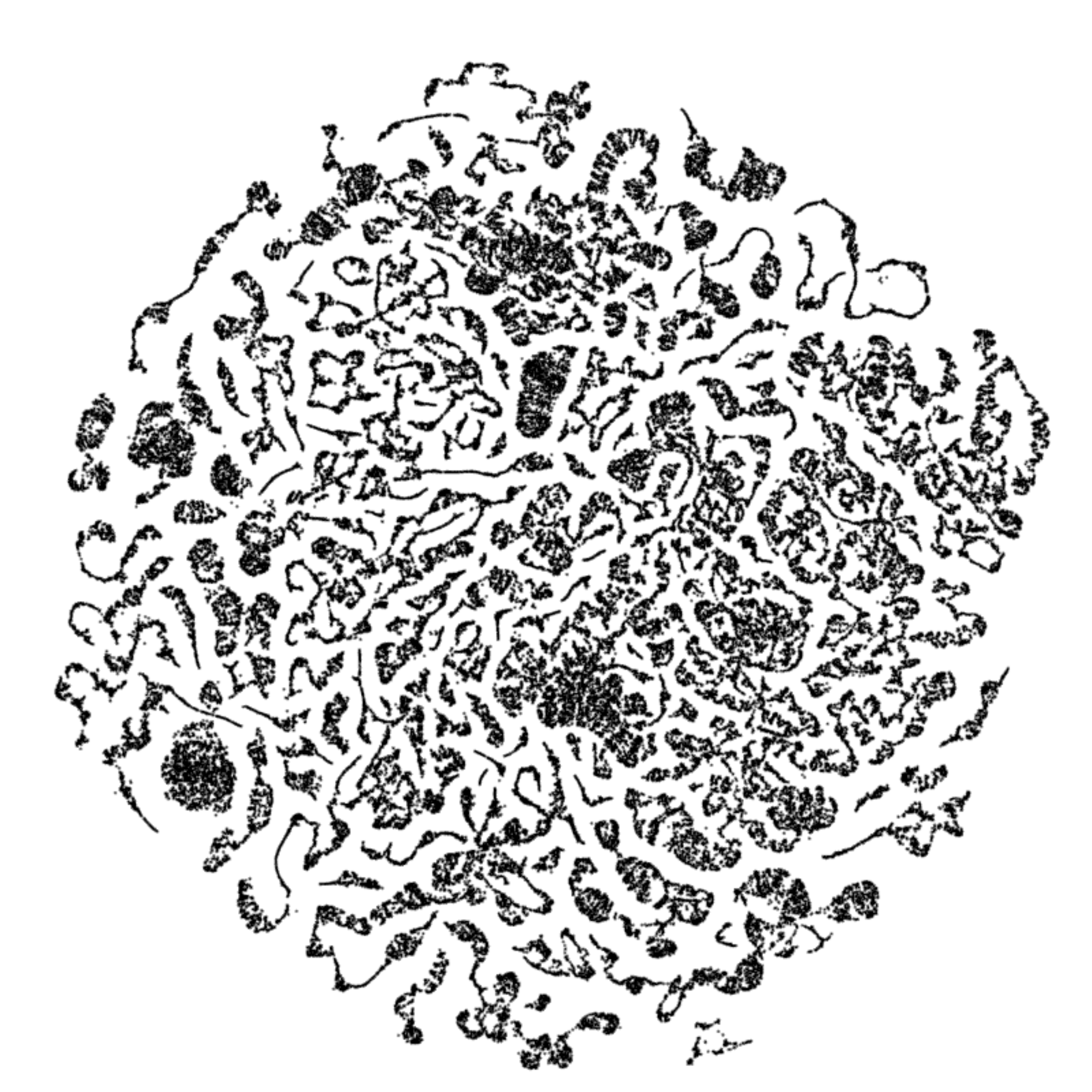}}}
	\subcaptionbox{degree=4}{\fbox{\includegraphics[width=\degreefigwidth,height=\degreefigwidth]{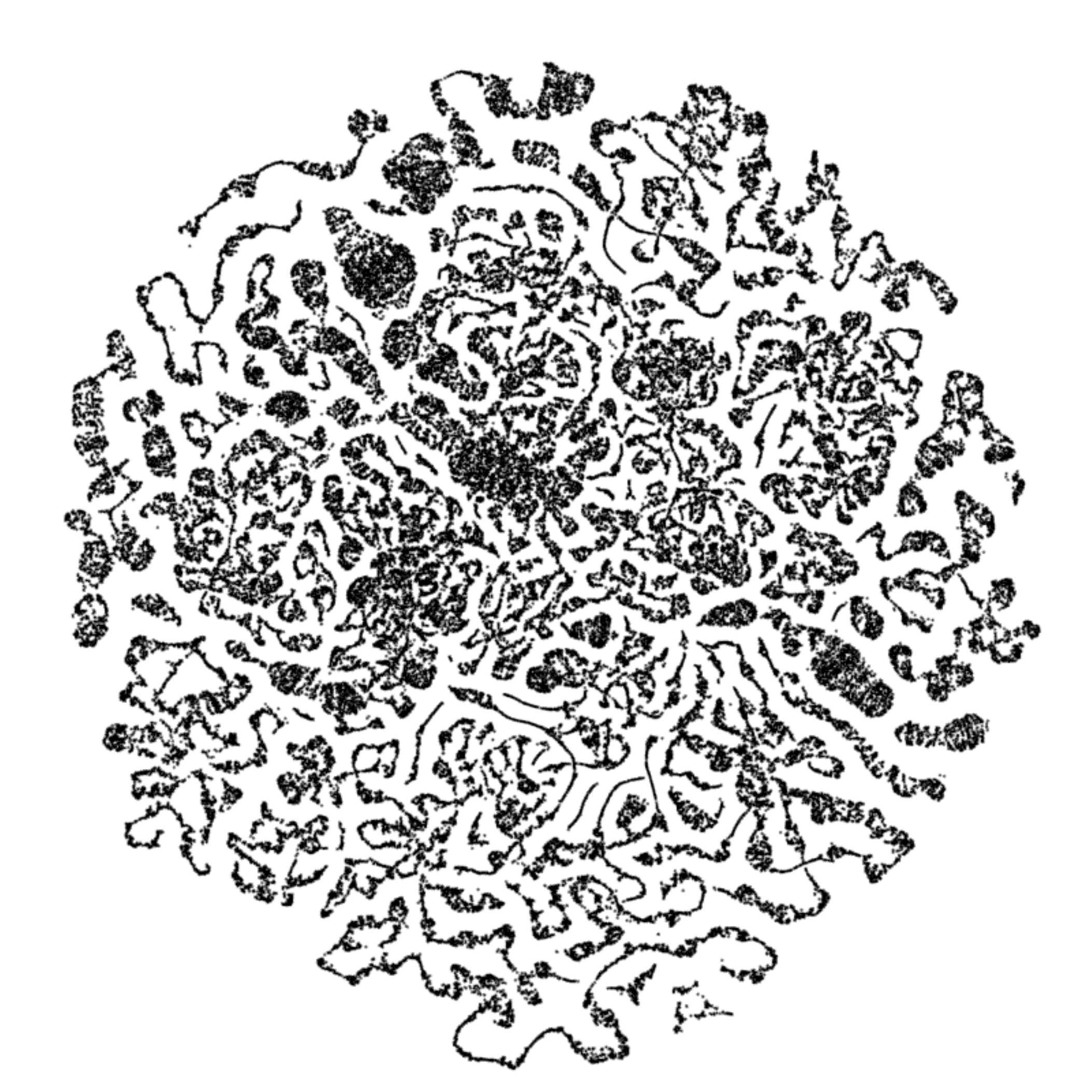}}}
	
	\vspace{\baselineskip}
	
	\subcaptionbox{degree=5}{\fbox{\includegraphics[width=\degreefigwidth,height=\degreefigwidth]{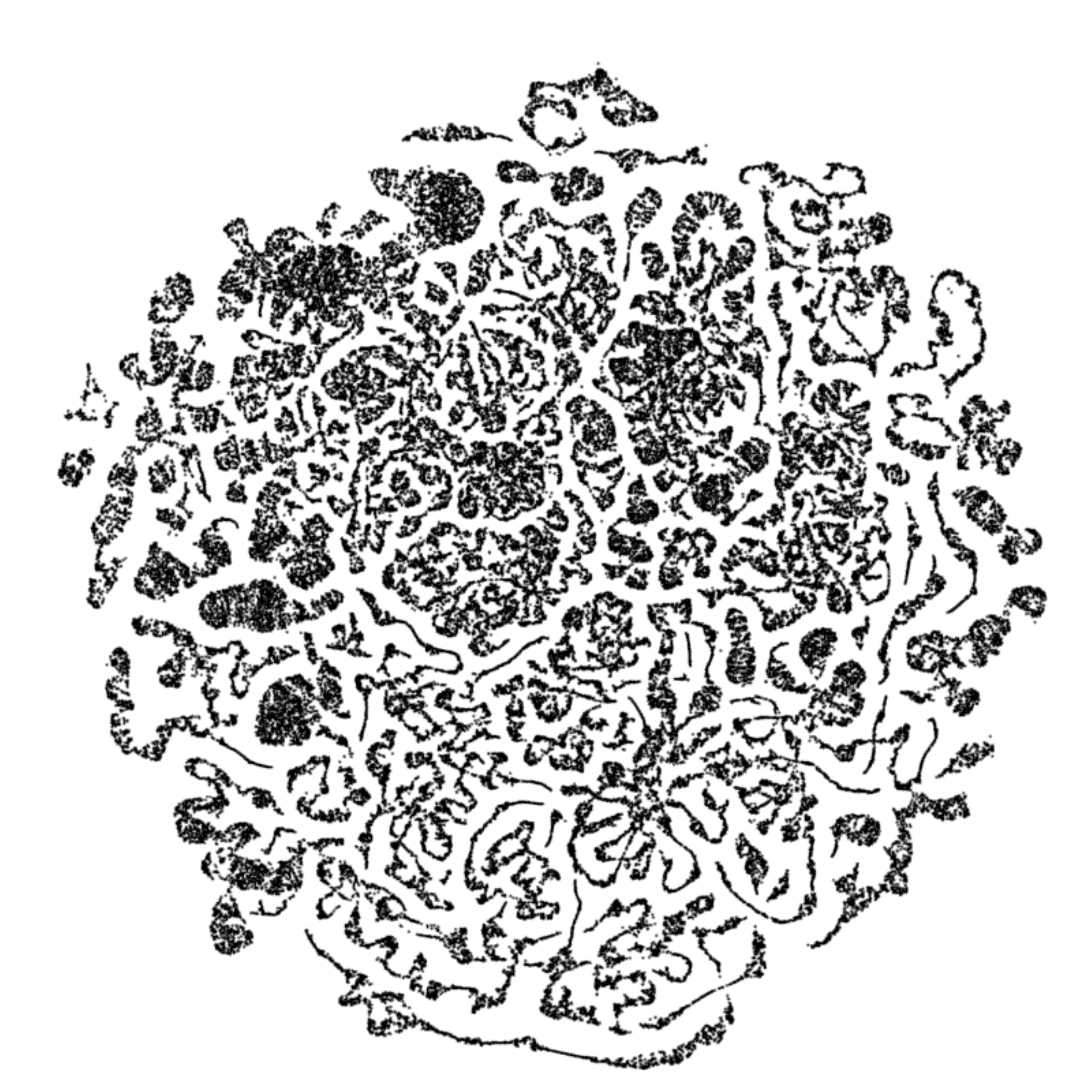}}}
	\subcaptionbox{degree=7}{\fbox{\includegraphics[width=\degreefigwidth,height=\degreefigwidth]{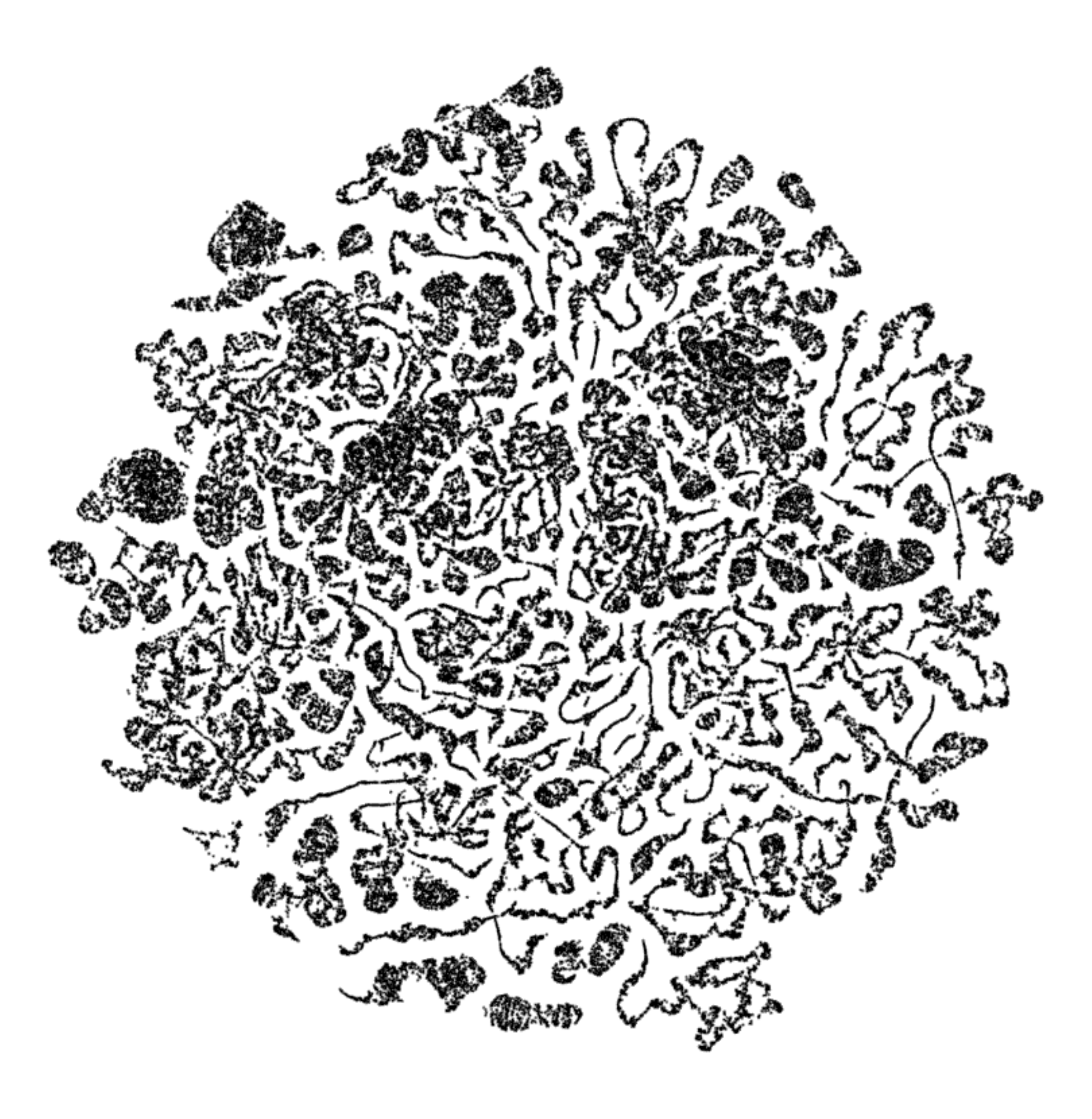}}}
	\subcaptionbox{degree=9}{\fbox{\includegraphics[width=\degreefigwidth,height=\degreefigwidth]{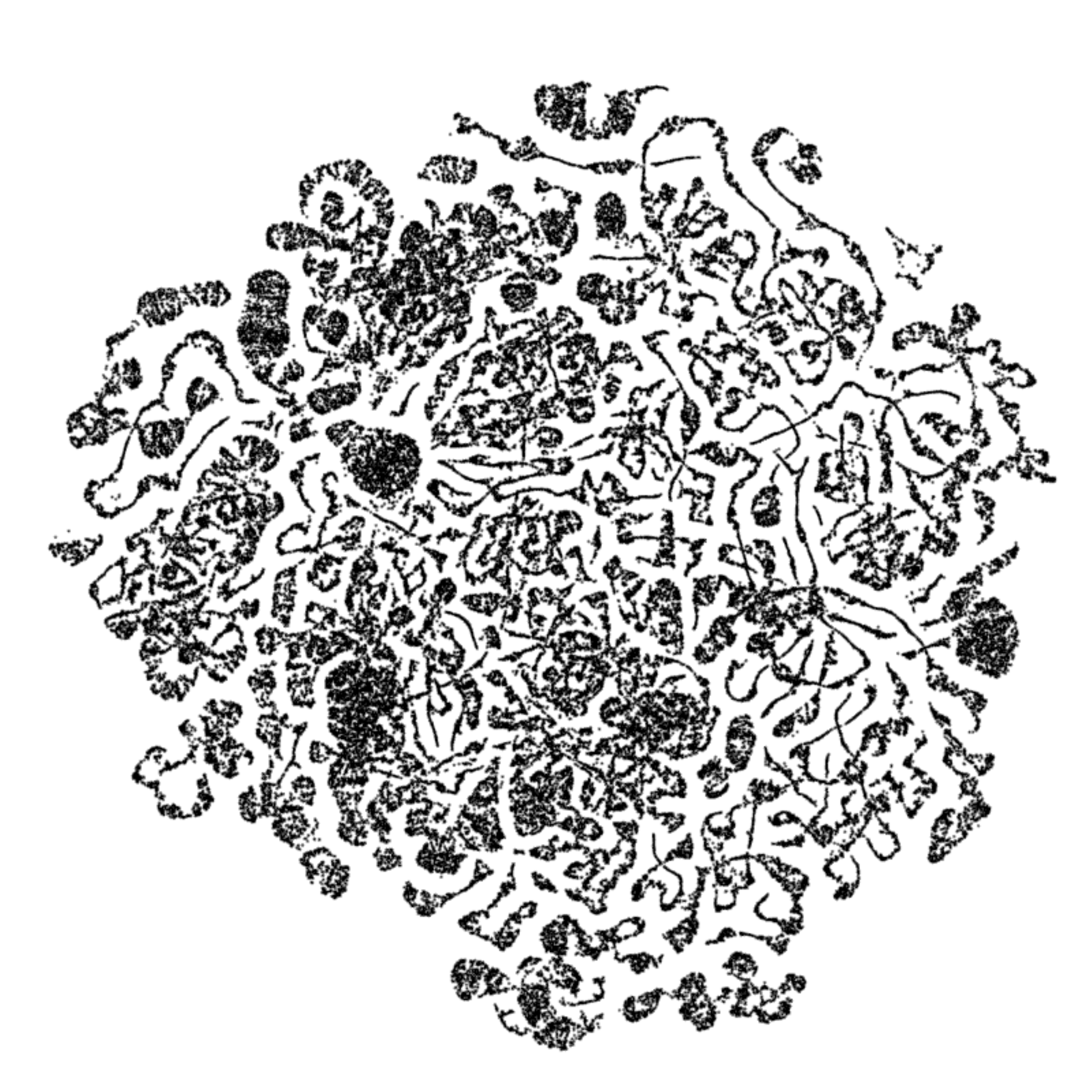}}}
	
	\vspace{\baselineskip}
	
	\subcaptionbox{degree=11}{\fbox{\includegraphics[width=\degreefigwidth,height=\degreefigwidth]{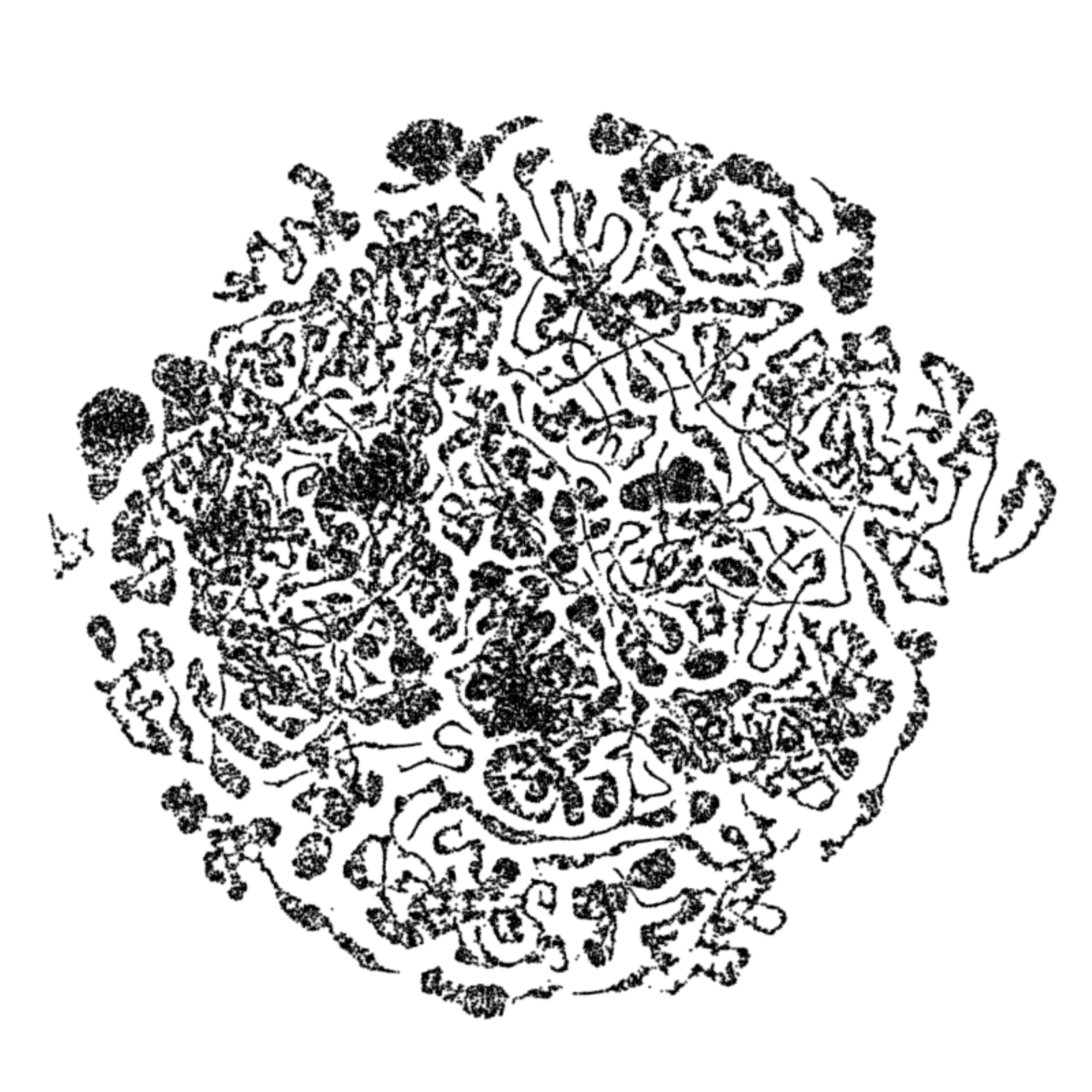}}}
	\subcaptionbox{degree=13}{\fbox{\includegraphics[width=\degreefigwidth,height=\degreefigwidth]{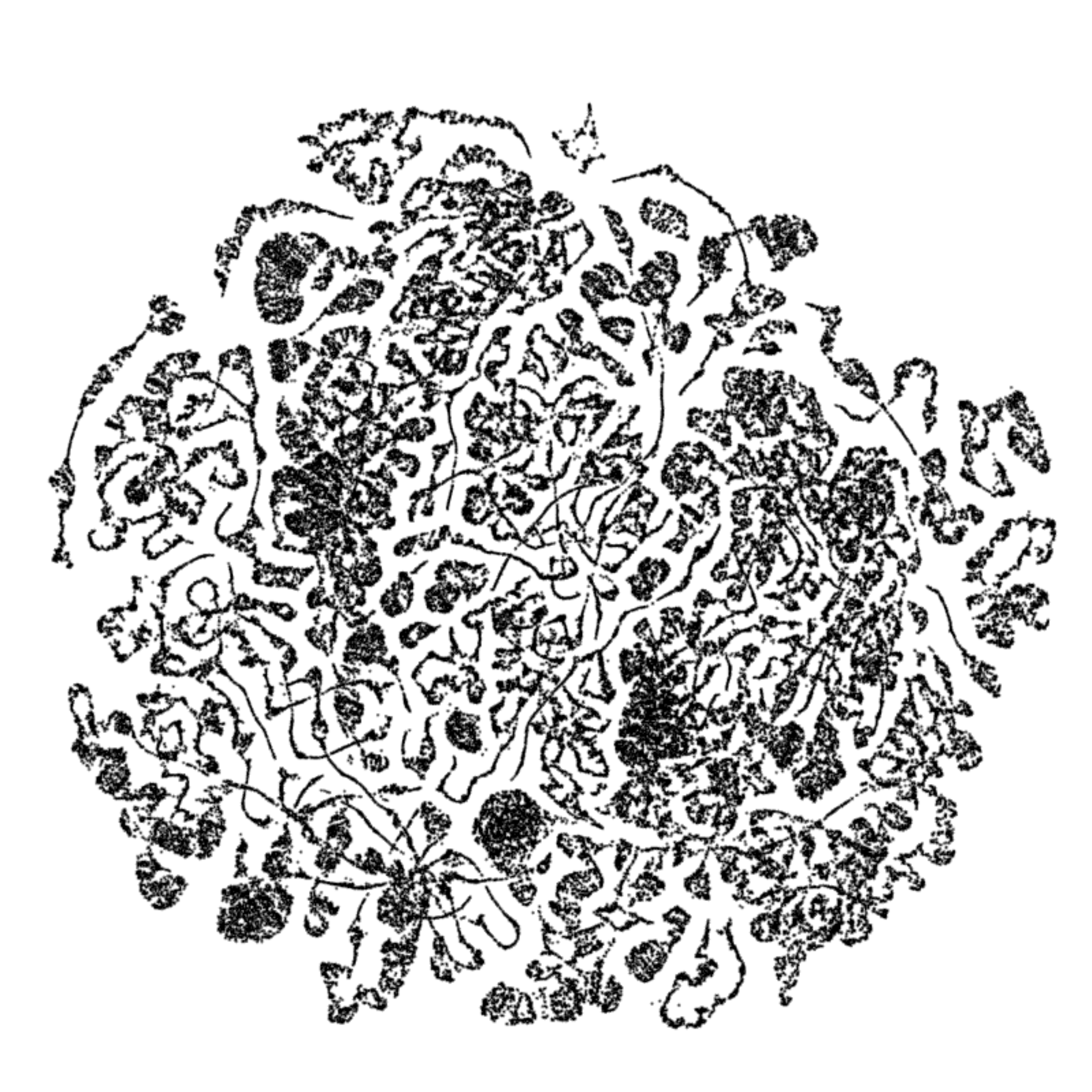}}}
	\subcaptionbox{degree=15}{\fbox{\includegraphics[width=\degreefigwidth,height=\degreefigwidth]{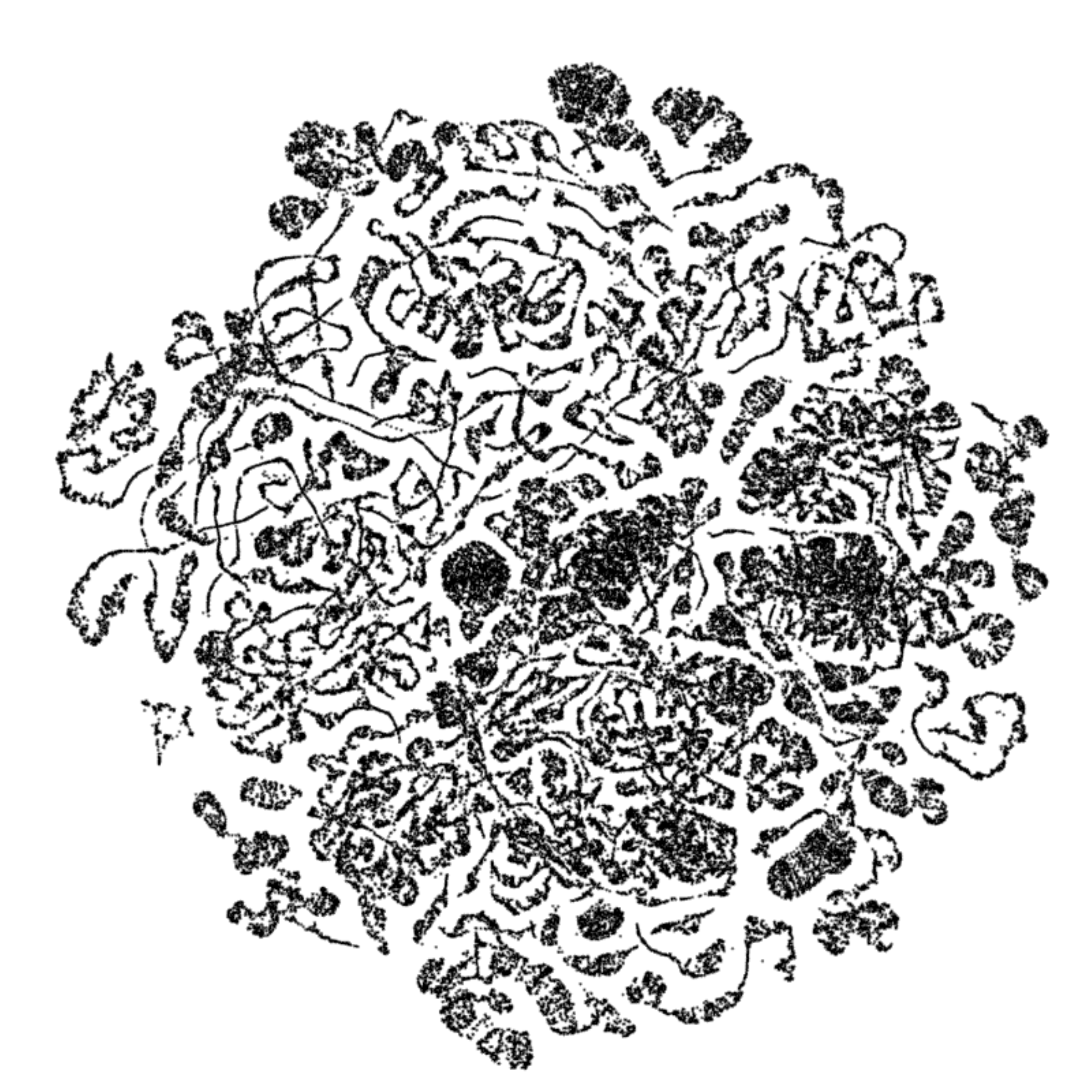}}}
	\caption{Visualizations of the \texttt{TOMORADAR} data set using t-SNE with various degree of freedom in the Student t-distribution. See Figure \ref{fig:tsnevis} for the standard t-SNE visualization (with degree=1).}
	\label{fig:degreestomoradar}
\end{figure}

\newcommand{\optfigwidth}{7.7cm}
\begin{figure}[p]
	\centering
	\subcaptionbox{t-SNE (original algorithm)}{\fbox{\includegraphics[width=\optfigwidth,height=\optfigwidth]{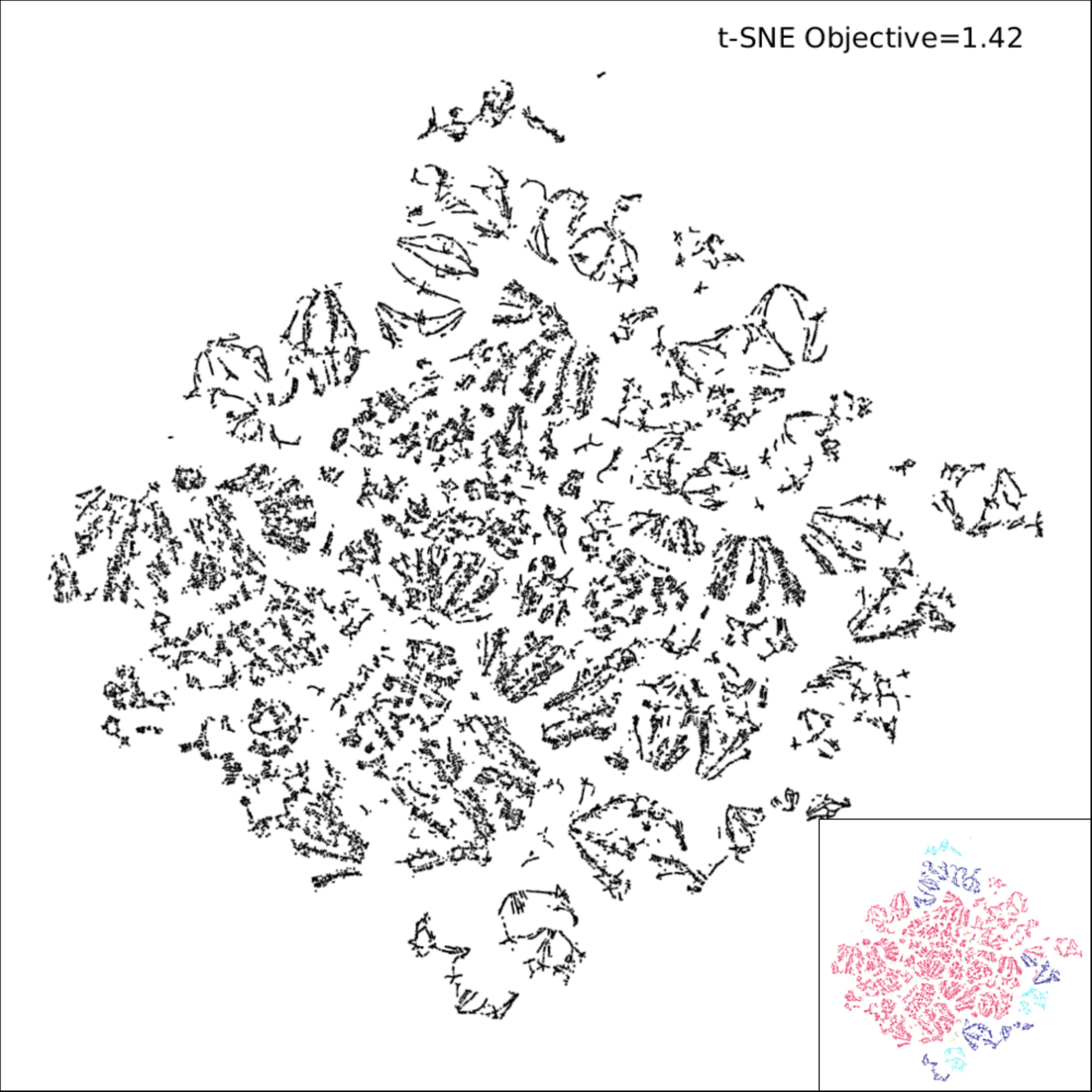}}}
	\subcaptionbox{t-SNE (MM algorithm)}{\fbox{\includegraphics[width=\optfigwidth,height=\optfigwidth]{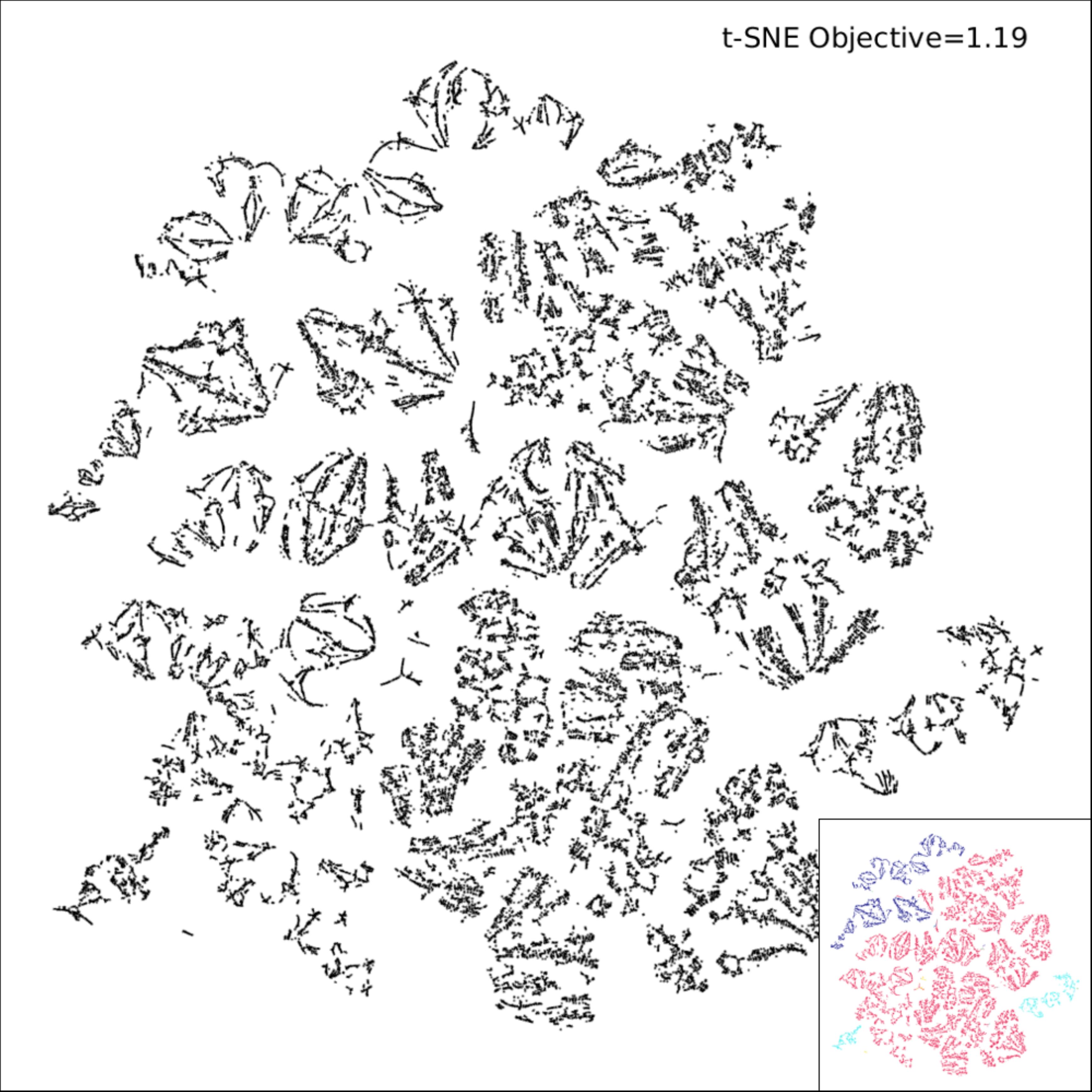}}}
	
	\vspace{\baselineskip}
	
	\subcaptionbox{t-SNE (opt-SNE algorithm)}{\fbox{\includegraphics[width=\optfigwidth,height=\optfigwidth]{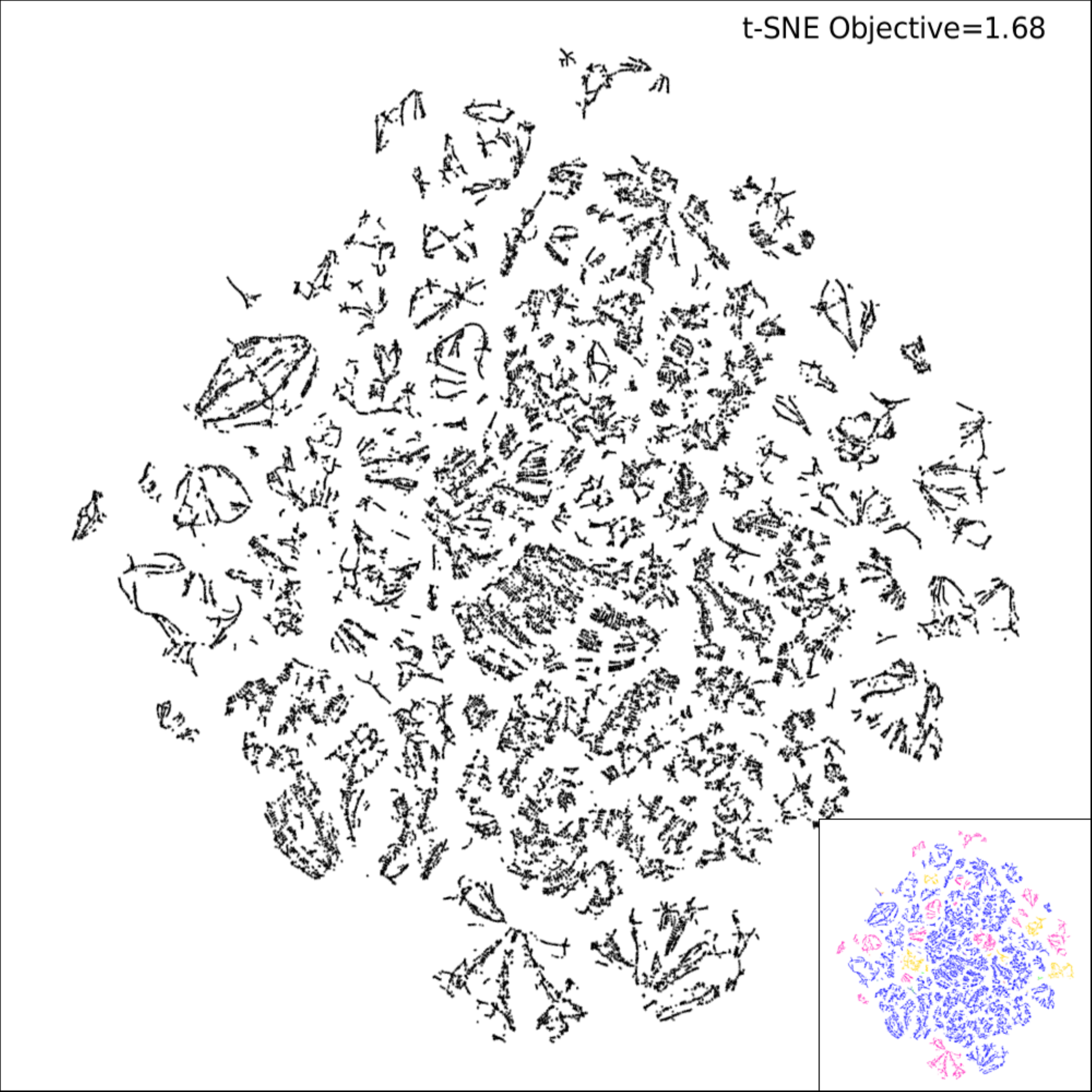}}}
	\subcaptionbox{SCE}{\fbox{\includegraphics[width=\optfigwidth,height=\optfigwidth]{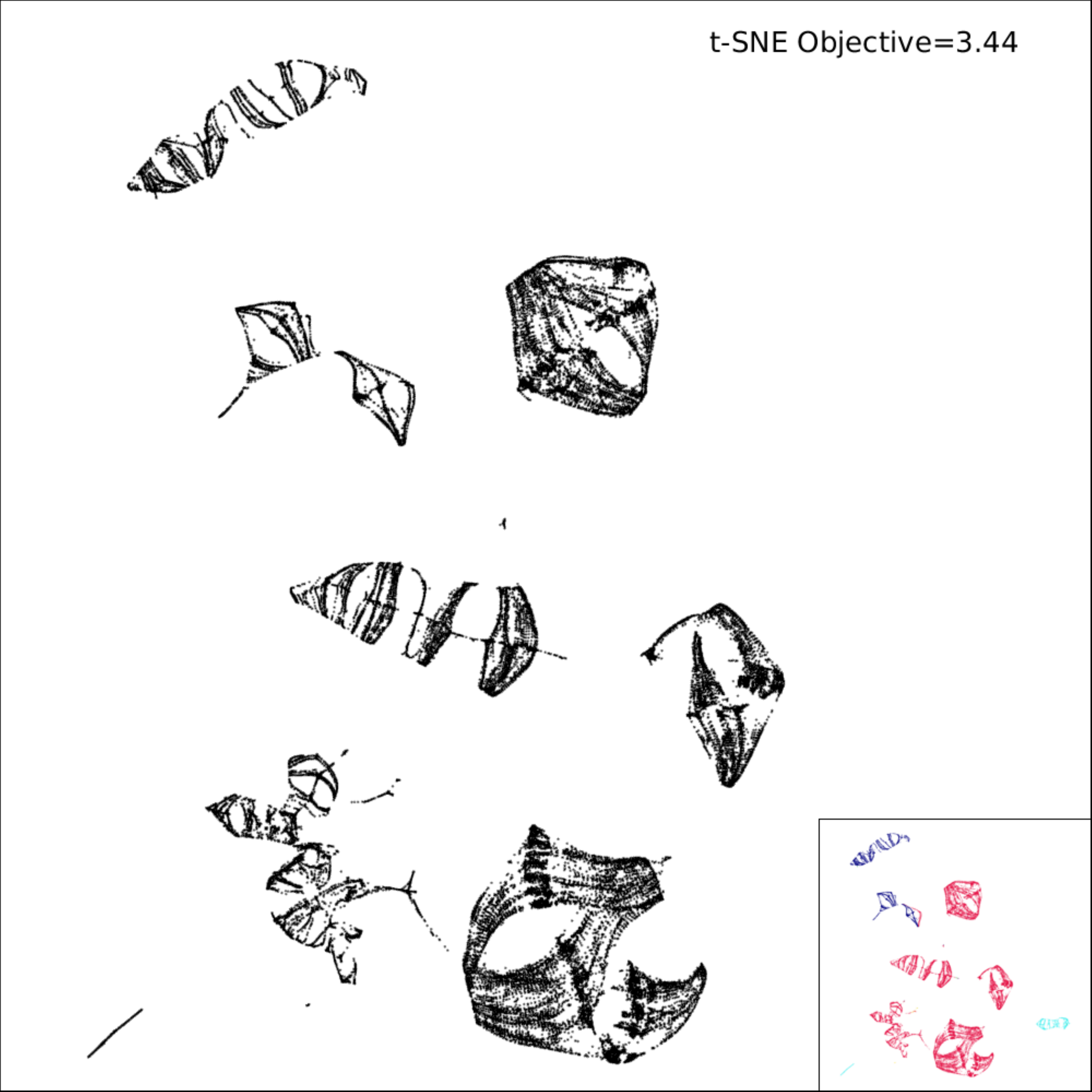}}}
	\caption{Visualizations of t-SNE with different optimization algorithms and SCE for the \texttt{SHUTTLE} data set.}
	\label{fig:opt_shuttle}
\end{figure}

\begin{figure}[p]
	\centering
	\subcaptionbox{t-SNE (original algorithm)}{\fbox{\includegraphics[width=\optfigwidth,height=\optfigwidth]{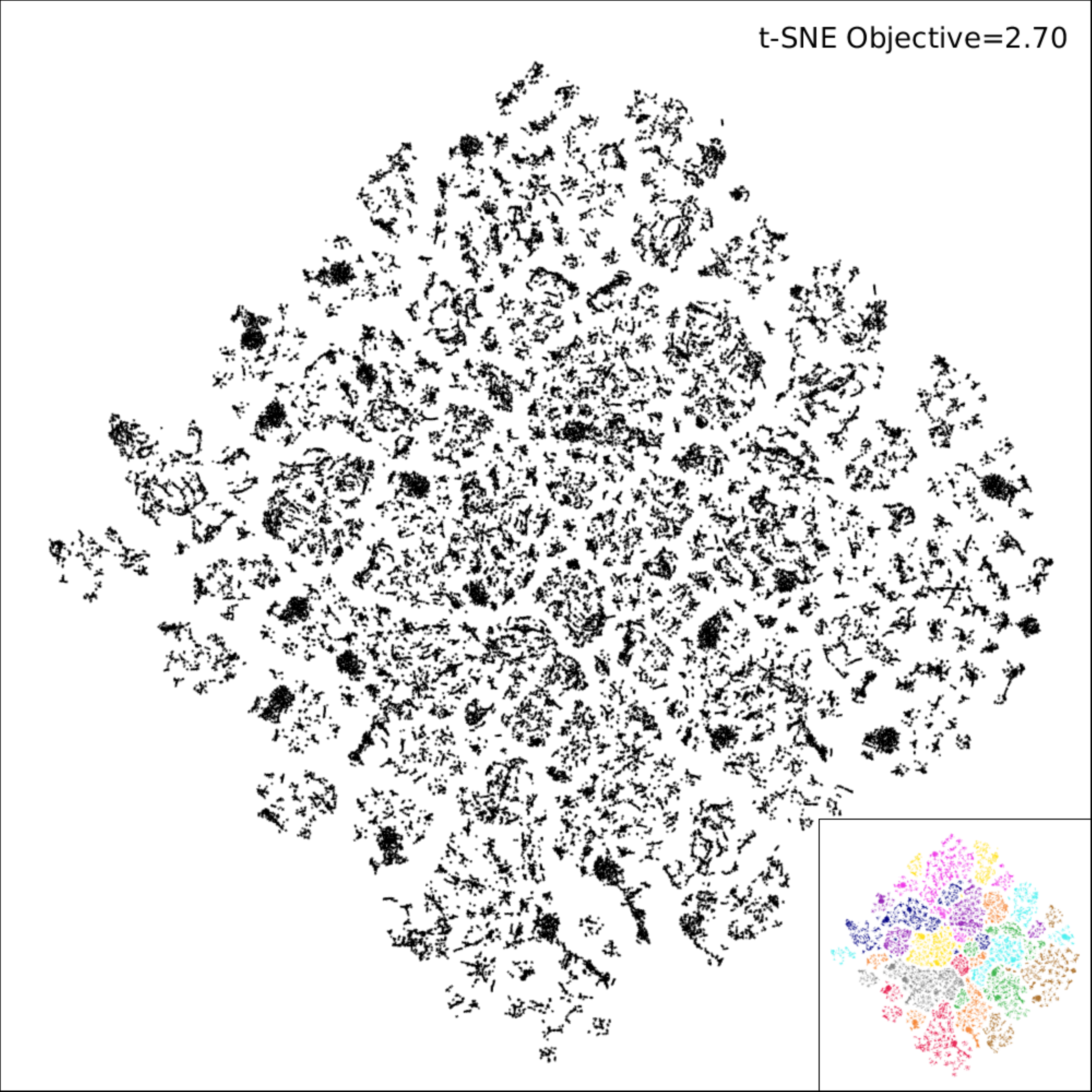}}}
	\subcaptionbox{t-SNE (MM algorithm)}{\fbox{\includegraphics[width=\optfigwidth,height=\optfigwidth]{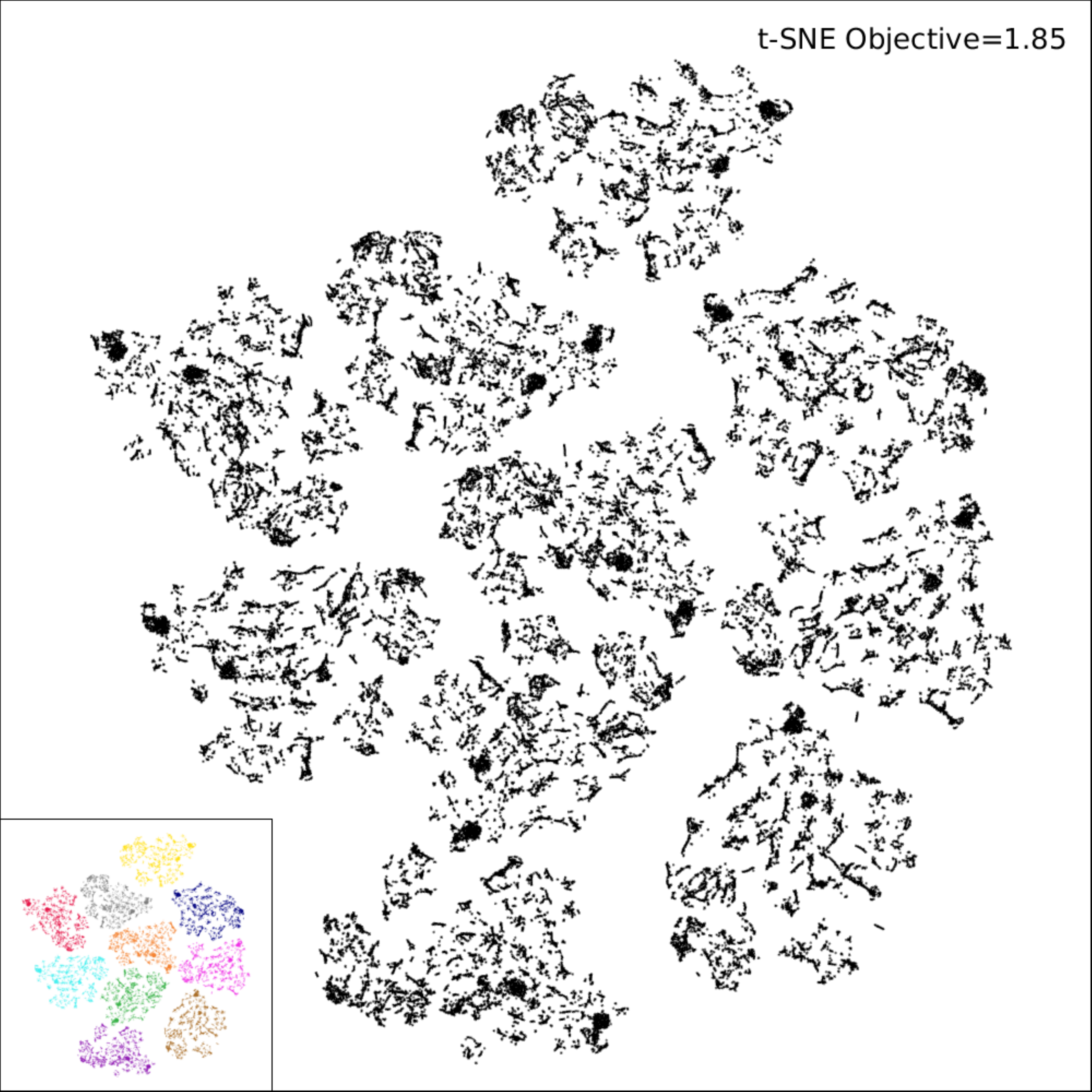}}}
	
	\vspace{\baselineskip}
	
	\subcaptionbox{t-SNE (opt-SNE algorithm)}{\fbox{\includegraphics[width=\optfigwidth,height=\optfigwidth]{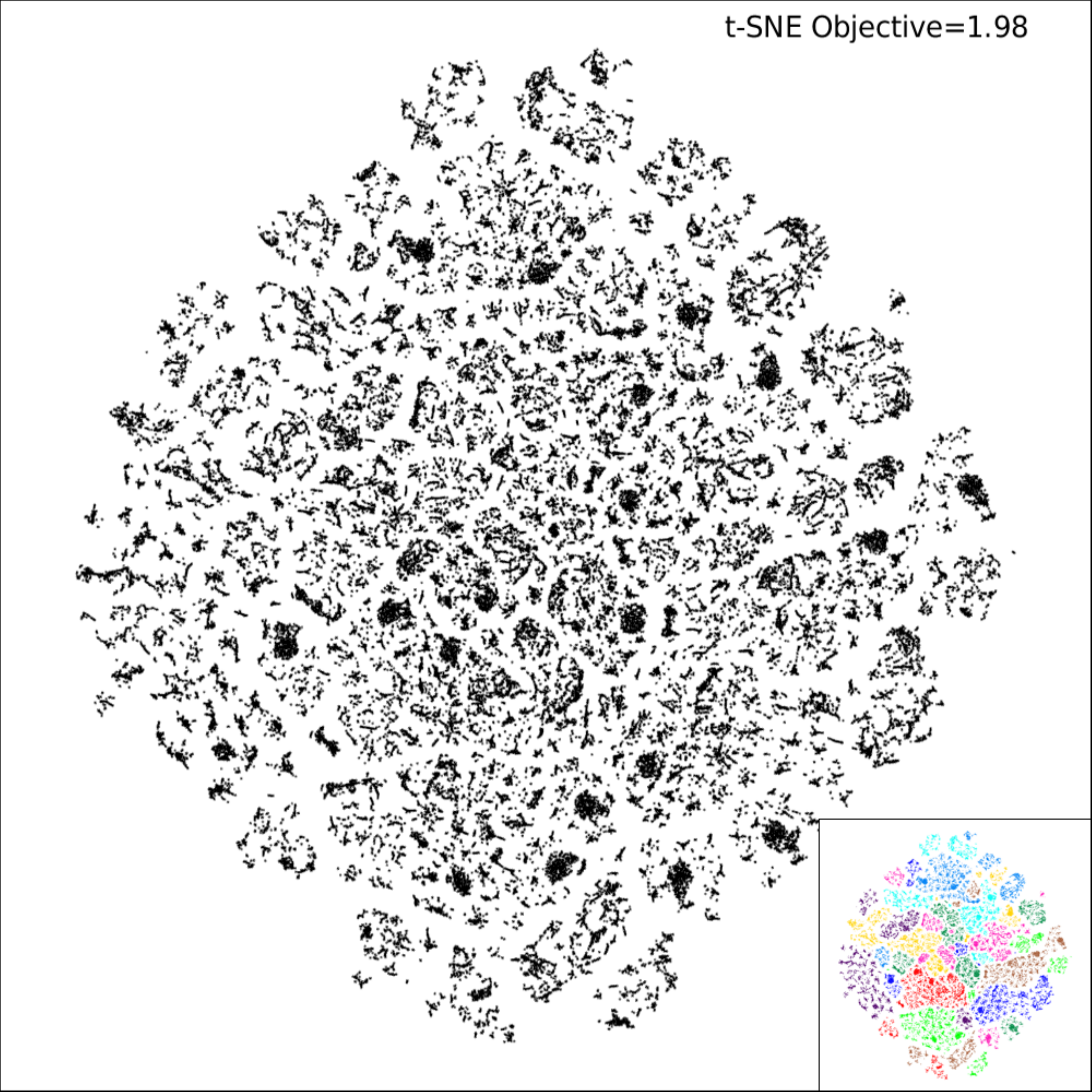}}}
	\subcaptionbox{SCE}{\fbox{\includegraphics[width=\optfigwidth,height=\optfigwidth]{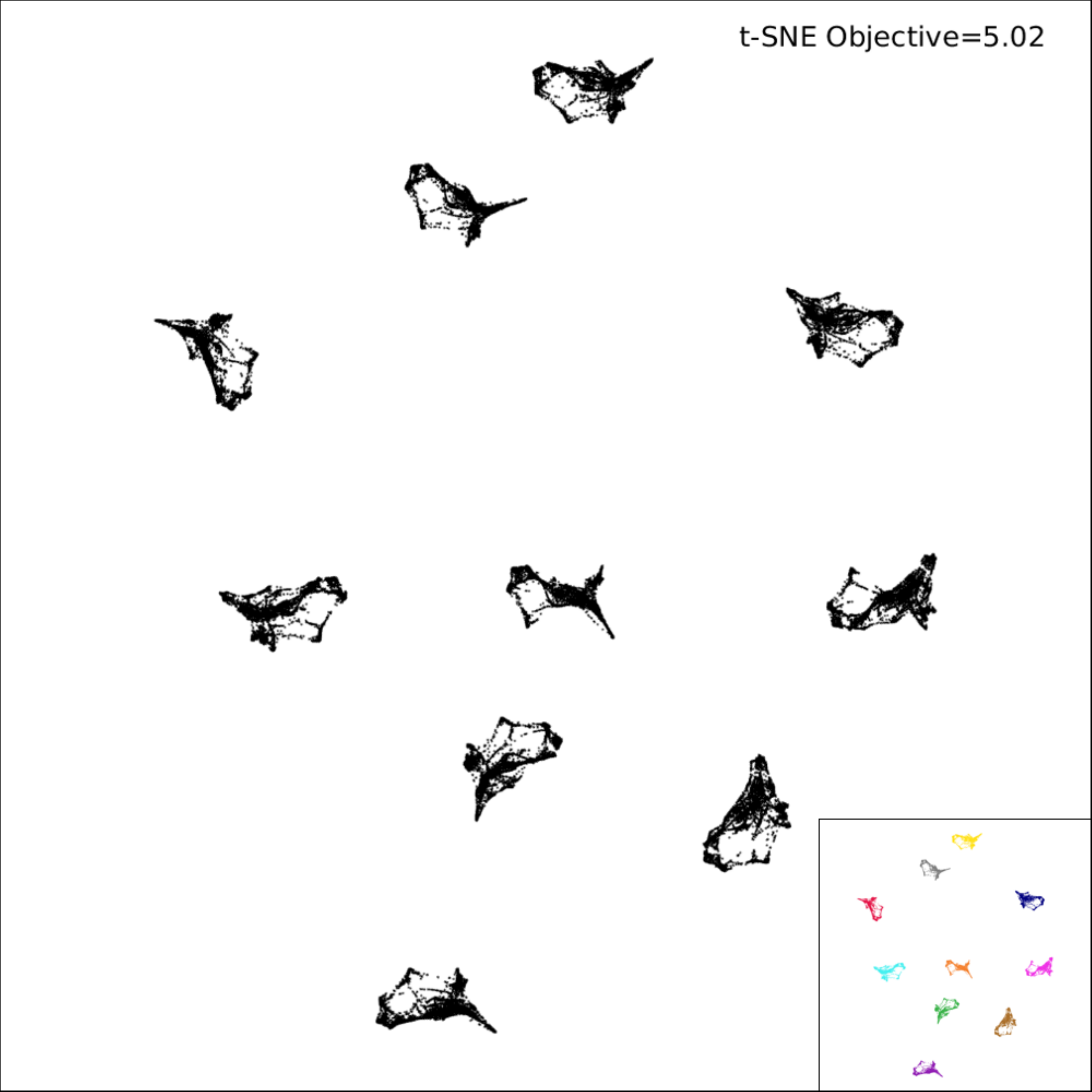}}}
	\caption{Visualizations of t-SNE with different optimization algorithms and SCE for the \texttt{IJCNN} data set.}
	\label{fig:opt_ijcnn}
\end{figure}

\begin{figure}[p]
	\centering
	\subcaptionbox{t-SNE (original algorithm)}{\fbox{\includegraphics[width=\optfigwidth,height=\optfigwidth]{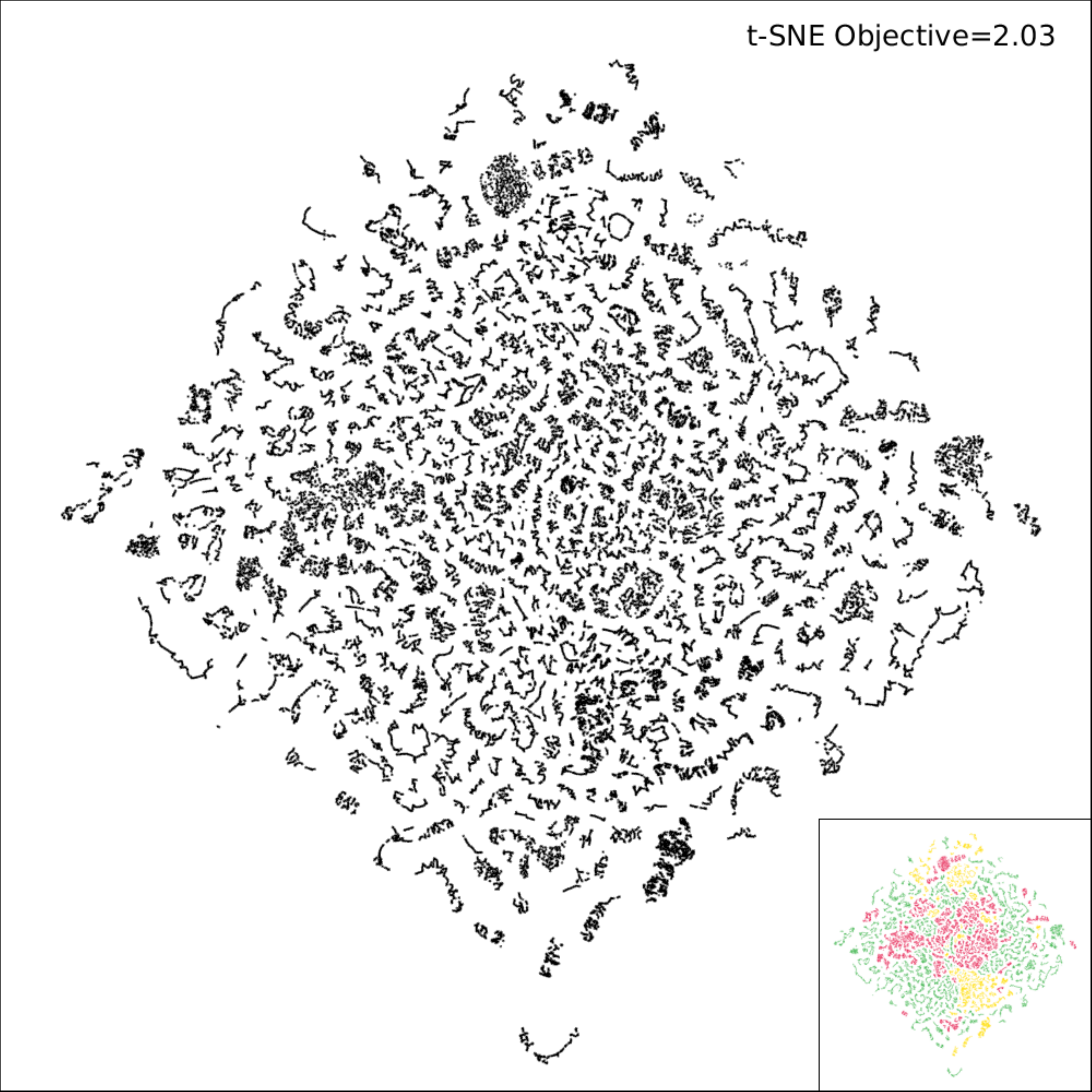}}}
	\subcaptionbox{t-SNE (MM algorithm)}{\fbox{\includegraphics[width=\optfigwidth,height=\optfigwidth]{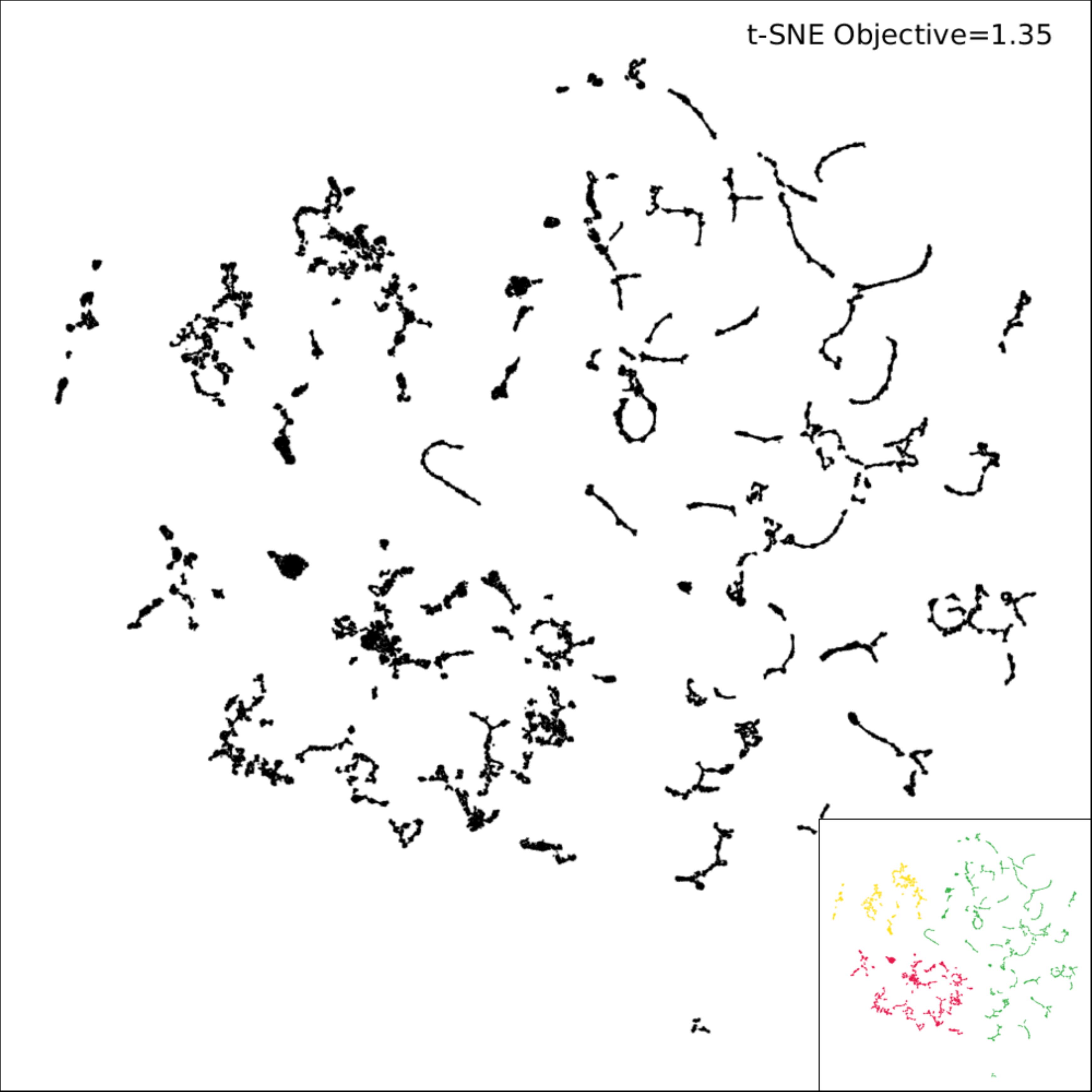}}}
	
	\vspace{\baselineskip}
	
	\subcaptionbox{t-SNE (opt-SNE algorithm)}{\fbox{\includegraphics[width=\optfigwidth,height=\optfigwidth]{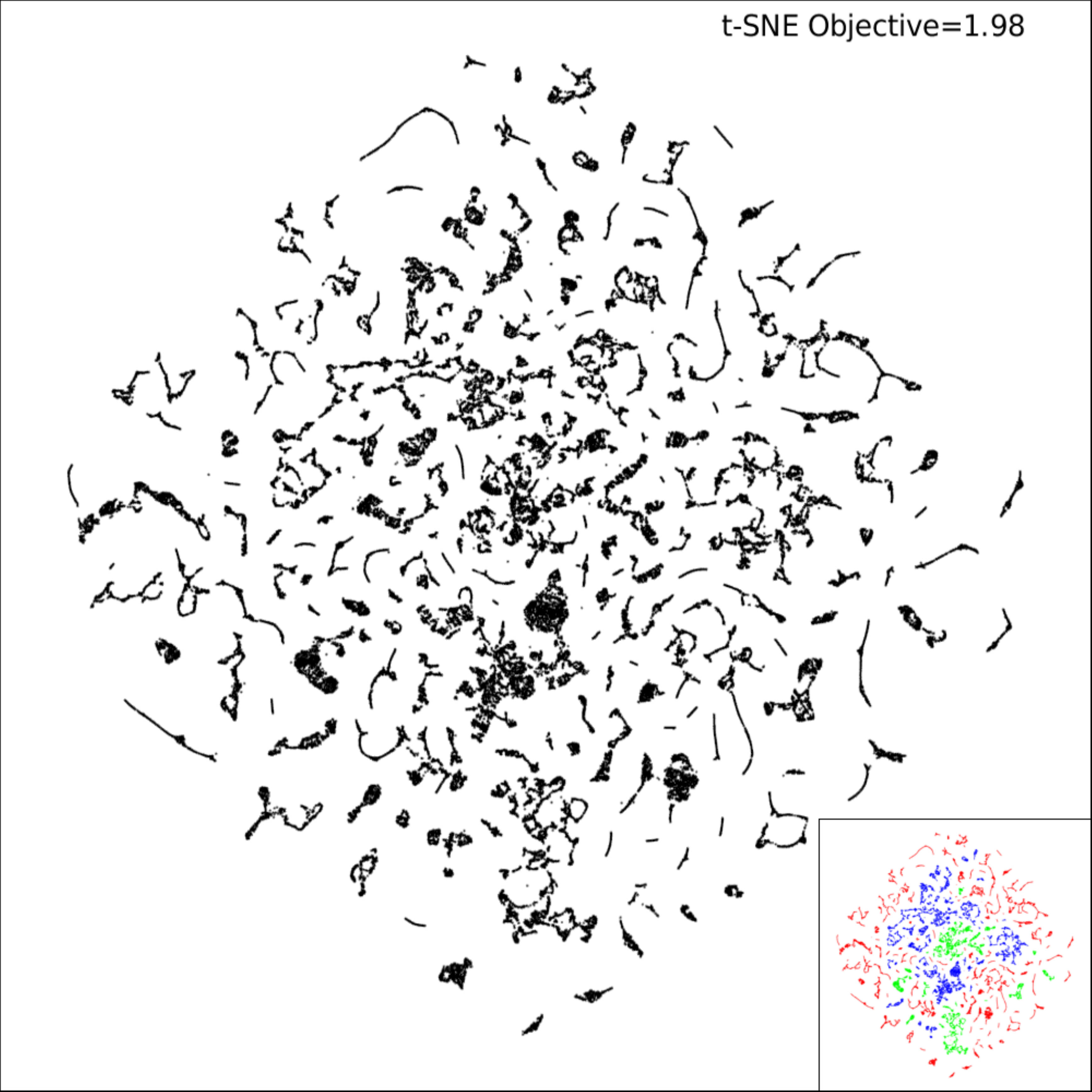}}}
	\subcaptionbox{SCE}{\fbox{\includegraphics[width=\optfigwidth,height=\optfigwidth]{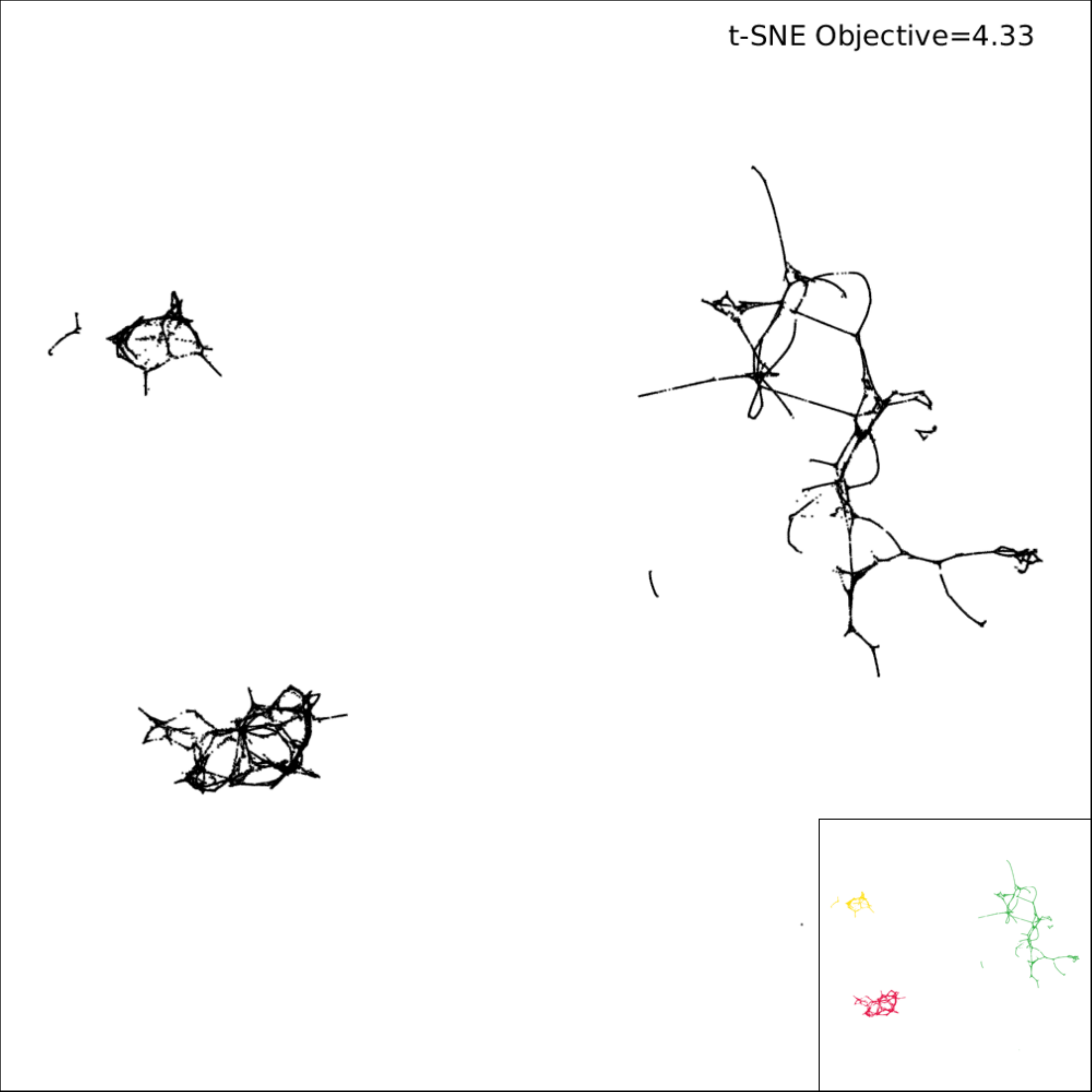}}}
	\caption{Visualizations of t-SNE with different optimization algorithms and SCE for the \texttt{TOMORADAR} data set.}
	\label{fig:opt_tomoradar}
\end{figure}

\clearpage
\newcommand{\synfigwidth}{4cm}
\begin{figure}[t]
\begin{center}
	\subcaptionbox{\texttt{2-MOON}}{\fbox{\includegraphics[width=\synfigwidth,height=\synfigwidth]{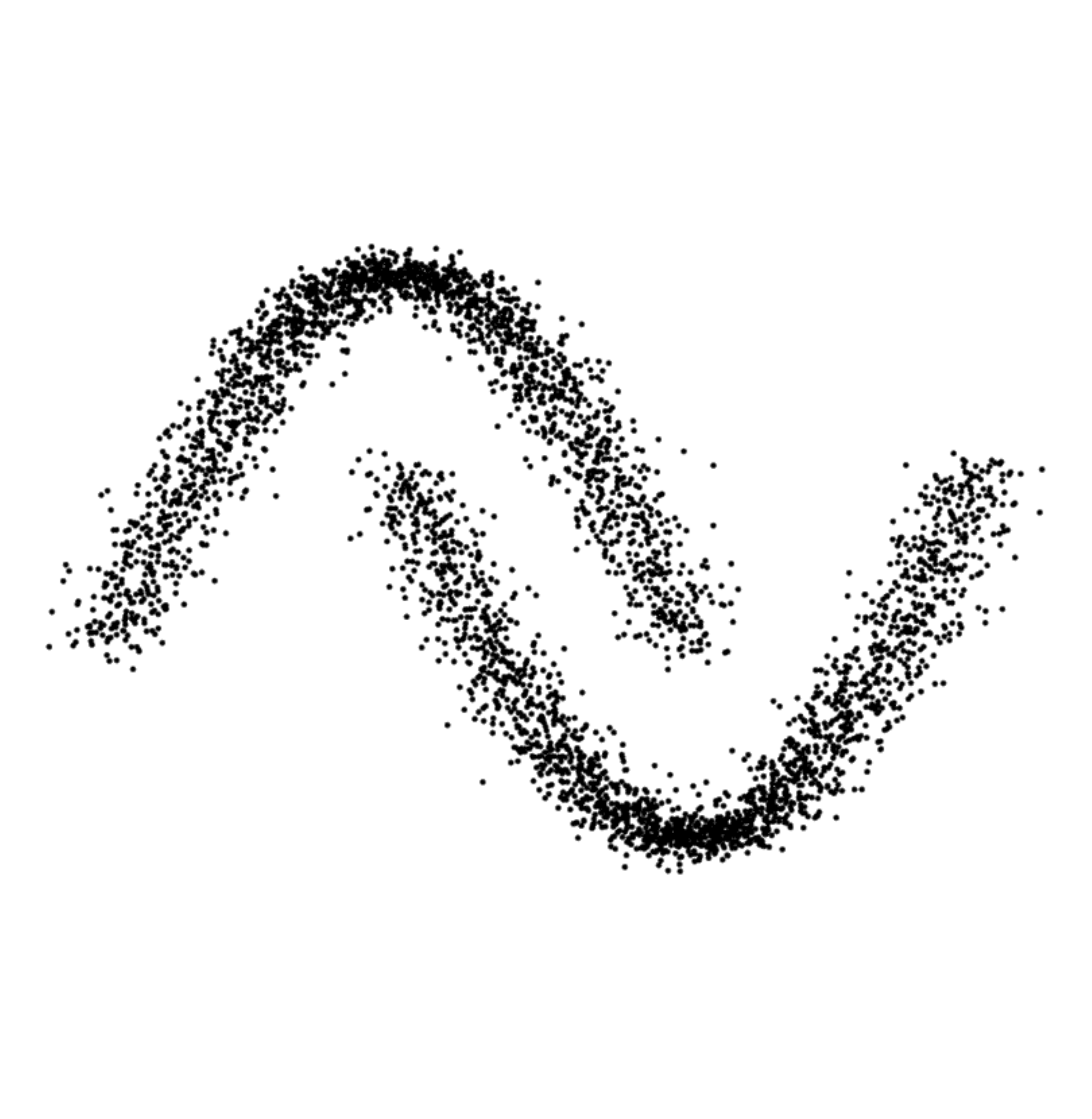}}}
	\subcaptionbox{\texttt{2-ROLL}}{\fbox{\includegraphics[width=\synfigwidth,height=\synfigwidth]{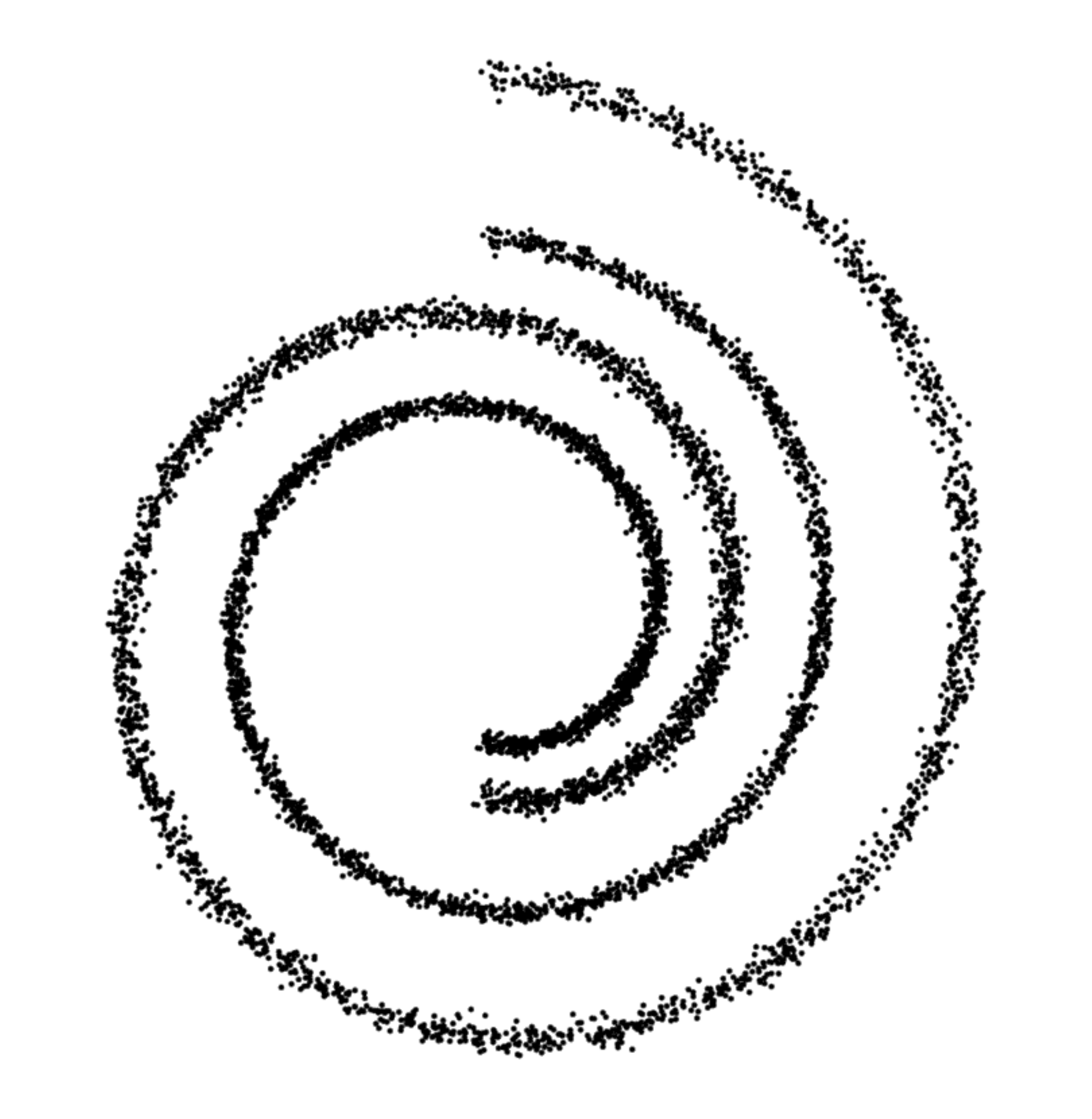}}}
	\caption{Counter examples that violate assumptions made in \cite{tsneprove} and in \cite{tsneanalysis}}
	\label{fig:synthetic}
\end{center}
\end{figure}

\begin{table}[t]
	\begin{center}
	\caption{Violation percentages of the condition in \cite{tsneprove} under various perplexities (perp.).}
	\label{tab:lindermancheck}
	\begin{tabular}{crrrrrrrrr}
		\hline\\[-2mm]
		perp. & \texttt{2-MOON} & \texttt{2-ROLL} & \texttt{COIL20} & \texttt{DIGITS} & \texttt{MNIST10K} & \texttt{SHUTTLE} & \texttt{MNIST} & \texttt{IJCNN} & \texttt{TOMORADAR}\\
		\hline\\[-2mm]
		5 & 98.33 & 98.97 & 61.77 & 74.65 & 83.44 & 97.90 & 90.76 & 96.14 & 98.45\\
		10 & 96.82 & 98.12 & 35.72 & 58.27 & 70.84 & 95.85 & 82.84 & 92.09 & 96.51\\
		15 & 95.49 & 97.26 & 23.12 & 47.97 & 62.23 & 94.12 & 77.00 & 88.91 & 94.75\\
		20 & 94.32 & 96.38 & 15.79 & 40.55 & 55.76 & 92.57 & 72.36 & 86.43 & 93.20\\
		30 & 92.29 & 94.71 & 9.74 & 30.35 & 46.47 & 89.92 & 65.18 & 82.55 & 90.58\\
		50 & 89.04 & 91.32 & 6.48 & 18.81 & 35.08 & 85.98 & 55.31 & 76.70 & 86.50\\
		100 & 82.94 & 82.12 & 3.40 & 7.44 & 21.66 & 78.69 & 41.21 & 69.19 & 79.42\\
		200 & 72.50 & 71.51 & 1.18 & 1.99 & 11.54 & 68.90 & 27.73 & 61.53 & 70.58\\
		500 & 37.15 & 41.08 & 0.01 & 0.14 & 2.19 & 51.20 & 13.72 & 50.05 & 55.90\\
		\hline
	\end{tabular}
	\end{center}
\end{table}

\clearpage

\begin{table}[p]
	\begin{center}
		\caption{Violation percentages of conditions in \cite{tsneanalysis} under various $\gamma$ values: (top) Condition 1a for $\gamma$-Spherical and (bottom) Condition 2 for $\gamma$-Well-separated.}
		\label{tab:aroracheck}
		\begin{tabular}{crrrrrrrrr}
			\hline\\[-2mm]
			$\gamma$ & \texttt{2-MOON} & \texttt{2-ROLL} & \texttt{COIL20} & \texttt{DIGITS} & \texttt{MNIST10K} & \texttt{SHUTTLE} & \texttt{MNIST} & \texttt{IJCNN} & \texttt{TOMORADAR}\\
			\hline\\[-2mm]
			$10^{-10}$ & 100.00 & 100.00 & 100.00 & 100.00 & 100.00 & 100.00 & 100.00 & 100.00 & 100.00\\
			$10^{-9}$ & 100.00 & 100.00 & 100.00 & 100.00 & 100.00 & 100.00 & 100.00 & 100.00 & 100.00\\
			$10^{-8}$ & 100.00 & 100.00 & 100.00 & 100.00 & 100.00 & 100.00 & 100.00 & 100.00 & 100.00\\
			$10^{-7}$ & 100.00 & 100.00 & 100.00 & 100.00 & 100.00 & 100.00 & 100.00 & 100.00 & 100.00\\
			$10^{-6}$ & 100.00 & 100.00 & 100.00 & 100.00 & 100.00 & 100.00 & 100.00 & 100.00 & 100.00\\
			$10^{-5}$ & 100.00 & 100.00 & 100.00 & 100.00 & 100.00 & 100.00 & 100.00 & 100.00 & 100.00\\
			$10^{-4}$ & 100.00 & 100.00 & 100.00 & 100.00 & 100.00 & 100.00 & 100.00 & 100.00 & 100.00\\
			$10^{-3}$ & 100.00 & 100.00 & 100.00 & 100.00 & 100.00 & 100.00 & 100.00 & 100.00 & 100.00\\
			$10^{-2}$ & 100.00 & 100.00 & 100.00 & 100.00 & 100.00 & 100.00 & 100.00 & 100.00 & 100.00\\
			$10^{-1}$ & 100.00 & 100.00 & 100.00 & 100.00 & 100.00 & 100.00 & 100.00 & 100.00 & 100.00\\
			$10^{0}$ & 100.00 & 100.00 & 100.00 & 100.00 & 100.00 & 100.00 & 100.00 & 100.00 & 100.00\\
			$10^{1}$ & 100.00 & 100.00 & 90.00 & 100.00 & 100.00 & 100.00 & 100.00 & 100.00 & 100.00\\
			$10^{2}$ & 100.00 & 100.00 & 0.00 & 10.00 & 16.67 & 100.00 & 10.00 & 100.00 & 100.00\\
			$10^{3}$ & 100.00 & 100.00 & 0.00 & 0.00 & 0.00 & 100.00 & 0.00 & 100.00 & 33.33\\
			$10^{4}$ & 100.00 & 100.00 & 0.00 & 0.00 & 0.00 & 57.14 & 0.00 & 100.00 & 33.33\\
			$10^{5}$ & 100.00 & 100.00 & 0.00 & 0.00 & 0.00 & 28.57 & 0.00 & 0.00 & 0.00\\
			$10^{6}$ & 100.00 & 100.00 & 0.00 & 0.00 & 0.00 & 28.57 & 0.00 & 0.00 & 0.00\\
			$10^{7}$ & 100.00 & 100.00 & 0.00 & 0.00 & 0.00 & 28.57 & 0.00 & 0.00 & 0.00\\
			$10^{8}$ & 50.00 & 0.00 & 0.00 & 0.00 & 0.00 & 14.29 & 0.00 & 0.00 & 0.00\\
			$10^{9}$ & 0.00 & 0.00 & 0.00 & 0.00 & 0.00 & 0.00 & 0.00 & 0.00 & 0.00\\
			$10^{10}$ & 0.00 & 0.00 & 0.00 & 0.00 & 0.00 & 0.00 & 0.00 & 0.00 & 0.00\\
			\hline
		\end{tabular}
		
		\vspace{2mm}		
		
		\begin{tabular}{crrrrrrrrr}
			\hline\\[-2mm]
			$\gamma$ & \texttt{2-MOON} & \texttt{2-ROLL} & \texttt{COIL20} & \texttt{DIGITS} & \texttt{MNIST10K} & \texttt{SHUTTLE} & \texttt{MNIST} & \texttt{IJCNN} & \texttt{TOMORADAR}\\
			\hline\\[-2mm]
			$10^{-10}$ & 54.14 & 90.35 & 52.59 & 82.88 & 95.31 & 95.91 & 98.69 & 98.07 & 99.99\\
			$10^{-9}$ & 54.14 & 90.35 & 52.59 & 82.88 & 95.31 & 95.91 & 98.69 & 98.07 & 99.99\\
			$10^{-8}$ & 54.14 & 90.35 & 52.59 & 82.89 & 95.31 & 95.91 & 98.69 & 98.07 & 99.99\\
			$10^{-7}$ & 54.14 & 90.35 & 52.59 & 82.89 & 95.31 & 95.91 & 98.69 & 98.07 & 99.99\\
			$10^{-6}$ & 54.14 & 90.35 & 52.59 & 82.89 & 95.31 & 95.91 & 98.69 & 98.07 & 99.99\\
			$10^{-5}$ & 54.14 & 90.36 & 52.60 & 82.89 & 95.32 & 95.91 & 98.69 & 98.07 & 99.99\\
			$10^{-4}$ & 54.17 & 90.38 & 52.66 & 82.96 & 95.36 & 95.92 & 98.71 & 98.08 & 99.99\\
			$10^{-3}$ & 54.48 & 90.63 & 53.26 & 83.71 & 95.73 & 96.04 & 98.89 & 98.23 & 99.99\\
			$10^{-2}$ & 57.62 & 92.80 & 58.99 & 89.91 & 98.30 & 97.15 & 99.78 & 99.21 & 100.00\\
			$10^{-1}$ & 92.32 & 100.00 & 86.78 & 100.00 & 100.00 & 99.87 & 100.00 & 100.00 & 100.00\\
			$10^{0}$ & 100.00 & 100.00 & 98.94 & 100.00 & 100.00 & 100.00 & 100.00 & 100.00 & 100.00\\
			$10^{1}$ & 100.00 & 100.00 & 100.00 & 100.00 & 100.00 & 100.00 & 100.00 & 100.00 & 100.00\\
			$10^{2}$ & 100.00 & 100.00 & 100.00 & 100.00 & 100.00 & 100.00 & 100.00 & 100.00 & 100.00\\
			$10^{3}$ & 100.00 & 100.00 & 100.00 & 100.00 & 100.00 & 100.00 & 100.00 & 100.00 & 100.00\\
			$10^{4}$ & 100.00 & 100.00 & 100.00 & 100.00 & 100.00 & 100.00 & 100.00 & 100.00 & 100.00\\
			$10^{5}$ & 100.00 & 100.00 & 100.00 & 100.00 & 100.00 & 100.00 & 100.00 & 100.00 & 100.00\\
			$10^{6}$ & 100.00 & 100.00 & 100.00 & 100.00 & 100.00 & 100.00 & 100.00 & 100.00 & 100.00\\
			$10^{7}$ & 100.00 & 100.00 & 100.00 & 100.00 & 100.00 & 100.00 & 100.00 & 100.00 & 100.00\\
			$10^{8}$ & 100.00 & 100.00 & 100.00 & 100.00 & 100.00 & 100.00 & 100.00 & 100.00 & 100.00\\
			$10^{9}$ & 100.00 & 100.00 & 100.00 & 100.00 & 100.00 & 100.00 & 100.00 & 100.00 & 100.00\\
			$10^{10}$ & 100.00 & 100.00 & 100.00 & 100.00 & 100.00 & 100.00 & 100.00 & 100.00 & 100.00\\
			\hline
		\end{tabular}
	\end{center}
\end{table}

\clearpage

\bibliographystyle{plain}
%\bibliography{refs}

\end{document}